\documentclass{sn-jnl}
\usepackage[numbers,sort]{natbib}
\usepackage[default]{sourcesanspro}
\usepackage{algpseudocode}
\usepackage{tablefootnote}
\usepackage{algorithm}
\usepackage{dirtytalk}
\usepackage{pdfrender}
\usepackage{pdfpages}
\usepackage{multirow}
\usepackage{amsfonts}
\usepackage{amsmath}
\usepackage{xspace}
\usepackage{subfig}
\usepackage{amsthm}
\usepackage{bm}

\definecolor{black}{rgb}		{0.0, 0.0, 0.0}
\definecolor{white}{rgb}		{1.0, 1.0, 1.0}
\definecolor{yellow}{rgb}		{1.0, 1.0, 0.8}
\definecolor{red}{rgb}			{0.6, 0.0, 0.2}
\definecolor{blue}{rgb}		{0.0, 0.2, 0.5}
\definecolor{green}{rgb}		{0.6, 0.8, 0.8}
\definecolor{dark_green}{RGB} {0, 140, 0}
\definecolor{gold}{rgb}		{0.6, 0.4, 0.1}
\definecolor{grey}{RGB}{0,0,0}
\definecolor{Gray}{gray}{0.8}
\definecolor{MediumGray}{gray}{0.9}
\definecolor{LightGray}{gray}{0.98}
\definecolor{LightCyan}{rgb}{0.88,1,1}
\definecolor{purple}{RGB}{128,0,128}
\definecolor{sl_blue}{RGB}{47, 60, 105}
\definecolor{orange}{RGB}{255,165,0}
\definecolor{Gray}{gray}{0.85}

\newcommand{\isep}{,..., }
\DeclareMathOperator*{\argmax}{argmax} 
\DeclareMathOperator*{\argmin}{argmin} 
\newcommand{\HOMER}{HOMER\xspace}
\newcommand{\sctwo}{StarCraft II\xspace}
\newcommand{\E}{\mathbb{E}}

\newcommand{\M}{\mathcal{M}}
\newcommand{\N}{\mathbb{N}}
\newcommand{\nextstate}{s'}
\newcommand{\states}{\mathcal{S}}
\newcommand{\actions}{\mathcal{A}}
\newcommand{\action}{a}
\newcommand{\crlaactions}{\ \mathcal{U}}
\newcommand{\crlaaction}{u}
\newcommand{\continuousactions}{\ \mathcal{U}}
\newcommand{\continuousaction}{u}
\newcommand{\rewards}{\mathcal{R}}
\newcommand{\transition}{\mathcal{P}}
\newcommand{\observations}{\Omega}
\newcommand{\observationf}{\mathcal{O}}
\newcommand{\observation}{o}

\newcommand{\MDP}{\mathcal{M}}
\newcommand{\xmark}{\text{\sffamily X}\xspace}

\newcommand{\Prob}{\mathcal{P}}
\newcommand{\red}{Red\xspace}
\newcommand{\blue}{Blue\xspace}
 
\newcommand{\unseenactions}{Time-Varying Actions (TVA)} 
\newcommand{\generalizability}{Generalizability}  
\newcommand{\hdasp}{high dimensional action space problem\xspace}  
\newcommand{\eg}{e.g., }  
\newcommand{\ie}{i.e., }  
\newcommand{\RL}{RL\xspace} 
\newcommand{\decPOMDP}{Dec-POMDP\xspace}

\newcommand{\cod}{curse of dimensionality\xspace} 
\newcommand{\aco}{ACD\xspace} 

\newcommand{\commentout}[1]{}
\newcommand{\papertitle}{Deep Reinforcement Learning for Autonomous Cyber Defence: A Survey}
\newcommand{\rats}{\mathbb{R}}

\newcommand{\ado}{ADO\xspace}

\newcommand{\wrt}{with regard to\xspace}
\global\long\def\JointPolicy#1#2{\langle #1, #2 \rangle}
\newtheorem{definitionsec}{Definition}[section] 
\newtheorem{theoremsec}{Theorem}

 \providecommand{\argsA}[2]{ {#1}_{#2} } 

\newcommand{\MoP}{Mixture of Policies\xspace}
\newcommand{\mop}{mixture of policies\xspace}

\providecommand{\gets}{$\leftarrow$}
\providecommand{\ExtensiveFormGame}{\mathcal{M}}
\providecommand{\NormalFormGame}{\mathcal{N}}
\providecommand{\Exploitability}{\mathcal{G}_{E}}
\providecommand{\AC}{\pi}                 
\providecommand{\aA}[1]{\argsA{\AC}{#1}}
\providecommand{\funcName}[1]{\textsc{#1}}
\global\long\def\mA#1{\argsA{\mu}{#1}}
\global\long\def\G#1#2{\argsA{\mathcal{G}}{#1}(#2)}

\global\long\def\JointPolicy#1#2{\langle #1, #2 \rangle}
\global\long\def\O#1#2{\argsA{O}{#1}(#2)}
\newcommand{\spacer}{\vspace{1mm}}
\providecommand{\reportonly}[1]{#1}
\newcommand{\acogyms}{\aco-gyms\xspace}
\newcommand{\acogym}{\aco-gym\xspace}

\newcommand{\envs}{envs\xspace}
\newcommand{\env}{env\xspace}
\renewcommand{\WP}{Wolpertinger\xspace}
\newcommand{\report}{survey\xspace} 
\providecommand{\update}[1]{#1}

\renewcommand{\R}{\mathcal{R}}
\renewcommand{\state}{s}
\renewcommand{\reportonly}[1]{}
\renewcommand{\cite}{\citet}
\renewcommand{\envs}{environments\xspace}

\renewcommand{\env}{environment\xspace}
\usepackage[T1]{fontenc}
\usepackage[ttdefault=true]{AnonymousPro}
\newcommand{\zero}{\texttt 0}
\definecolor{red}{rgb}{0.6, 0.0, 0.2}

\jyear{2024}%
\usepackage{csvsimple,filecontents}
\usepackage{graphicx}
\usepackage{multirow}
\usepackage{amsfonts}
\usepackage{xstring}
\usepackage{amsmath}
\usepackage{hhline}
\newcommand{\citeenv}[1]{
    \IfEqCase{#1}{%
        {CybORG}{\citep{cyborg_acd_2021}}%
        {Yawning Titan}{\citep{YAWNING}}%
        {NASim}{\citep{schwartz2019autonomous}}%
        {Intelligent Communication Jamming}{\citep{e24101441}}%
        {Multicast Scheduling Problem}{\citep{li2022multicast}}%
        {Active Network Management}{\citep{HENRY20211000921,HENRY20211000922}}%
        {SUMO}{\citep{SUMO2018}}%
        {Iroko}{\citep{ruffy2018iroko}}%
        {Categorised N-Rooms}{\citep{zahavy2019learn}}%
        {Selective Predator-Prey}{\citep{song2019solving,PredatorPreyExample}}%
        {Multi-Step Plan Environment}{\citep{dulac2015deep}}%
        {Platform Jumping}{\citep{wei2018hierarchical}}%
        {MicroRTS}{\citep{huang2021gym}}%
        {Pommerman}{\citep{DBLP:journals/corr/abs-1809-07124}}%
        {DinerDash}{\citep{chen2020dinerdash}}%
        {Half Field Offense}{\citep{hausknecht2016half}}%
        {MuJoCo}{\citep{todorov2012mujoco,GymMuJoCoDocs}}%
        {Petting Zoo}{\citep{PettingZooDocs}}%
        {Nocturne}{\citep{nocturne2022}}%
        {RecoGym}{\citep{rohde2018recogym}}%
        {RecSim}{\citep{ie2019recsim}}%
        {RL4RS}{\citep{2021RL4RS}}%
        {Zork}{\citep{zahavy2019learn}}%
        {Dou Dizhu}{\citep{pmlr-v139-zha21a}}%
        {SMAC}{\citep{samvelyan19smac,ellis2022smacv2}}%
    }[#1]
}

\newcommand{\convert}[1]{%
    \IfEqCase{#1}{%
        {Y}{\checkmark}%
        {N}{\xmark}%
        {C}{$\nearrow$}%
    }[{\rotatebox[origin=c]{90}{\ \textbf{#1}~\citeenv{#1}\ }}]
}%
\newcommand{\pconvert}[1]{%
    \IfEqCase{#1}{%
        {Name}{\textbf{Name}}%
    }[#1] 
}%

\theoremstyle{thmstyleone}%
\theoremstyle{thmstyletwo}%

\theoremstyle{thmstylethree}%

\begin{document}

\title[\papertitle]{\papertitle}


\author*[1]{\fnm{Gregory} \sur{Palmer}}\email{gregory.palmer@baesystems.com}

\author[1]{\fnm{Chris} \sur{Parry}}

\author[1]{\fnm{Daniel} \sur{Harrold}}

\author[1]{\fnm{Chris} \sur{Willis}}



\affil[1]{\orgname{BAE Systems} \orgdiv{Applied Intelligence Labs}, Chelmsford Office \& Technology Park, Great Baddow, Chelmsford, Essex, CM2 8HN}


\abstract{
The rapid increase in the number of cyber-attacks in recent years
raises the need for principled methods for defending networks against
malicious actors.
\emph{Deep reinforcement learning} (DRL) has emerged as a  
promising approach for mitigating these attacks.
However, while DRL has shown much potential for cyber defence, 
numerous challenges must be overcome before DRL can be applied 
to the autonomous cyber defence (\aco) problem \emph{at scale}.
Principled methods are required for environments
that confront learners with \emph{very} high-dimensional state spaces, 
large multi-discrete action spaces, and adversarial learning.
Recent works have reported success in solving these problems individually. 
There have also been impressive engineering efforts towards solving all three
for real-time strategy games.
However, applying DRL to the full \aco problem remains an open challenge.  
Here, we survey the relevant DRL literature 
and conceptualize an idealised \aco-DRL agent.
We provide: 
i.) A summary of the domain properties that define the \aco problem;
ii.) A comprehensive comparison of \update{current \aco environments} used 
for benchmarking DRL approaches;
iii.) An overview of state-of-the-art approaches for scaling DRL to domains
that confront learners with the \cod, and;
iv.) A survey and critique of current methods for limiting the exploitability
of agents within adversarial settings from the perspective of \aco.
We conclude with open research questions that we hope will 
motivate future directions for researchers and practitioners 
working on \aco.

}




\maketitle

\section{Introduction} \label{sec:introduction}
 
\begin{figure}[h]
\centering
\resizebox{0.35\columnwidth}{!}{
\centering
\includegraphics[width=\columnwidth]{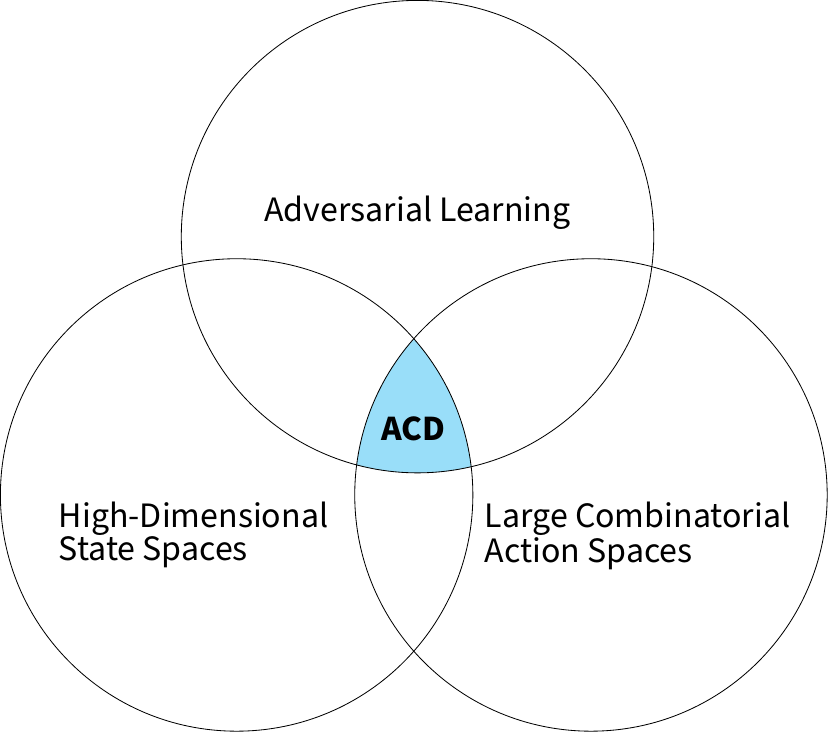}}
\caption{Three challenges that an idealised DRL-\aco agent must conquer.}
\label{fig:aco_requirements}
\end{figure}

The rapid increase in the number of cyber-attacks in recent years has 
raised the need for responsive, adaptive, and scalable 
\emph{autonomous cyber defence} (\aco) solutions~\citep{9596578,albahar2019cyber,theron152}. 
Adaptive solutions are desirable due to cyber-criminals increasingly
showing an ability to evade conventional security systems,
which often lack the ability to detect new types of attacks~\citep{9277523}.
The \aco problem can be formulated as an 
adversarial game involving a Blue agent tasked with defending
cyber resources from a Red attacker~\citep{baillie2020cyborg}.
%
%
%
%
Deep reinforcement learning (DRL) has been
identified as a suitable machine learning (ML) paradigm to apply 
to \aco~\citep{adawadkar2022cyber,li2019reinforcement,liu2020deep}.
However, current \say{out of the box} DRL solutions do not 
scale well to many real world scenarios.
This is primarily due to \aco lying at the intersection of three open problem 
areas for DRL, namely:
i.) The efficient processing and exploration of vast
high-dimensional state spaces~\citep{abel2016exploratory};
ii.) Large combinatorial action spaces, 
and;
iii.)~Minimizing the exploitability of DRL agents in adversarial games~\citep{gleave2019adversarial} (see \autoref{fig:aco_requirements}).  

%
%
%
The DRL literature features a plethora of efforts
addressing the above challenges individually. 
Here, we survey these efforts and define 
an idealised DRL agent for \aco.
%
%
This \report provides an overview of the current state of the field, 
\update{identifies methods that have the potential to fill current
gaps within the \aco-DRL literature},
defines long term objectives,  
and poses research questions
for DRL practitioners and \aco researchers to dig their teeth into.

Our contributions can be summarised as follows:

\spacer
\noindent{\textbf{1.)}} To enable an extensive evaluation of future DRL-\aco 
%
approaches, we provide an overview of \aco benchmarking 
environments.
\update{In order to evaluate DRL approaches for \aco,
benchmarking environments are required that confront
learners with the same properties as the full 
\aco problem.
To address this challenge, we outline desirable criteria 
to assess the suitability of ACD-gyms, and conduct 
a critical evaluation of \aco benchmarking environments (\autoref{sec:envs}).}   

%
%
%
%

\spacer
\noindent{\textbf{2.)}} We identify suitable
methods for addressing the \cod for \aco (\autoref{sec:hdss_approaches}). 
This includes a summary of approaches for
state-abstraction (\autoref{sec:state_abstraction}), 
efficient exploration (\autoref{sec:exploration}) and mitigating
catastrophic forgetting (\autoref{sec:knowledge_retention}), as well as a critical evaluation 
of high-dimensional action approaches (\autoref{sec:hdas_approaches}).

\spacer
\noindent{\textbf{3.)}} We formally define the \aco problem from 
the perspective of adversarial learning (\autoref{sec:adv_learning}).
Even within \say{simple} adversarial games, finding (near) optimal policies 
is non-trivial. 
%
%
We therefore review principled methods
for limiting exploitability (\autoref{sec:best_resposne_techniques}), 
and map out paths towards scaling these approaches to 
the full \aco challenge (\autoref{sec:scale_adv_learning}).

\spacer
%
%
\commentout{
This \report is structured as follows.
We begin with background and formal definitions,
%
%
then take a closer look at our target
domains, formally defining the properties
of \aco (\autoref{sec:envs}).
This allows us to evaluate
non-\aco environments; 
approaches for dealing with high-dimensional state spaces (\autoref{sec:hdss_approaches});
and action spaces (\autoref{sec:hdas_approaches}); and,
adversarial learning approaches (\autoref{sec:adv_learning}). 
We conclude with a discussion and a list of
open research questions (\autoref{sec:challenges}).}

\section{\update{Related Surveys}} \label{sec:related_work}

A number of surveys have been conducted in recent years that provide an overview of
 the different types of cyber attacks (\eg intrusion, spam, and malware) and the ML 
methodologies that have been applied in response~\citep{li2018cyber,liu2019machine,reda2021taxonomy,9277523,thiyagarajan2020review}. 
Given that ML methods themselves are susceptible to adversarial 
attacks, there have also been efforts towards assessing the risk
posed by adversarial learning techniques for cyber security~\citep{duddu2018survey,rosenberg2021adversarial}.
However, while these works evaluate existing threats to ML models in general 
(\eg white-box attacks~\citep{moosavi2016deepfool} and model poisoning attacks~\citep{kloft2010online}),
our \report focuses on the adversarial learning process for DRL agents within \aco, desirable
solution concepts, and a critical evaluation of existing techniques towards limiting
exploitability.
%

\cite{9596578} surveyed the DRL for cyber security literature,
providing an overview of works where DRL-based security methods 
were applied to cyber–physical systems, autonomous intrusion detection techniques, and 
multiagent DRL-based game theory simulations for defence strategies against
cyber-attacks.
\update{A recent survey by \cite{vyas2023automated}, meanwhile, provides a detailed 
overview of recent DRL approaches for \aco, 
a requirements analysis, and an overview of cyber-defence gym environments, 
along with algorithms provided within open-sourced ACD gyms.}

\update{In contrast to the above surveys, 
our work focuses on generation after next solutions. 
We capture the challenges posed by the 
\aco problem at scale and survey the DRL literature for suitable methods designed
to address these challenges separately, providing the building blocks for an
 idealised \aco-DRL agent. 
Therefore, one of the objectives of our work is to identify methods 
that to-date have not been applied to cyber-defence environments,
but have the potential to fill current gaps in the DRL for \aco literature.
We provide summary of surveys on ML for \aco in \autoref{tab:relatedsurvey}.}

%
%
%
%
%

\begin{table}[h]
\begin{center}
\resizebox{\columnwidth}{!}{
\begin{tabular}{||p{2.2cm}  | p{7cm} | c | c | c | c ||} 
 \hline
 Work & Summary & Envs & HDO & HDA & ADV  \\ [0.5ex] 
 \hline\hline
 \cite{nguyen2021deep} & 
Contains an overview of literature using standard DRL approaches for defending cyber–physical systems,
and considers game theoretic evaluations of defence strategies. 
& \xmark & \xmark & \xmark & \checkmark  \\ [0.5ex] 
 \hline
\cite{adawadkar2022cyber} & 
Focuses on DRL techniques for Intrusion Detection Systems (IDS), 
Intrusion Prevention Systems (IPS), 
and Identity and Access Management (IAM). &
\xmark & \checkmark & \xmark & \xmark  \\ [0.5ex] 
\hline
\cite{sharmila2023comprehensive} & 
A survey on cyber-defence and digital forensics, 
which includes a discussion on ML and data mining techniques. 
 & \xmark & \xmark & \xmark & \xmark  \\ [0.5ex]
 \hline
\cite{duddu2018survey} & Provides an overview of the vulnerabilities of ML approaches towards
adversarial attacks within the context of cyber-defence, along with defence strategies. 
& \xmark & \xmark & \xmark & \checkmark  \\ [0.5ex]
 \hline
\cite{ozkan2024comprehensive} &
Covers state-of-the-art (SOTA) ML techniques for \aco,
including ChatGPT-like AI tools for \aco.
A summary of common DRL approaches is also provided.
& \xmark & \checkmark & \xmark & \checkmark  \\ [0.5ex]
\hline
\cite{cengiz2023reinforcement} &
Reviewed the literature on DRL's penetration testing and
approaches for addressing different types
of attacks, including Denial of Services (DoS) and Distributed 
Denial of Services (DDoS), Spoofing, Jamming Attacks and SQL injections.
& \xmark & \checkmark & \checkmark & \checkmark  \\ [0.5ex]
 \hline
\cite{9277523} & 
Focuses on the challenges of intrusion, spam detection, and malware detection,
and provides a brief descriptions of each ML method, 
frequently used security datasets, essential ML tools.
The survey also provides a detailed summary of the risks faced
by ML models toward adversarial attacks. 
& \xmark & \xmark & \xmark & \checkmark  \\ [0.5ex] 
\hline
\cite{li2018cyber} & 
Provides a summary on using ML to combat cyber attacks, and
subsequently analyzes adversarial attacks on AI models,
their characteristics, and provide an overview of corresponding defence methods. 
 & \xmark & \xmark & \xmark & \checkmark  \\ [0.5ex] 
\hline
\cite{reda2021taxonomy} & 
Provides a comprehensive review of ML based defence countermeasures
for the False Data Injection attacks on Smart Grid infrastructure.
& \xmark & \xmark & \xmark & \checkmark  \\ [0.5ex] 
\hline
\cite{liu2019machine} &  
Proposes a taxonomy of supervised and unsupervised ML based Intrusion Detection Systems, 
and a summary of metrics, and benchmark datasets are introduced. 
& \xmark & \xmark & \xmark & \checkmark \\ [0.5ex] 
\hline
\cite{vyas2023automated} & Provides a detailed survey on current DRL approaches for cyber-defence.
The authors conducted a requirements analysis through comparing 40 papers.
One of the main takeaways was that the development of ACD Gyms 
and automated cyber-defence and attack solutions currently 
comprise separate research and development strands. 
The authors note that progress of each area is heavily dependent
on the others, therefore requiring a joint effort. 
& \checkmark & \checkmark & \checkmark & \checkmark \\ [0.5ex] 
\hline
\end{tabular}}
\end{center}
\caption{An overview of current \aco surveys and topics covered. 
We contrast existing works with respect to the focus areas 
of our survey, including an up-to-date overview of current \aco environments (Envs), 
methods relevant for tackling \aco's variable-sized and intractable state space (HDO),
high-dimensional action space approaches for \aco (HDA), 
and the adversarial learning (ADV) challenge.}
\label{tab:relatedsurvey}
\end{table}
\clearpage

\section{Background \& Definitions} \label{sec:background}

\begin{figure}[h]
\centering     
\subfloat[MDP]{\label{fig:MDP}\includegraphics[width=0.24\columnwidth]{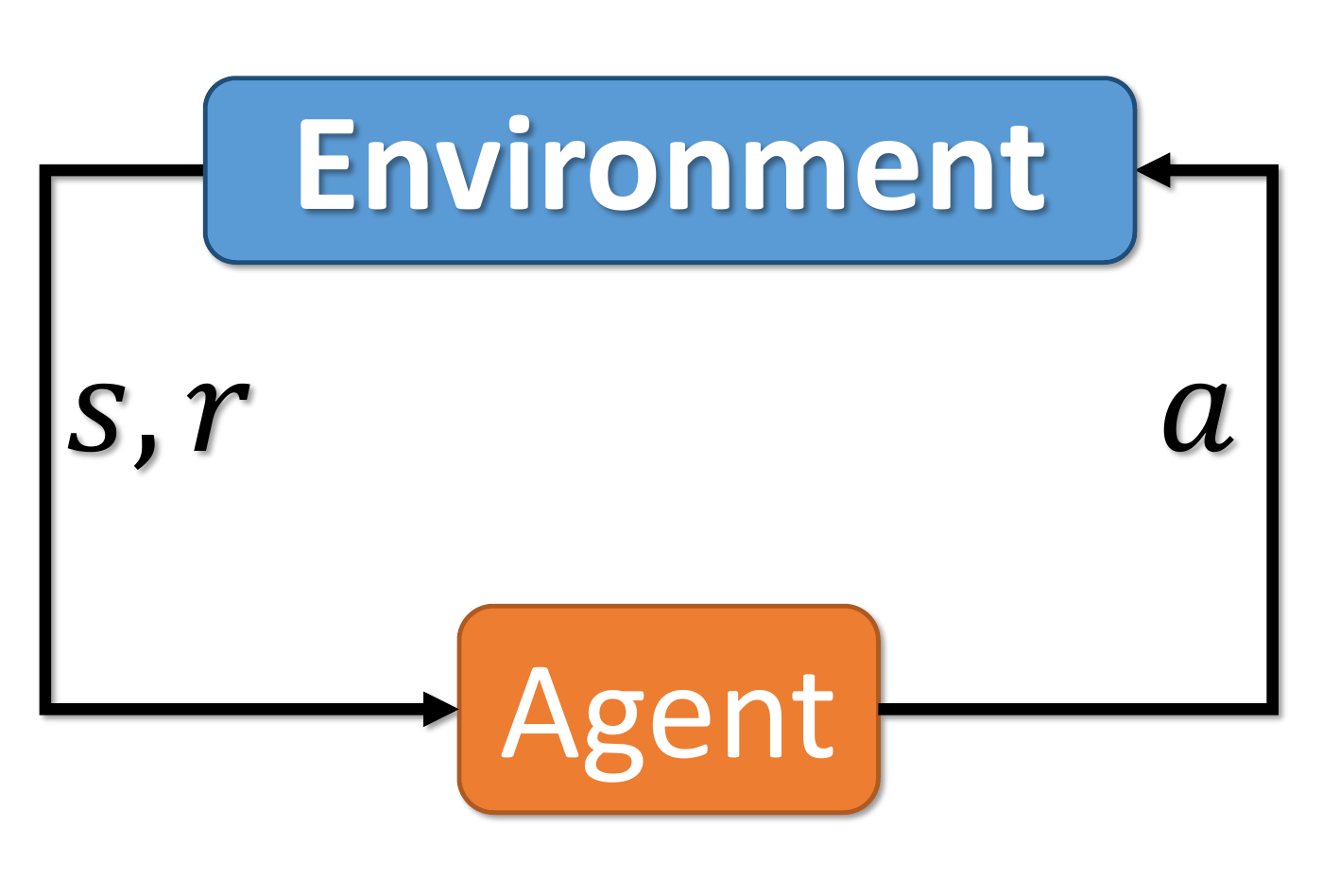}}
\subfloat[POMDP]{\label{fig:POMDP}\includegraphics[width=0.24\columnwidth]{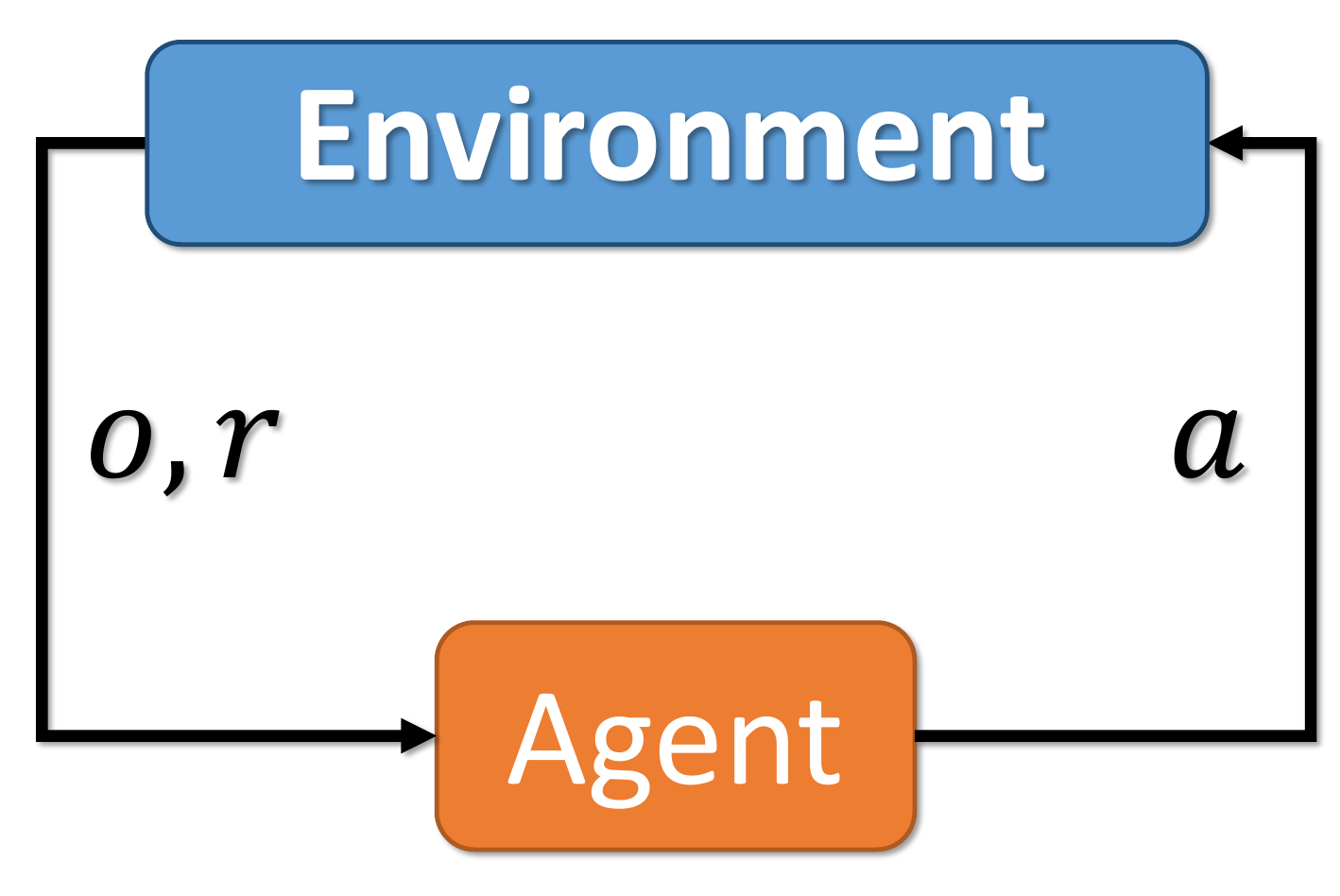}}
\subfloat[Markov Game (MG)]{\label{fig:MG}\includegraphics[width=0.24\columnwidth]{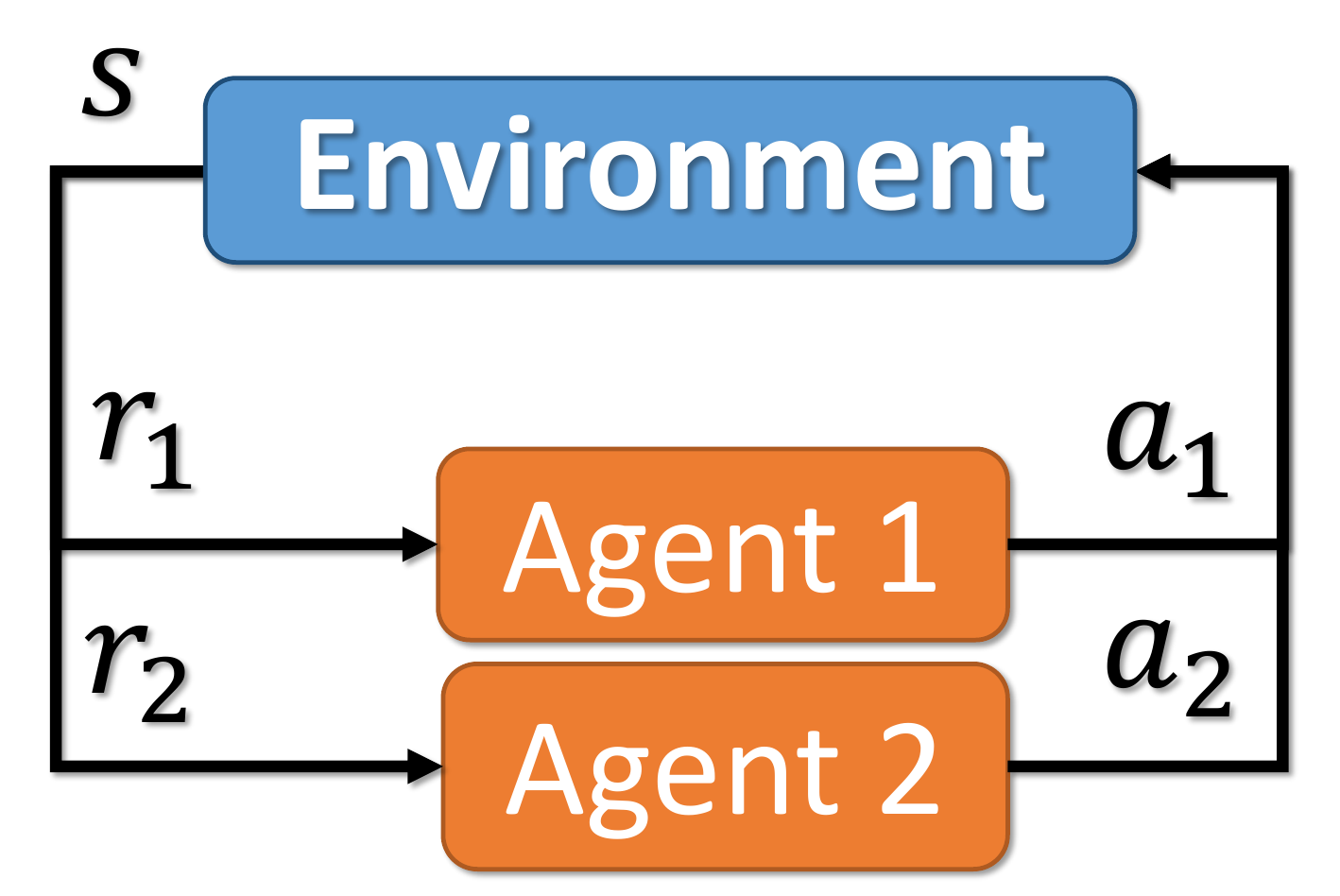}}
\subfloat[POMG]{\label{fig:POMG}\includegraphics[width=0.24\columnwidth]{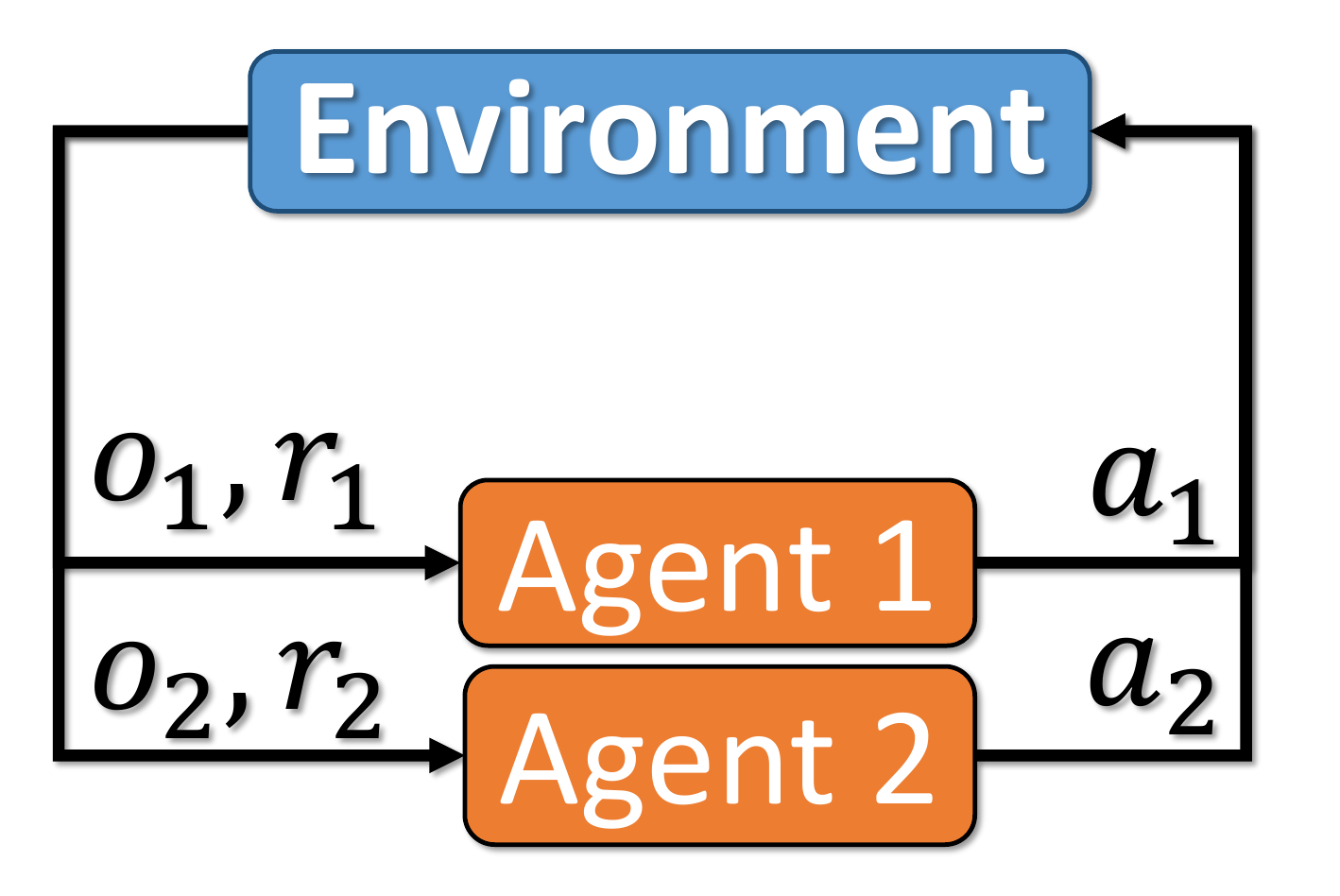}}

\subfloat[Dec-POMDP]{\label{fig:DecPOMDP}\includegraphics[width=0.32\columnwidth]{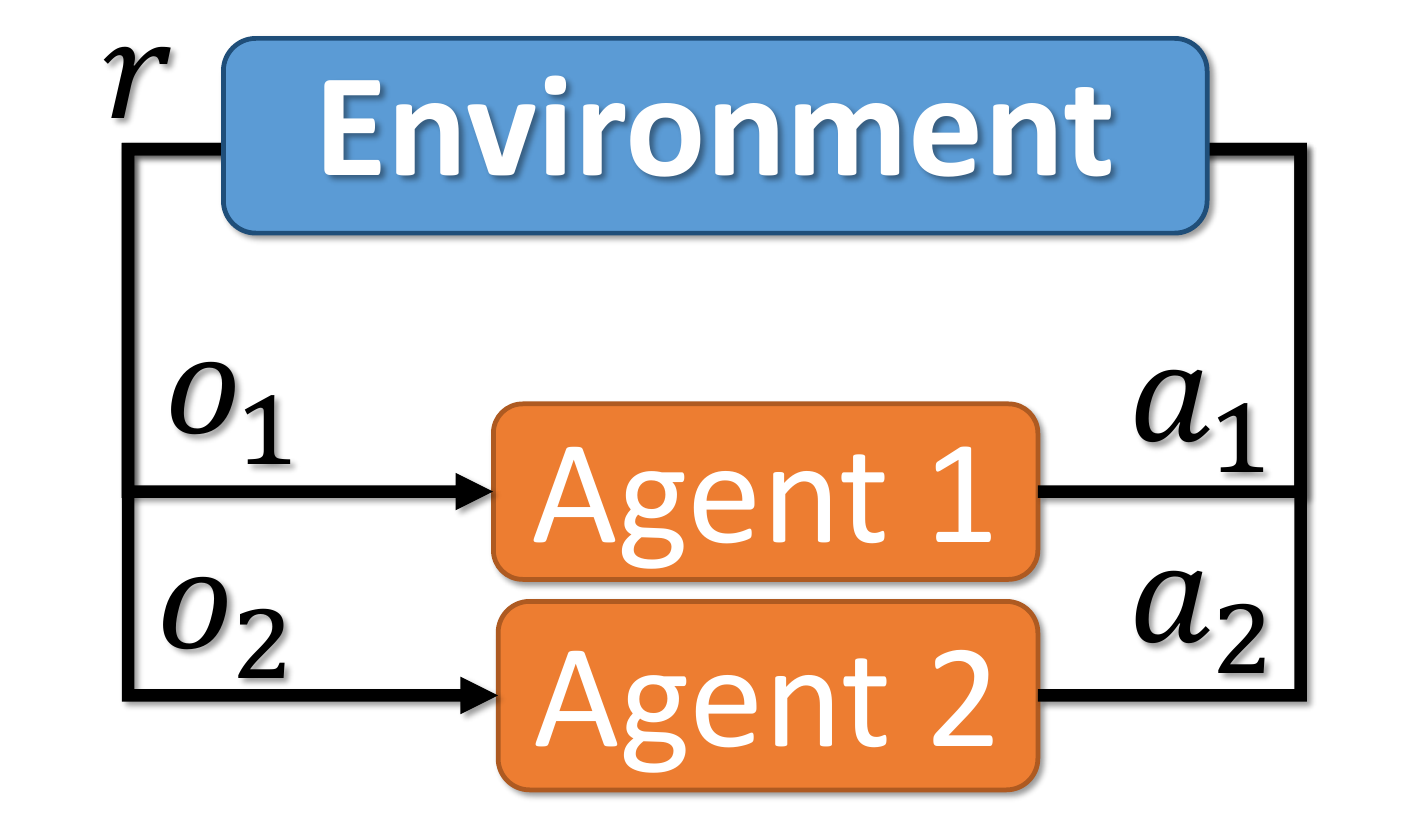}}
\subfloat[Slate-MDP]{\label{fig:SlateMDPs}\includegraphics[width=0.32\columnwidth]{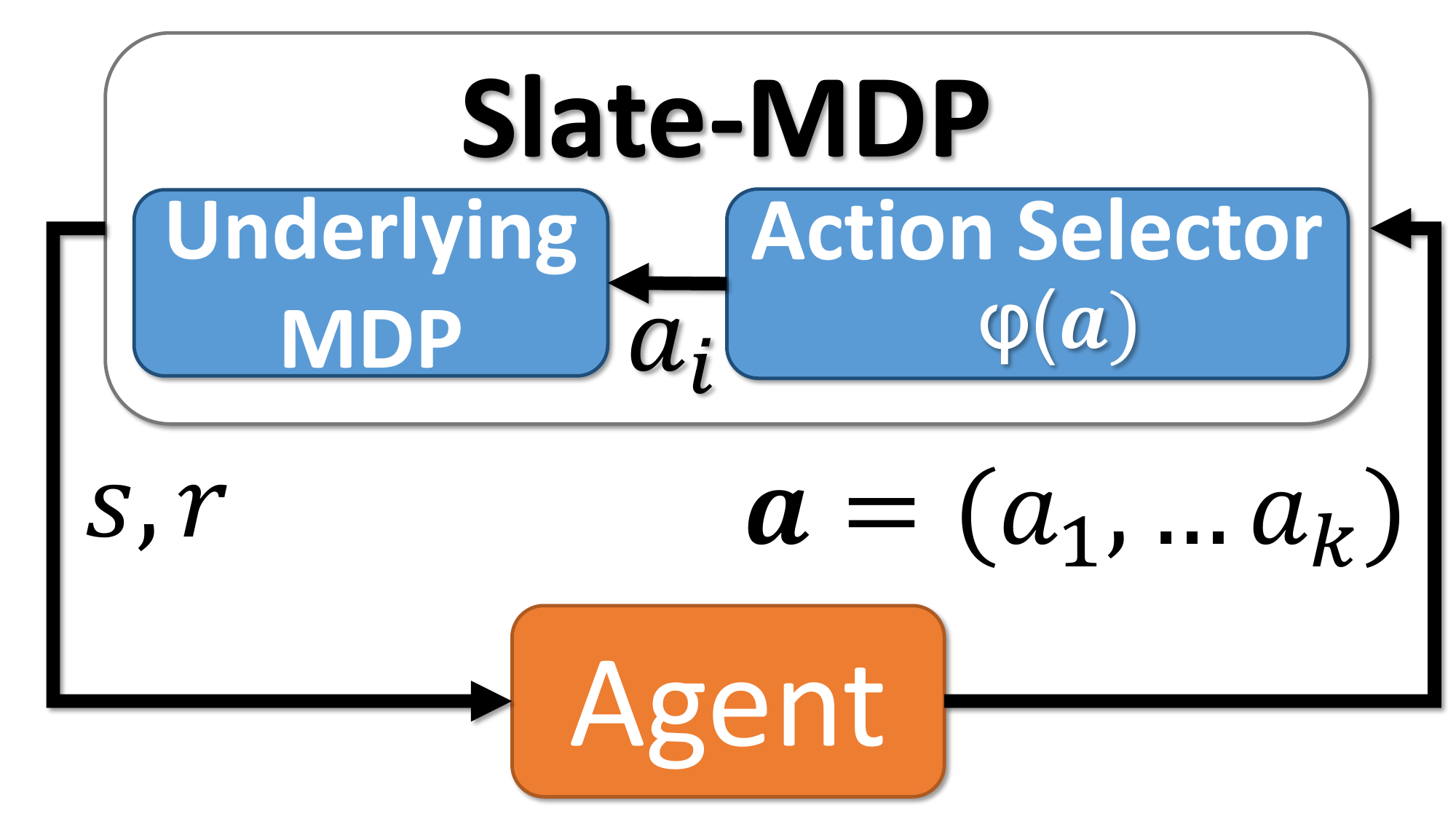}}
\subfloat[Parameterized Action MDP]{\label{fig:PAMDP}\includegraphics[width=0.32\columnwidth]{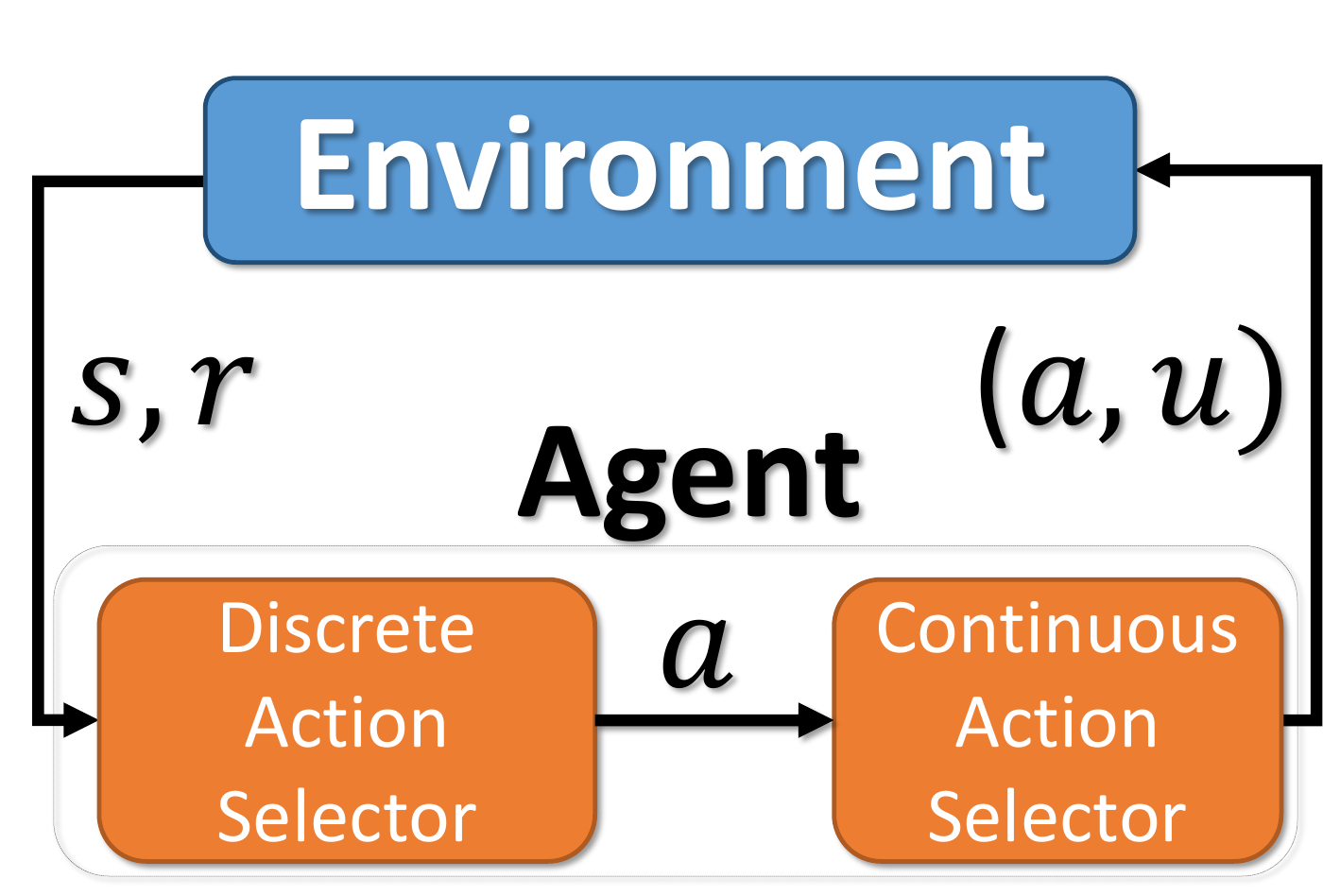}}
\caption{An overview of the problem formulations discussed in this \report. 
Within these formulations we have the following variables: states $s$, rewards $r$, actions $a$, and observations $o$.
For the Parameterized Action MDP we differentiate between discrete actions $a$ and continuous actions $u$.}
\label{fig:rl_problems_overview}
\end{figure}

Below we provide the definitions
and notations that we will rely on throughout this  
\report.
First, we will formally define the different types of 
models through which the interactions between \RL agents and
\update{cyber-defence} environments can be described.
\update{While these formulations are not specific to \aco, 
and similar introductions can be found in existing textbooks,
we consider the extent to which each model can be used to 
capture variations of the cyber-defence challenge.}
We will encounter each type in this 
survey (See \autoref{fig:rl_problems_overview} for an overview).
%
%
%
%

\subsection{(Partially Observable) Markov Decision Process} \label{sec:mdp}

Markov Decision Processes (MDPs) describe a class of problems
-- fully observable environments -- that defines the field of 
\RL, providing a suitable model to formulate interactions 
between reinforcement learners and their environment~\citep{sutton2018reinforcement}. 
Formally:
An MDP is a tuple 
$\M = \langle \states, \actions, \rewards, \transition \rangle$, where: 
$\states$ is a finite set of states; 
for each state $\state \in \states$ there exists a finite set of possible actions $\actions$; 
$\rewards$ is a real-valued payoff function 
$\rewards : \states \times \actions \times \states' \rightarrow \rats$, 
where $\rewards_\action(\state, \state')$ is the expected payoff following 
a state transition from $\state$ to $\state'$ using action $\action$; 
$\transition$ is a state transition probability matrix 
$\transition : \states \times \actions \times \states' \rightarrow [0, 1]$, 
where $\transition_{\action}(\state, \state')$ is the probability of state $\state$ transitioning into state $\state'$ using action $\action$. 
In addition, MDPs can have terminal (absorbing) states at which the episode ends.
Numerous environments lack the full 
observability property of MDPs~\citep{oliehoek2015concise}, 
\update{which is also a reasonable assumption made for both cyber-attacking~\citep{horta2024evaluating},
and cyber-defence agents~\citep{elhami2022control}.}
Here, a Partially Observable MDP (POMDP) extends an MDP $\M$ by adding 
$\langle \observations, \observationf \rangle$, where: 
$\observations$ is a finite set of observations;
and $\observationf$ is an observation function defined as
$\observationf : \states \times \actions \times \observations \rightarrow [0, 1]$, where $\observationf(\observation \rvert \state, \action)$
is a distribution over observations $\observation$ that may occur in state $\state$ after taking action $\action$.

\subsection{(Partially Observable) Markov Games} \label{sec:partially_observable_markov_games}

\update{
The above formulations fail to explicitly account for the presence of \red (or other) 
agents that interact with the environment through receiving observations and taking actions.
Many environments, including the cyber-defence scenarios that are the focus of this \report, 
feature more than one agent.} 
Here, game theory offers a solution via Markov games 
(also known as stochastic games~\citep{shapley1953stochastic}). 
%
%
\reportonly{\begin{definitionsec}[Markov Games]}
A Markov game is defined as a tuple $(n, \states, \actions, \transition, \rewards)$, 
that has a finite state space $\states$;
for each state $\state \in \states$ a joint action space $(\actions_1 \times ... \times \actions_n)$, 
with $\actions_p$ being the number of actions available to player $p$;
a state transition function \update{$\transition : \states \times \actions_1 \times ... \times \actions_n \times \states' \rightarrow [0,1]$}, 
returning the probability of transitioning from a state \update{$\state$} to \update{$\state'$} given an action profile $\action_1 \times ... \times \action_n$;
and for each player $p$ a reward function: $\mathcal{\rewards}_p : \update{\states} \times \actions_1 \times ... \times \actions_n \times \update{\states'} \rightarrow \rats$ \citep{shapley1953stochastic}. 
We allow \emph{terminal states} at which the game ends. 
Each state is fully-observable. 
\reportonly{\end{definitionsec}}
%
%
%
%
%
A Partially Observable Markov Game (POMG) is an extension of Markov Games that includes 
$\langle \observations, \observationf \rangle$ 
a set of observations $\update{\observation \in}\ \observations$; 
and an observation probability function defined as
$\observationf_p : \states \times \actions_1 \times ... \times \actions_n \times \observations \rightarrow [0,1]$. 
For each player $p$ the observation probability function $\observationf_p$ is a distribution over 
observations $\observation$ that may occur in state $\state$, given an action profile $\action_1 \times ... \times \action_n$.
%
%
POMGs where, at each step, all $n$ agents receive an identical reward
are known as Decentralized-POMDP (\decPOMDP)~\citep{oliehoek2015concise}.
\update{This formulation is applicable to decentralized cyber-defence solutions
with perfectly aligned objectives.}

\subsection{Slate Markov Decision Processes} \label{sec:slateMDPs}

%
\update{
Some problem formulations, such as \emph{human-machine teaming} scenarios, can require a custom model. 
For instance, in security operation centres cyber-security personnel are often confronted
with vast amounts of data, reducing the likelihood of a timely decision being made for
mitigating an attack.
Here, recommender systems can provide a solution, taking  on the responsibility of \emph{data triage automation}
and recommending suitable cybersecurity response and mitigation measures~\citep{lyons2014recommender,pawlicka2021systematic}.
The benefit of this approach is that human operators can remain in the loop 
for tasks where full automation is not currently feasible or permissible~\citep{pawlicka2021systematic}.
Formally, this type of problem can be defined as a Slate-MDP~\citep{sunehag2015deep}.}
%
%
%
%
%
%
%
%
\reportonly{\begin{definitionsec}[Slate-MDPs]}
Given an underlying MDP $\MDP = \langle \states, \actions, \transition, \rewards  \rangle$, 
a Slate-MDP is a tuple $\langle \states, \actions^{l}, \transition', \rewards'  \rangle$.
Within this formulation $\actions^l$ is a finite discrete action space 
$\actions^l = \{\bm{\action}_1, \bm{\action}_2, ..., \bm{\action}_N\}$,
representing the set of all possible slates to recommend given the current state $\state$. 
Each slate can be formulated as $\bm{\action}_i = \{\action_i^1, \action_i^2,... \action_i^K\}$, with $K$ representing the
size of the slate.  
Slate-MDPs assume an action selection function $\varphi : \states \times \actions^l \rightarrow \actions$.
State transitions and rewards are, as a result, determined via functions
$\transition' : \states \times \actions^l \times \states' \rightarrow [0, 1]$
and
$\rewards' : \states \times \actions^l \times \states' \rightarrow \rats$
respectively. 
Therefore, given an underlying MDP $\MDP$, we have 
$\transition'(\state, \bm{\action}, \state') = \transition(\state, \varphi(\bm{\action}), \state')$
and
$\rewards'(\state, \bm{\action}, \state') = \rewards(\state, \varphi(\bm{\action}), \state')$.
Finally, there is an assumption that the most recently executed action
can be derived from a state via a function $\psi : \states \rightarrow \actions$.
Note, there is no requirement that $\psi(\state_{t+1}) \in \bm{\action}_t$, therefore, the action
selected can also be outside the provided slate~\footnote{There are environments that treat 
$\psi(\state_{t+1}) \notin \bm{\action}_t$ as a failure property, 
upon which an episode terminates~\citep{sunehag2015deep}.}. 
\reportonly{\end{definitionsec}}
%

\subsection{Parameterized Action MDPs} \label{sec:autoref}

Parameterized Action MDPs (PA-MDPs) are a generalization of MDPs
where the agent must choose from a discrete set of
parameterized actions~\citep{wei2018hierarchical,masson2016reinforcement}.
More formally, PA-MDPs assume a finite discrete set of actions 
$\actions_d = \{\action_1, \action_2, ..., \action_n\}$ and 
for each action $\action \in \actions_d$ a set of continuous 
parameters $\continuousaction_\action \subseteq \rats^{m_\action}$, where $m_\action$
represents the dimensionality of action $\action$.
Therefore, an action is a tuple $(\action, \continuousaction)$ 
in the joint action space,
%
$\actions = \bigcup_{\action \in \actions_d} \{(\action, \continuousaction) \rvert \continuousactions_\action\}$.
%
\update{An \aco example of a PA-MDP could consist of timed actions, 
such as defining a start and end time during which an update
to a firewall's access control list should be applicable.}
 

\subsection{Types of Action Spaces} \label{sec:action_spaces}

From the above definitions we see that environments
have different requirements \wrt their action spaces~\citep{9231687}.
%
%
\update{Below we provide an overview of different types of action spaces, 
along with their suitability for common cyber-defence scenarios. 
As a running example, we consider the action spaces in relation to an \aco agent in 
charge of managing a system-wide firewall via an access control list (ACL).
When adding or removing a rule from this list, the agent is required 
to define an action signature consisting of~\citep{PrimAITE}:
\begin{enumerate}
\item Action (do nothing, create rule, delete rule);
\item Permission (DENY, ALLOW);
\item Source IP Address;
\item Destination IP Address;
\item Protocol, and;
\item Port.
\end{enumerate}} 

\spacer
\noindent \textbf{Discrete} actions \update{consist of a set} $\action \in \{\update{1, 2,} ... N\}$, with $N \in \N$ available actions in a given state. 
\update{As we shall discuss throughout this paper, 
traditional DRL approaches struggle when a discrete 
action space has a large number of actions.
For cyber-defence in particular, there is a combinatorial increase in the number 
of possible actions with respect to the number of action dimensions. 
Therefore, a discrete action space can quickly become intractable for a cyber-defence 
agent tasked with defending a large network.
In-fact, even on a relatively small network 
the ACL action space defined 
above can consist of thousands of discrete actions.}


\spacer
\noindent \textbf{MultiDiscrete} actions \update{are defined by an action vector $\bm{\action}$, each $\action_i$ is a discrete action with $N$ possibilities}. 
\update{
The benefit of this formulation is that it results in a linear increase in the number 
of policy outputs with respect to the number of degrees of freedom, and provides a level of independence 
for each individual action dimension~\citep{tavakoli2018action}. 
Therefore, in relation to our ACL example,
our action vector would contain six elements defining an action signature. 
However, while a combinatorial action space does drastically reduce the number of 
individual action components, invalid actions may frequently be sampled, as the selected
action vector may contain invalid parameters along a subset of axes $\action_i$~\citep{huang2022closer}, 
e.g., deleting a rule that does not exist within the ACL.
Furthermore, the challenge of exploring a more extensive set of action combinations remains~\citep{10327797}}.

\spacer
\noindent \update{\textbf{Continuous} actions, $\action \in \rats\update{^K}$, consist of action vectors of $K$, potentially bounded, real numbered actions.}
\update{At a first glance a continuous action space does not appear to be a good fit for \aco problems, 
such as our ACL example, given that we clearly have a set of discrete actions
along each axis.
However, as we shall discuss in Section~\ref{sec:hdas_approaches:wolpertinger}, 
a continuous action space can be applicable for \aco 
when given a suitable approach for embedding discrete 
actions into a continuous space. 
}

\spacer
\noindent \textbf{Slate} actions \update{are a set of actions} $\bm{\action} = \{\action_1, \action_2, ..., \action_n\}$ 
from which one can be selected\update{, typically by a human in the loop.
Therefore, this formulation makes a lot of sense for human-machine teaming scenarios.
For example, we may wish to implement an ACD policy that presents a network administrator 
wit a set of access control rules upon observing a threat.
The network administrator is subsequently tasked with deciding which rule to apply, 
or to ignore the suggestions.} 

\spacer
\noindent \textbf{Parameterized} actions are mixed discrete-continuous actions, \eg a tuple $(\action, \continuousaction)$ where $\action$ is a discrete action, and $\continuousaction$ is a continuous action.
\update{For instance, as stated for PA-MDPs, the continuous actions can be used to define timespan during which an ACL rule needs to be applicable.}


\subsection{Reinforcement Learning} \label{sec:rl}

The goal of an \RL algorithm is to learn a policy $\pi$ 
that maps states to a probability distribution over
the actions 
$\pi : \states \rightarrow  P(\actions)$, 
so as to maximize the expected return 
$\E_\pi [\sum^{H-1}_{t=0} \gamma^t r_t ]$. 
Here, $H$ is the length of the horizon, and
$\gamma$ is a discount factor $\gamma \in \update{(}\zero, 1]$ 
weighting the value of future rewards.
Many of the approaches discussed in this \report use the Q-learning algorithm 
introduced by Watkins \citep{watkins1989learning,watkins1992q} as their foundation. 
%
%
Using a dynamic programming approach, the algorithm learns action-value estimates 
(Q-values) independent of the agent's current policy. 
Q-values are estimates of the discounted sum of future rewards (the return) that 
can be obtained at time $t$ through selecting an action $\action \in \actions$ in a 
state $\state_t$, providing the optimal policy is selected in each state that follows. 
Q-learning is an \emph{off-policy} temporal-difference (TD) learning algorithm. 

In environments with a low-dimensional state space Q-values can be maintained using a Q-table. 
%
Upon choosing an action $\action$ in state $\state$ according 
to a policy $\pi$, the Q-table is updated by bootstrapping the immediate reward $r$ 
received in state $\nextstate$ plus the discounted expected future reward from the next state, 
using a 
scalar $\alpha$ to control 
the learning rate: 
%
$Q_{k+1}(\state, \action) \gets Q_{k}(\state,\action) + 
\alpha \big(r + \gamma \max_{\update{\action \in \actions}} \left[Q_k\left(\nextstate, \action \right) 
- Q_k\left(\state, \action \right)\right]\big)$.
%
Many sequential decision problems 
have a high-dimensional state space. 
Here, Q-values can be approximated using a function approximator, 
for instance using 
a neural network. 
%
\commentout{The parameters $\theta$ of the function approximator can be learned via experiences gathered 
by the agent while exploring their environment, choosing an action $\action$ in state $\state$ 
according to a policy $\pi$, and updating the Q-function by bootstrapping the immediate reward $r$ 
received in state $\nextstate$, plus the expected future reward from the next state~\citep{mnih2015human,mnih2013playing}: 
%
\begin{align} 
\theta_{k+1} &= \theta_{k} - \dfrac{1}{2} \alpha \nabla \big(Y^{Q}_{k} - Q\left(\state, \action;\theta_k\right)\big)^2 \\
             &= \theta_{k} + \alpha \big(Y^{Q}_{k} - Q\left(\state, \action;\theta_k\right)\big) \nabla_{\theta_k}Q\left(\state, \action; \theta_k\right).
\end{align}
%
Here, $Y_{k}^{Q}$ is the bootstrap target which sums the immediate reward $r$ and the current estimate of the return obtainable from 
the next state $\nextstate$ assuming optimal behaviour, discounted by $\gamma \in \left(0, 1\right]$, 
We provide a recap of relevant DRL approaches 
in \autoref{sota_drl_approaches}.
\begin{equation} \label{eq:targetValue}
Y^{Q}_{k} \equiv r + \gamma \max_{\action \in \actions} Q\left(\nextstate, \action; \theta_k\right).
\end{equation}
The Q-value $Q\left(\state, \action; \theta_k\right)$ therefore moves towards the target by following the gradient $\nabla_{\theta_k}Q\left(\state, \action; \theta_k\right)$. 
}
\update{For a recap of popular DRL approaches that will be discussed throughout this \report, 
including variations of Deep Q-Network (DQN)~\citep{mnih2013playing,mnih2015human,hasselt2010double}
and Proximal Policy Optimization~\citep{schulman2017proximal}, we recommend 
the following surveys~\citep{wang2022deep,han2023survey,mousavi2018deep}.}

\subsection{Joint Policies}

Our domain of interest can feature multiple agents. 
%
%
For each agent $p$, the strategy $\pi_p$ represents a mapping from the 
state space to a probability distribution over actions: 
$\pi_p : \update{\observations_p} \rightarrow \Delta(\actions_p)$\update{, where
$\observations_p$ represents agent $p$’s observation space.}
%
Transitions within 
Markov games 
are determined by a joint policy.
%
The notation $\bm{\pi}$ refers to a joint policy of all agents. 
Joint policies excluding agent $p$ are defined as $\bm{\pi}_{-p}$. 
The notation $\langle \pi_p, \bm{\pi}_{-p}\rangle$ refers to a joint policy with 
agent $p$ following $\pi_p$ while the other agents follow $\bm{\pi}_{-p}$. 

\subsection{Multi-Agent Learning} \label{sec:background:marl_pathologies}

A popular approach towards scaling DRL to 
large combinatorial action spaces is to 
apply multi-agent deep reinforcement learning (MADRL).
\update{In this survey we will see this formulation 
featured in two contexts: i.) Environments
that require a decentralized cyber-defence solution (See \autoref{sec:envs}), and;
ii.) Centralized controllers using an \emph{action decomposition} approach that, 
in essence, decomposes an action into actions
provided by multiple agents (see \autoref{sec:action_decomposition}).
Both contexts require approaches designed to facilitate agent cooperation 
and coordination.}
Formally, given an expected gain, 
$\G{i}{\bm{\pi}} = \mathbb{E}_{\bm{\pi}} \{ \sum_{k=0}^\infty \gamma^k r_{i,t+k+1} \rvert x_t = x \}$,
the underlying policies must find a Pareto optimal solution,
\ie a joint policy $\bm{\hat{\pi}}$ from which no 
agent $i$ can deviate without making at least one other 
agent worse off~\citep{matignon2012independent}. 
%

\begin{figure}[h]
\subfloat[Independent Learners]{\label{fig:actions:IL}\includegraphics[width=0.32\columnwidth]{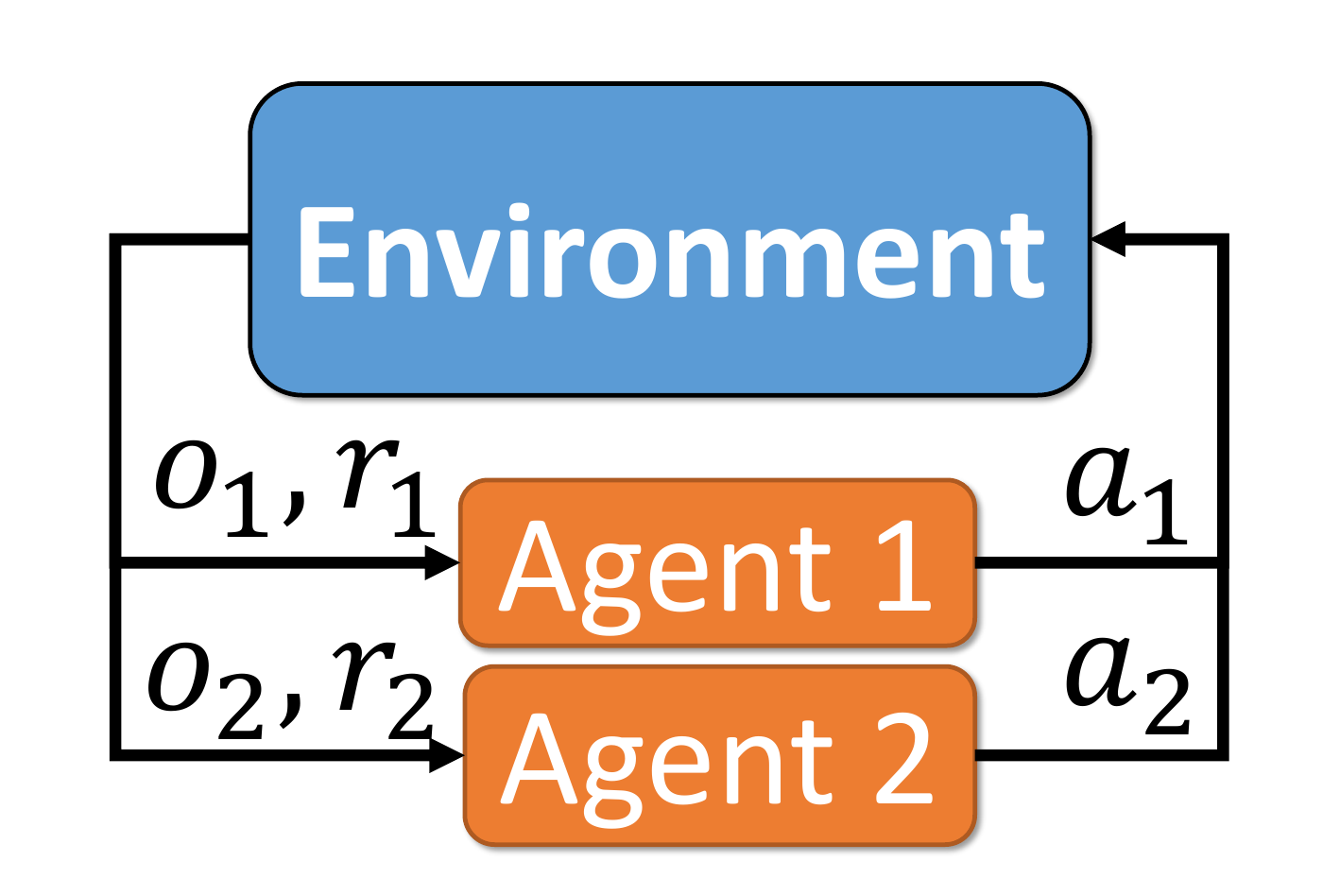}}
\subfloat[Centralized Training for Decenctralized Execution]{\label{fig:actions:CTDE}\includegraphics[width=0.32\columnwidth]{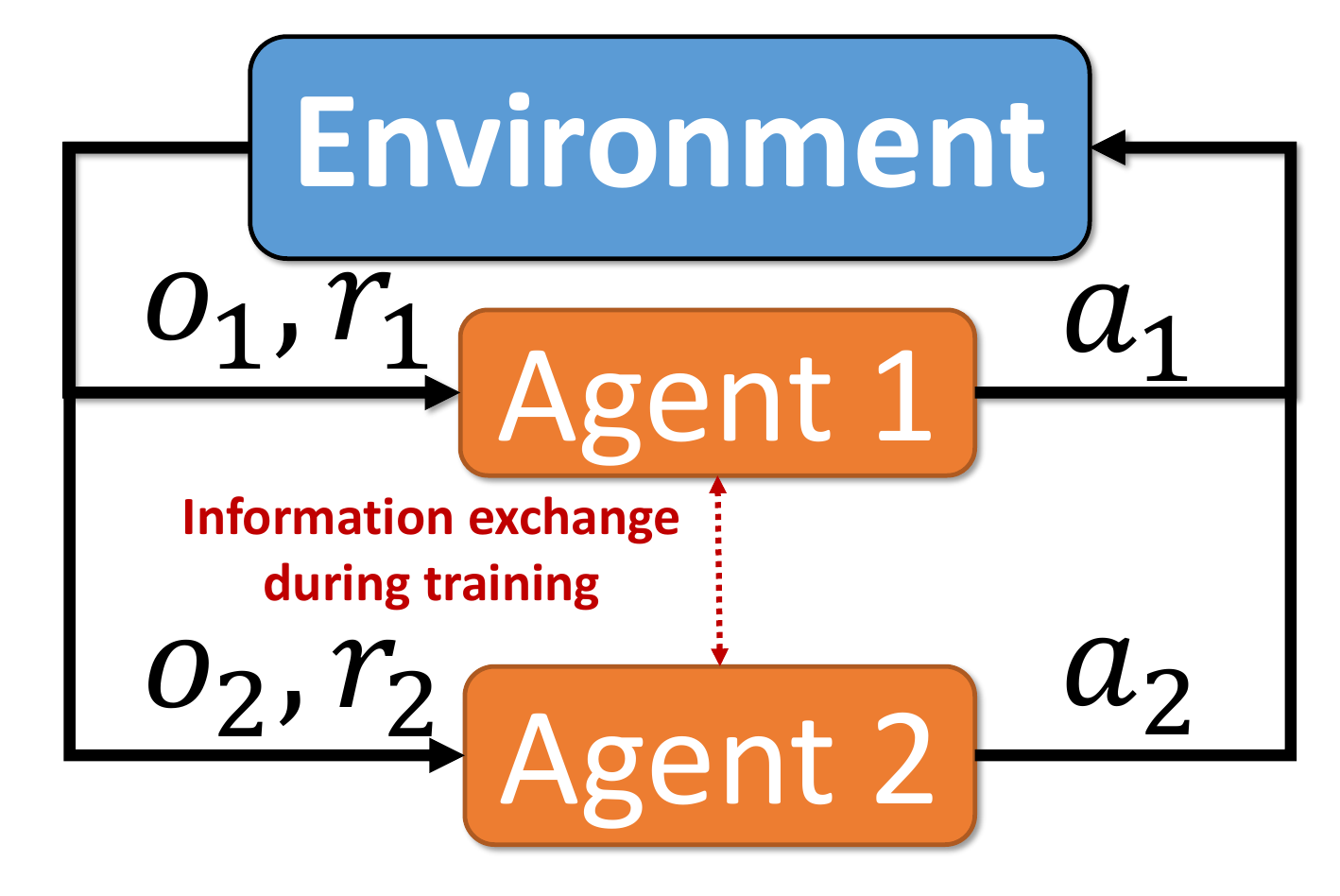}}
\subfloat[Centralized Controller]{\label{fig:actions:CC}\includegraphics[width=0.32\columnwidth]{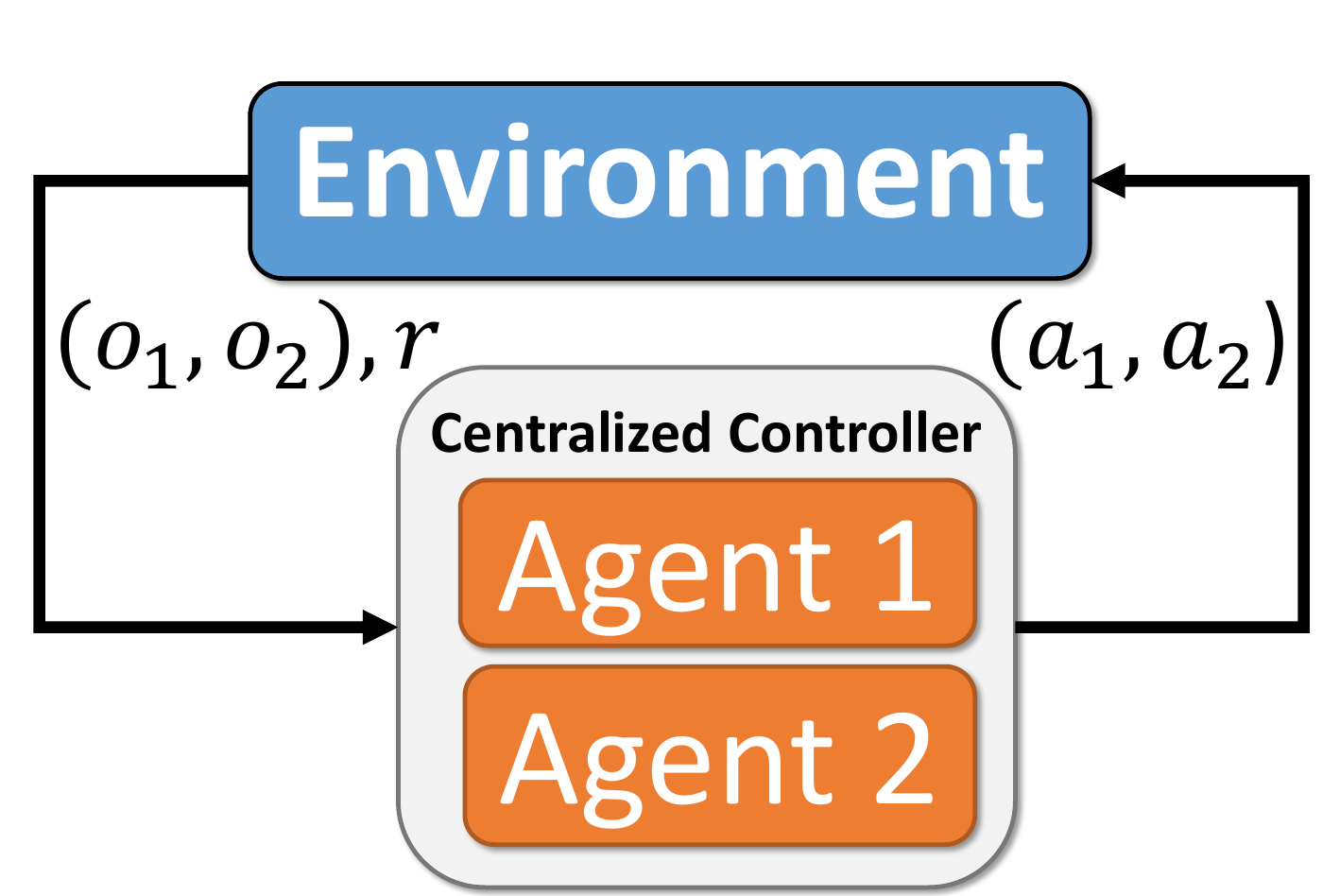}}
\caption{An overview of multi-agent reinforcement learning training schemes.}
\label{fig:marl_approaches_overview}
\end{figure}

There are three categories of training schemes 
for cooperative MA(D)RL (illustrated in \autoref{fig:marl_approaches_overview}): 
independent learners (ILs), who treat each other as part of the environment; 
the centralized controller approach, which does not scale with the number of agents; 
and centralized training for decentralized execution (CTDE).

Even within stateless two player matrix games 
with a small number of actions per agents,
ILs fail to consistently converge upon Pareto optimal solutions~\citep{busoniu2008comprehensive,nowe2012game,matignon2012independent,kapetanakis2002reinforcement,claus1998dynamics,matignon2007hysteretic,
panait2006lenient,panait2008theoretical}. 
\update{However, ILs are frequently used as a baseline for approaches discussed in this 
paper.} 
Therefore, to better understand the challenges that confront
action decomposition approaches
we shall first briefly consider the multi-agent learning pathologies 
that learners must overcome to converge upon a Pareto optimal joint-policy
from the perspective of ILs~\footnote{For a detailed recap 
please read~\citep{JMLR:v17:15-417,lauer2000algorithm,kapetanakis2002reinforcement,palmer2020independent}.}:
%

\spacer
\noindent \textbf{Miscoordination} occurs when there are two or more incompatible Pareto-optimal 
equilibria~\citep{claus1998dynamics,kapetanakis2002reinforcement,matignon2012independent}. 
One agent choosing an action from an incompatible equilibria is sufficient to lower the gain.
%
%
Formally:
%
two equilibria $\bm{\pi}$ and $\bm{\hat{\pi}}$ are incompatible \emph{iff} the gain received for pairing at 
least one agent using a policy $\pi$ with other agents using a policy $\hat{\pi}$ results in a lower gain compared 
to when all agents are using $\pi$:
$\exists i, \bm{\pi}_i \neq \bm{\hat{\pi}}_i, \mathcal{G}_{i, \langle \bm{\hat{\pi}}_i, \bm{\pi}_{-i} \rangle} < \mathcal{G}_{i, \bm{\pi}}$.

\spacer
\noindent \textbf{Relative Overgeneralization:} \label{sec:relative_overgeneralization}
ILs are prone to being drawn to sub-optimal but wide peaks in the reward space, 
as there is a greater likelihood of achieving collaboration there \citep{panait2006lenience}. 
Within these areas a sub-optimal policy yields a 
higher payoff on average when each selected action is paired with an arbitrary action chosen by 
the other agent \citep{panait2006lenience,wiegand2003analysis,palmer2020independent}. 
%
%

\spacer
\noindent \textbf{Stochasticity of Rewards and Transitions:}
Rewards and transitions can be stochastic, which has implications 
for approaches that use optimistic learning
to overcome the relative overgeneralization pathology
~\citep{palmer2020independent,palmer2018negative,palmer2018lenient}.

\spacer
\noindent \textbf{The Alter-Exploration Problem:} 
In MA(D)RL increasing the number of agents also increases \emph{global exploration},
the probability of at least one of $n$ agents exploring: $1 - (1 - \epsilon)^n$.
Here, each agent explores according to a probability $\epsilon$~\citep{matignon2012independent}. 
%

\spacer
\noindent \textbf{The Moving Target Problem} 
is a result of agents updating their policies in 
parallel~\citep{bowling2002multiagent,sutton1998introduction,tuyls2012multiagent,tuyls2007evolutionary}. 
This pathology is amplified when using experience replay memories $\mathcal{D}_i$, 
due to transitions becoming deprecated~\citep{foerster2017stabilising,omidshafiei2017deep,palmer2018lenient}.
%

\spacer
\noindent \textbf{Deception:} \label{sec:deception}
Deception occurs when utility values are calculated using rewards 
backed up from follow-on states from which pathologies such as miscoordination and relative 
overgeneralization can also be back-propagated \citep{JMLR:v17:15-417}. 
States with high local rewards can also represent a problem, drawing 
ILs away from optimal state-transition trajectories~\citep{JMLR:v17:15-417}.

\spacer

Non-trivial approaches are required in order to consistently converge upon a Pareto
optimal solution. 
Solutions to these challenges will be considered in more detail in Sections \ref{sec:hdas_approaches:marl_and_coordination} and \ref{sec:action_decomposition}.

\section{\update{Cyber-Defence Environments for DRL}} \label{sec:envs}

\update{To develop an understanding of how DRL can be used effectively 
for cyber-defence tasks, simulation and emulation environments
are required that allow researchers and practitioners 
to model realistic threats and network conditions~\citep{molina2021farland}. 
Publicly available cyber-security datasets, such as MITRE’s ATT\&CK framework~\citep{strom2018mitre},
have helped further advance ML approaches for cyber-defence~\citep{ahn2022malicious,al2019automating,al2020learning,ampel2021linking,bagui2022detecting,subbaratinam2022machine}.
However, in order to learn, 
DRL agents require interaction with a dynamic environment
that allows advanced autonomous decision making while providing 
feedback in the form of a reward signal~\citep{walter2021incorporating}.
This requirement has resulted in the development of a number 
of cyber-defence benchmarking environments for training
and evaluating DRL agents, which we shall refer to as \acogyms.}

\update{In the majority of \acogyms 
the Red agent is tasked with moving through the network graph, 
compromising nodes in order to progress. 
The end goal in most cases is to reach and impact a high-value target node.
In all \acogyms, the task of the Blue agent is conceptually identical -- to identify and counter the Red agent's
intrusion and advances using the available actions.} 

\update{In this section we critically evaluate six popular and topical \acogyms, 
including 
\update{the CybORG environment~\citep{cyborg_acd_2021,cage_cyborg_2022};
NASim~\citep{schwartz2019autonomous,janisch2023nasimemu};
CyberBattleSim (CBS)~\citep{10216719,shashkov2023adversarial,cyberbattlesim,walter2021incorporating};
FARLAND~\citep{molina2021farland};
YAWNING TITAN (YT)~\citep{YAWNING},
and; PrimAITE~\citep{PrimAITE}.} 
First we will provide a summary of each of these environments.
Next, we have identified desirable criteria 
to assess the suitability of \acogyms for developing and 
assessing techniques for tackling challenges in the ACD domain.
We shall compare and contrast the six \acogyms based on these 
criteria.
A final summary of our evaluation is provided in~\autoref{tab:aco_envs}, 
towards the end of the section.}


\subsection{\update{Cyber-Defence Environments Overview}}

\subsubsection{\update{Cyber Operations Research Gym (CybORG)}}

\update{The Cyber Operations Research
Gym (CybORG)~\citep{cyborg_acd_2021} 
is a framework for implementing a 
range of \aco scenarios.
It provides a common interface 
at two levels of fidelity in the 
form of a finite state machine
and an emulator. 
The latter provides for each node an operating system,
hardware architecture, users, and passwords~\citep{cage_cyborg_2022}.
CybORG allows for both the training of \blue and \red agents.
Therefore, in principle, it provides an \aco environment for 
adversarial learning. 
However, in practice further modifications are required in order
to support the loading of a DRL opponent.  
This is due to environment wrappers (for reducing the size 
of the observation space and mapping actions) only being available
to the agent that is being trained. 
Currently, the loaded opponent instances only receive
raw unprocessed dictionary observations. 
Furthermore, while CybORG does allow for customizable configuration files,
in practice we find that specifying a custom network is 
non-trivial. 
Nevertheless, thanks to the annual Cyber Autonomy Gym for Experimentation (CAGE) challenges, that
have been run by The Technical Cooperation Program (TTCP) since 2021, 
CybORG now features a diverse set of cyber defence problem scenarios.} 

\update{For \textbf{TTCP CAGE Challenges 1 \& 2}~\citep{cage_challenge_announcement,cage_challenge_2_announcement}, 
a Blue cyber-defence agent is tasked with defending a computer network 
that consists of a 
user network for staff, 
an enterprise network, 
and an operational network which contains key manufacturing and logistics servers. 
A Green agent is used to model the effect of network users, 
while a Red cyber-attacking agent confronts the Blue cyber-defence agent with a  
post-exploitation lateral movement scenario. 
The Red agent begins each episode with root access on a user host in Subnet 1. 
Its objective is to reach an operational server and begin
to degrade network services. 
Therefore, the Blue agent’s objective is to minimise the presence of the Red 
agent and ensure that network functionality is maintained~\citep{kiely2023autonomous}.
Blue is capable of running a detailed analyses on hosts and removing malicious software. 
The Blue agent can also opt to remove a host and restore it using
a clean backup, for cases where the Red agent is too well established.
In addition, Blue has the option of creating decoy services.
The main difference between CAGE Challenges 1 and 2, 
is that the latter provides a more representative action space~\citep{kiely2023autonomous}.}

\update{\textbf{TTCP CAGE Challenge 3} provides a scenario that is  
tailored to some of the unique requirements of military 
systems~\citep{cage_challenge_3_announcement}.
The challenge is to develop a decentralised 
cyber-security system consisting of a swarm of unmanned aerial vehicles (UAVs)
that form a mobile ad-hoc network (MANET).
The MANET allows soldiers patrolling a border
to communicate with each other, using the UAVs 
to relay messages.
However, in this scenario Red have managed to install
hardware Trojans on a subset of drones in the swarm.
Once activated Trojans can deploy a worm that will
attempt to compromise UAVs, capable of accessing
Blue's data regarding troop movement, and 
introducing false information to mislead command \& control.
Blue's task is to detect, isolate and reclaim 
compromised UAVs, while ensuring that the flow 
of data on the MANET is maintained. 
The default scenario consists of a team of 18 UAVs, 
each implemented with a Linux system. 
Communication is enabled via radio-based networking 
with any other systems (UAVs or troops) that are 
within communication range.}

\update{Agents can take either local or remote actions, 
with local actions only acting on the UAV on which they 
are executed, while remote actions can impact connected UAVs. 
Two UAVs are considered connected if there is a communication 
route between them.
Considerations while executing remote actions include that:
i.) An action can fail if any UAV along the communication path
drops the remote action’s network traffic, and; ii.)  
Remote actions consume bandwidth in the communication range 
of each UAV along the shortest path. 
For the latter, if the bandwidth limit is exceeded, then
actions will be randomly selected or dropped until 
the bandwidth usage returns below the limit.}

\update{Red agent actions include ExploitDroneVulnerability, SeizeControl, 
FloodBandwidth, BlockTraffic, and AllowTraffic. 
If a Red agent successfully executes ExploitDroneVulnerability, 
then it gains low privileged code execution rights on the 
target UAV. 
This action can be followed by the SeizeControl action to 
create a new Red agent that takes control over the target UAV
and removes the Blue agent. 
In response, Blue agents can inhibit Red agents via actions,
including: RetakeControl, RemoveOtherSessions, BlockTraffic, 
and AllowTraffic. 
The effectiveness of the Blue cyber-defence approach 
is determined by the extent to which data transfers
between soldiers (Green agents) are successful.
Blue is considered to have lost an episode if the
swarm reaches a state from which Blue cannot
recover.}

\update{\textbf{TTCP CAGE Challenge 4} sticks with the theme of
Red attempting to disrupt a swarm of UAVs 
from CAGE Challenge 3~\citep{cage_challenge_4_announcement}.
However, this time attacks are performed against
a base-station from which the activities of the
UAVs are coordinated.
The network that Blue is tasked with defending
consists of a range of operational and back-office
enterprise networks, supporting various military 
operations.
The UAVs are controlled via a contractor subnet,
which connects to the base via the internet.
To increase security levels the network
is highly segmented, and as a result, once
again requires a multi-agent solution. 
Each agent is assigned a zone that it must 
protect.
Similar to CAGE Challenge 3 communication 
between agents is restricted with respect
to bandwidth limitations.}

\subsubsection{\update{CyberBattleSim}}

\update{CyberBattleSim~\citep{10216719,shashkov2023adversarial,walter2021incorporating,cyberbattlesim} 
provides a deliberately simplistic and abstract cyber-defence environment, 
with the developers' original goal being to prohibit direct applications 
of developed cyber-attacking approaches to real-world systems. 
This concern is due to the original version of the framework focusing
on the training of Red cyber-attacking agents. 
The goal of the Red agents is to take ownership of a portion of the network through
exploiting vulnerabilities on hosts~\citep{cyberbattlesim}. 
A basic stochastic Blue defender agent was provided with the original version 
to detect and mitigate ongoing attacks.}

\update{This environment has benefited from its code being made openly available, 
which has resulted in a number of useful extensions.
\cite{10216719} added the facility for training Blue cyber-defence agents,
thereby providing the necessary components for adversarial learning~\citep{MARLon}, 
while \cite{walter2021incorporating} included deceptive elements such as 
honeypots and decoys.  
For the latter extension, 
the defensive benefits of the deceptive elements were tested 
through applying DRL based attackers using a capture the flag scenario.
The authors found that the attacker’s progress was
dependent on the number and location of the deceptive elements.} 

\update{Thanks to its simplicity, CyberBattleSim provides an environment within which 
to conduct rapid experiments
and benchmark novel methods~\citep{cyberbattlesim}.
However, upon conducting an evaluation of adversarial learning,  
\cite{shashkov2023adversarial} caution that the practical use of discovered defender 
policies are limited due to Blue's action space lacking options such as patching services.
The current implementation only offers options to shut-down and reboot
compromised services.} 

\subsubsection{\update{FARLAND}}

\update{The developers of CyberBattleSim~\citep{cyberbattlesim} adopted
the stance that a cyber-defence training environment needs to provide
safeguards from the potential nefarious use of automated agents 
trained within it. 
In contrast, the developers of FARLAND
focus on confronting Blue cyber-defence agents with a high fidelity 
network simulation environment, with a gradual increase in the
complexity of models and the level of skill required to reconfigure networks
and mitigate cyberattacks~\citep{molina2021farland}.
The developers note the cyber-defence problem
can't be defined as a single network defence 
game with a simple set of rules. 
Instead a spectrum of games need to be mastered with respect to:
i.) Potential adversarial attack techniques and procedures (TTPs); 
ii.) The quality of service goals and network characteristics, and; 
iii.) The actions available to the defender, 
along with Blue's own security goals.}

\update{To enable a gradual progression the developers turned to approaches such
as unsupervised curriculum learning and generative programs that help
determine the difficulty of the task that Blue is confronted with. 
Therefore, FARLAND provides the facility for conducting adversarial learning
through updating Red behaviour via probabilistic programming,
allowing an attacker to find a parameterization that would optimally
bias the blue agent policy, and allow Red to take actions that are 
most likely to allow it to reach its objective~\citep{molina2021farland}.}

\update{\cite{molina2021farland} stress that Blue cyber-defence agents must not only
be well equipped to handle attacks performed by common adversaries,
such as those provided by MITRE’s ATT\&CK framework~\citep{strom2018mitre},
but must be able to handle increasingly sophisticated
adversaries. 
While a trained DRL agent can be effective in defending a network 
from attacks that match previous training conditions, 
insights are required regarding the performance
of a cyber-defence policy once the threats, network conditions, and security goals change~\citep{molina2021farland}.}

\update{With respect to cyber-defence scenarios, 
FARLAND games take place within a preconfigured
virtual software defined network, 
consisting of hosts, switches, network functions,
and a network controller. 
Throughout the course of an episode the Red adversary 
attempts to compromise hosts by using a subset of 
techniques described within MITRE’s ATT\&CK framework~\citep{strom2018mitre}
or using a set of deceptive actions. 
The MITRE ATT\&CK framework provides a curated knowledge base and model
for Red cyber-attacking behaviour, focusing on how external adversaries
operate within a compromised network. 
The Blue DRL agent is tasked with learning how to mitigate
these attacks, with a game ending upon one of the following
conditions occurring: 
i.)~The attacker achieves its objective; 
ii.)~The cyber-defence agent prematurely terminates the game (and receives an associated penalty); 
iii.)~The red agent concedes the game upon being contained by the blue agent, or; 
iv.)~A predefined time-limit is reached~\citep{molina2021farland}.}

\update{In addition to Blue and Red agents, FARLAND also features Gray agent 
behaviour in the form of a probabilistic model,
characterizing the expected behavior of authorized network users. 
The framework also provides an emulation environment for policy validation, 
that can operate on a system with an additional virtualization layer. 
This layer limits the risks of running adversarial programs on 
shared High Performance Computers~\citep{molina2021farland}.}

\subsubsection{\update{Network Attack Simulator (NASim)}}

\update{This \aco environment is a simulated computer network complete with vulnerabilities, 
scans and exploits designed to be used as a testing environment for AI agents and 
planning techniques applied to network penetration testing~\citep{schwartz2019autonomous}. 
NASim can support scenarios that involve a large number of machines. 
However, \cite{nguyen2020multiple} find that the probability of
\red agent attacks succeeding in NASim are far greater than in
reality. 
As a result the authors add additional rules to increase 
the task difficulty.
NASim also does not support adversarial learning.}

\update{\cite{janisch2023nasimemu} have extended NASim to enhance its realism, 
releasing a framework that contains a modified version of NASim and
an emulator (NASimEmu), which have a shared interface.
Therefore, agents can be trained within the extended version of NASim
and subsequently evaluated on the emulator. 
Changes that the authors have made to NASim include: 
\begin{itemize}
\item New dynamic scenarios that support random variations; 
\item Allowing agents to be trained and tested on multiple scenarios in 
parallel;
\item The size of host vectors across scenarios remaining constant;
\item The random permutation of nodes and segment IDs at the beginning
of each episode to prevent memorization;
\item Observations that retain revealed information;
\item The Red agent decides when an episode ends via a \emph{terminal action};
\item Observations can optionally be returned in the form of a graph with nodes
representing subnets and individual hosts;
\item The solution allows for observations to be visualised. 
\end{itemize}}

\update{The emulator has been implemented using industry-level tools, 
including Vagrant, VirtualBox, and Metasploit. 
These developments provide a means through which
to evaluate the extent to which cyber-attack agents
can bridge the sim-to-real gap between the modified
version of NASim and the provided emulator.
However, \cite{janisch2023nasimemu} also provide transparency
with respect to the limitations of the framework which, despite
their best efforts to make NASimEmu realistic,  
centre around the fact that the simulation remains an abstraction
of the penetration testing challenge.
Factors here include that: i.) Service versions cannot be differentiated;
ii.) Configurations between hosts on subnets are not correlated;
iii.) No facility is provided for storing and using discovered credentials;
iv.) Exploit actions are assumed to work if the corresponding service is 
running on the host, which will not always be the case in practice, \ie due to patching;
v.) The firewall currently blocks or allows all traffic between subnets, and;
vi.) Only the attacker can be trained / modelled currently.}

\subsubsection{\update{Primary-level AI Training Environment (PrimAITE)}}

\update{PrimAITE~\citep{PrimAITE} is a primary-level cyber-defence environment for training 
and evaluating AI approaches, providing a means through which to model 
relevant platforms and contexts. 
This includes the modeling of network characteristics such as
connections, IP addresses, ports, traffic loading, operating systems and services.
PrimAITE is highly configurable, 
providing a means through which to 
model a variety of mission profiles
and Red cyber-attack scenarios.
Lay-down configuration files can also be 
used to define Information Exchange Requirements (IERs)
that model pattern-of-life (POL) for both green agent (background) and red agent (adversary) behaviour, 
via the modelling of traffic loading on the network.
POL configurations allow for attacks (and defence) 
to be modelled within the confines of a node, including adversarial 
behaviour and mission data with a sliding scale of criticality.}

\subsubsection{\update{YAWNING TITAN}}

\update{The YAWNING TITAN environment intentionally abstracts away 
the additional information used by cyber-defence environments
such as CybORG, FARLAND and PrimAITE, 
facilitating the rapid integration and 
evaluation of new methods~\citep{YAWNING}.
It is built to train \aco agents to defend 
arbitrary network topologies that can
be specified using highly customisable 
configuration files. 
The developers adhered to the 
following design principles:
i.)~Simplicity over complexity;
ii.)~Minimal hardware requirements;
iii.)~Support for a wide range of algorithms;
iv.)~Enhanced agent/policy evaluation support, and;
v.)~Flexible environment and game-rule setup.}

\update{Due to deliberately abstracting the cyber-defence challenge,
YAWNING TITAN serves as an environment for modelling a large
variety of different network topologies, without the requirement
for specifying properties such as communication protocols between different types
of hosts~\citep{andrew2022developing}.
The environment also provides a lightweight database 
for saving game modes, which can be defined using a GUI.}

\update{Each machine in the network has parameters that 
affect the extent to which they can be impacted 
by \blue and \red agent behaviour, including
vulnerability scores on how easy it is for a node to be
compromised. 
The environment is highly configurable, 
allowing users to define success conditions 
for the Blue agent through defining new reward functions, as well as defining new 
Blue and Red actions~\citep{andrew2022developing}. 
However, while YAWNING TITAN is an easy to configure 
lightweight \aco environment, it currently 
lacks the ability to train \red agents,
and does not support adversarial learning.
The documentation does, however, provide detailed instructions
on how new custom rules-based Red agents can be added to the framework.}

\subsection{Desirable Properties}

\update{To implement, train and evaluate DRL agents for \aco, 
training environments are required that can
pose learners with the same challenges that 
a cyber-defence agent will be confronted with
once deployed.
Therefore, we shall now consider the desirable properties of cyber defence environments
for training and evaluating AI agents, and critically evaluate the extent to which the 
\acogyms listed above meet them.
We hypothesize that a DRL agent that performs well on a 
cyber-defence environment featuring all of the properties listed below 
stands a better chance of bridging the sim-to-real gap.
However, we note that even cyber-defence environments that 
do not meet all of the described properties still may have their
merits. 
For instance, despite being deliberately abstract, CyberBattleSim and YAWNING TITAN
provide a facility for rapid benchmarking and short development 
evaluation cycles for improving novel approaches.
We have identified a total of eleven desirable properties  
to assess the suitability of cyber-defence environments for 
developing and assessing techniques for tackling challenges
in the \aco domain. 
These can be seen listed in~\autoref{tab:aco_envs}, 
while a critical evaluation of cyber-defence environments
with respect to the properties is provided below.}

\subsubsection{Fidelity} 

\update{Current cyber-defence environments range 
from deliberately abstract (YAWNING TITAN, CyberBattleSim, NASim)
to increasingly high fidelity environments (PrimAITE and FARLAND).
Naturally, one of the limitations of the former 
is that approaches that perform well on the environment
will be of very limited practical use~\citep{shashkov2023adversarial}.    
Nevertheless, low-fidelity environments also have advantages.
For example, through emphasising simplicity over complexity
and having minimal hardware requirements, environments such 
as YAWNING TITAN provide a means through which to conduct rapid benchmarking
and improving approaches that sit on the lower end of 
the technology readiness level (TRL) scale~\citep{YAWNING}.
As solutions are scaled through the TRLs they can increasingly
be confronted with challenges that high fidelity environments
bring to the table, \eg more realistic observations in the form 
of network events and traffic between hosts~\citep{molina2021network},
as well as more representative, data driven, attacks by Red agents~\citep{strom2018mitre}.}

\subsubsection{Emulators} 

\update{For evaluation purposes, both FARLAND and CybORG recognise that 
a compatible simulation-emulation approach 
is required to facilitate model validation~\citep{molina2021farland,cyborg_acd_2021}.
However, \cite{molina2021farland} observe that 
the CybORG design does not explicitly allow the modelling 
of asymmetrical behavior and goals between the Red and Blue agents.
It is worth noting that the developers of PrimAITE 
list an integration with external threat emulation tools, using either offline data or through
integrating data at runtime, under future ambitions~\citep{PrimAITE}.}

\subsubsection{\textbf{Representative Actions:}} 

\update{
An idealised \acogym provides a large number of represenative actions, 
such as scanning/monitoring for Red activity;
isolating, making safe, and reducing the vulnerability of nodes; 
adding firewall rules, and;
deploying decoy nodes.
Naturally there is a trade-off here between 
offering cyber-defence (and attacking) agents
a granular space over which to make decisions,
and sufficient abstraction to simplify the 
learning task.
Without the use of observation and action wrappers 
that provide either a MultiDiscrete formulation or a reduced flattened Discrete 
action space, \acogyms feature potentially large,
rich combinatorial action spaces that are intractable for
traditional DRL approaches. 
}

\update{
As we have outlined in Section \ref{sec:action_spaces},
MultiDiscrete action formulations offer a number of
benefits for \aco, including a linear increase in the number 
of policy outputs along each degree of freedom~\citep{tavakoli2018action},
\eg through treating the type of action, target host, source host, 
and credentials as separate axes~\citep{shashkov2023adversarial}. 
%
%
A multi-discrete space, 
such as the one that is a property of \aco problems, 
can cause an exponential scaling 
of the size of the space when flattened,
such that it quickly becomes computationally intractable 
for standard DRL approaches~\citep{tavakoli2018action}.
As a result, \acogyms that already provide a MultiDiscrete actions wrapper,
such as CybORG, are highly desirable.}

\update{Naturally, \acogyms that provide an option for choosing the granularity of the 
action space can provide a significant advantage when attempting to scale an 
approach through the TRLs.
Here, one of the more recent additions to the cyber-defence landscape, 
PrimAITE, provides options for choosing different types of action spaces, 
including: 
i.) \emph{NODE}, where The Blue agent can influence the status of nodes by switching them off, resetting,
or patching operating systems and services;
ii.) \emph{Access Control List} (\emph{ACL}), where 
the Blue agent can update the (highly configurable) access control list rules, thereby
determining the configuration of the system-wide firewall, and;
iii.) \emph{ANY}, where actions from both the NODE and ACL setting can be selected.
The ACL action space enables definitions that follow the standard 
ACL rule format (\eg DENY/ALLOW, source IP address, destination IP address, protocol and port).
The NODE action space, meanwhile, significantly reduces the
dimensionality of the action space compared to the ANY and ACL settings.
The ANY setting represents the most representative action space.
Therefore, solutions that can master it are more likely to scale.}
%

\subsubsection{Source Code Availability} 

\update{The availability of code can significantly speed-up research efforts.  
In addition, open-sourced cyber-defence environments can serve
as a starting point for others to add their own extensions.
A perfect example here is CyberBattleSim, 
which started out as a framework for training
cyber-attacking agents, and now features the 
facility for also training Blue agents 
and conduct adversarial learning~\citep{10216719,MARLon}.
Thanks to the efforts from \cite{walter2021incorporating} the
environment now also features an extended Blue action space,
allowing the defender to launch honeypots and decoys.  
Similarly, \cite{wiebe2023learning} have
extended CybORG CAGE Challenge 2 into multi-agent variation called CyMARL,
implemented with the PyMARL environment interface~\citep{samvelyan2019starcraft}.
The extensions also include additional game types, actions, and network
topologies.}

\update{Openly available training and evaluation environments
also help ensure the reproducibility of results,
and thereby the quick and fair benchmarking of novel
methods~\citep{lucchi2020robo}.
With respect to the \acogyms discussed above,
FARLAND is the only one for which we are
unable to find openly available source code.}

\subsubsection{Trainable Blue and Red agents}

\update{An idealised \acogym should provide the facility for optimizing both
\blue cyber-defence and \red cyber-attacking agents.
We observe that while there are environments that provide
a means to optimize both \blue and \red behaviour (CybORG,
FARLAND and the extended version of CyberBattleSim from \cite{10216719})
other environments focus on either trainable \blue (YAWNING TITAN 
and PrimAITE) or \red agents (NASim and NASimEmu). 
While trainable Blue and Red agents are a prerequisite 
for adversarial learning, we note that an environment 
featuring both trainable Blue and Red agents doesn't 
necessarily indicate that the environment supports 
the training of Blue and Red learning approaches 
against each other.
We shall define the necessary conditions for
adversarial learning below.}

\subsubsection{Adversarial Learning:} 
\update{To obtain robust and resilient cyber-defence agents, 
that can generalize across a plethora of Red cyber-attacking approaches,
we are interested in \aco environments that can confront learners 
with the adversarial learning challenge.
However, despite the cyber-defence challenge representing an
adversarial game by definition, 
adversarial learning is far from a common feature found
within \acogym implementations. 
In-fact, only CybORG, FARLAND and CyberBattleSim 
(the extended version from \cite{10216719})
currently provide the facility for optimizing
\emph{both} Blue and Red approaches.
Here it is worth noting that for CybORG
we find that the implementation of an additional 
observation wrapper is necessary in order to load
trained Blue or Red DRL policies as opponents.} 

\update{This reluctance for providing full support 
for this adversarial learning challenge
may be partially routed in concerns 
regarding the potential nefarious use 
of cyber-attacking agents trained within
a high-fidelity gym environment~\citep{cyberbattlesim}. 
However, \cite{molina2021network} note that there 
is limited value in evaluating a cyber-defence policy
based on its performance against Red agents with arbitrary behaviours
and fixed TTPs.
Instead, Blue agents must be equipped to generalize across
opponents that can change their behaviour, and thereby the
game itself~\citep{molina2021network}.
\autoref{sec:adv_learning} will focus on methods
for ensuring an adversarial learning process leads to 
a policy that can generalize across opponents within 
a given environment. 
However, the point made by \cite{molina2021network}
highlights that to ensure a policy that can generalize across
different types of attacks once deployed, the ability for
Red to launch a diverse array of attacks within the training
environment is critical.}
%

\subsubsection{Cooperative Multi-Agent Learning (CMAL)}

\update{Many real-world network defence scenarios can benefit 
from decentralized solutions, \eg using a multi agent approach~\citep{molina2021farland}.  
For instance, with an ever growing number of application areas for hardware, 
such as intelligent drones and smart vehicles, there is an 
increased interest in decentralized cyber-defence solutions for MANETs~\citep{kusyk2018survey}.
Given the challenges of defending a network where hosts can be on the move and
communications between nodes can drop out at any point in time, decentralized 
defence systems have also been a topic for TTCP CAGE Challenges 3 and 4~\citep{cage_challenge_3_announcement,cage_challenge_4_announcement}.
In addition, the developers of PrimAITE have listed the implementation of a suitable 
standardised framework to allow multi-agent integration as an ambition for future
enhancements~\citep{PrimAITE}.}

\update{Multi-agent learning environments provide an opportunity to benchmark the benefits of 
approaches that can share critical information, such as network topology, 
known or suspected threats, and event logs, thereby enabling a more informed decision-making 
process~\citep{contractor2024learning}.
Depending on the levels of self-interest of mobile hosts within a MANET, 
the problem can either be formulated as a Dec-POMDP (host objectives are 
aligned, \ie a team-game) or POMG (objectives are somewhat aligned, \ie
we have a general-sum game).
Therefore, this type of network represents a cooperative multi-agent learning
challenge.  
The formulation is also relevant for networks spanning multiple geographic locations~\citep{gady2020cyber}.
In essence, any cyber-defence challenge where physical restraints, 
such as data transmission capacity and latency,  
can necessitate a multi-agent (reinforcement) learning approach.}

\subsubsection{Stochastic} 

\update{The \aco problem features stochastic factors, \eg 
user activity, 
equipment failures and 
the likelihood of a Red attack succeeding. 
To meet this criteria stochastic state transitions are required.
Random starting positions (or ports and IP addresses) alone are insufficient.
For example, FARLAND provides probabilistic models that characterize network dynamics
and the behaviors of adversarial (red) and benign (gray) agents~\citep{molina2021network}.
CybORG features randomization with the choice for the session on 
which decoys are created \citep{cyborg_acd_2021}, while YAWNING TITAN assigns 
an abstracted vulnerability score to each node, determining the likelihood
with which a \red attack will succeed.}

\subsubsection{Partially Observable} 

\update{Given the impracticality of being able to observe and process
the full state of a network, the typical assumption made 
for cyber-security agents is that they act on partial
and imperfect state information, with Blue typically
having access to more information than Red~\citep{molina2021network}. 
For example, Blue agents are often assumed to have knowledge
regarding which nodes on the network are connected, and
can easily access information on services running on each host and 
the ports that are being used~\citep{molina2021network}.
Therefore, the accurate modelling of partial observability,
and determining which features are observable, are key considerations.}

\update{In \acogyms Blue must typically perform a scan/monitor action 
to observe Red activities.
The degree of the partially observability
challenge may differ across \acogyms, with  
more sophisticated Blue policies being required  
to detect (observe) Red agent activity within more
high fidelity environments~\citep{molina2021network}.
Here, again we observe different levels of abstractions within cyber-defence
environments, ranging from node vulnerability features (YAWNING TITAN) to
more granular features of the state of various components such as hardware,
operating system, file system, and services (PrimAITE).
The latter type of environments will be critical for obtaining
\aco agents that can bridge the \emph{sim-to-real gap}.} 

\subsubsection{Graph-Based Observations} 

\update{With \aco taking place on networks, \acogyms contain underlying
dynamics unique to graph-based domains.
Most notably, in the \acogyms discussed in this section, 
a Red agent can only move through the network by jumping
from a node which it has previously compromised to a connected node. 
As such, the topology and connectivity
of the graph directly influences the behaviour 
of the Red agent, and, therefore, the game itself.}

\update{\cite{shashkov2023adversarial} observe that one of the weaknesses of CyberBattleSim is that the 
trained agents only work in the particular network they are trained on, 
and their strategies would be ineffective in a different topology. 
This, again, re-iterates the need for a more appropriate solution for formatting observations,
\eg through using graph-based observations.
A graph-based representation of 
the environment also holds much
potential as a suitable data structure 
for cyber-defence tasks~\citep{bilot2023graph}.}

\update{To process graph-based observations, 
graph neural networks (GNNs) have emerged 
as a powerful tool~\citep{JIANG2022117921}.
These are capable of processing a graph's uneven structure, 
irregular size of unordered nodes, and dynamic 
neighbourhood compositions~\citep{kipf2016semi}.
Given the challenge that network configurations include a set of hosts, 
which may change during an episode~\citep{molina2021network},
GNNs are also well equipped to handle arbitrary 
sized network topologies with respect to the number 
of hosts, thereby providing a solution that can
scale to large networks~\citep{bilot2023graph}.}

\update{DRL approaches using GNNs as feature extractors have
much potential for autonomous networking solutions~\citep{tam2024survey},  
including cyber-defence~\citep{munikoti2023challenges}.
However, interestingly the current versions of the majority of 
cyber-defence environments discussed in this section
do not feature a graph-based observation option,
with NASimEmu~\citep{janisch2023nasimemu} being the only exception.
Therefore, in \autoref{tab:aco_envs}
we evaluate the environments with respect
to the amount of effort that would be required to
add a graph-based observation wrapper to the 
environment, for DRL-GNN approaches to become applicable.}

%
%

\begin{table}[h]
\centering
\resizebox{\columnwidth}{!}{
\begin{tabular}{|l||c|c|c|c|c|c|c|c|c|c|}
\hline
\multicolumn{11}{|c|}{\textbf{Autonomous Cyber Defence Environments}}\\
\hline
\textbf{Environment Name} &  \textbf{CC1} &  \textbf{CC2}  &  \textbf{CC3} &  \textbf{CC4} &  \textbf{CBS} &  \textbf{FARLAND} &  \textbf{NASim} &  \textbf{NASimEmu} &  \textbf{PrimAITE}  &  \textbf{YT} \\
\hline
High Fidelity & \xmark & $\nearrow$ & $\nearrow$ & $\nearrow$ & \xmark & \checkmark & \xmark & \xmark & \checkmark & \xmark \\ 
\hline
Representative Actions & \xmark & \checkmark & \checkmark & \checkmark & \xmark & \checkmark & \xmark & \checkmark & \checkmark & \xmark \\
\hline
Emulator & \checkmark & \checkmark & \checkmark & \checkmark & \xmark & \checkmark & \xmark & \checkmark & \xmark & \xmark \\ 
\hline
Available Source Code & \checkmark & \checkmark & \checkmark & \checkmark & \checkmark & \xmark & \checkmark & \checkmark & \checkmark & \checkmark \\
\hline
Trainable Blue & \checkmark & \checkmark & \checkmark & \checkmark & \checkmark & \checkmark & \xmark & \xmark & \checkmark & \checkmark \\
\hline
Trainable Red & \checkmark & \checkmark & \checkmark & \checkmark & \checkmark & \checkmark & \checkmark & \checkmark & \xmark & \xmark \\
\hline
Adversarial Learning & $\nearrow$ & $\nearrow$ & $\nearrow$ & $\nearrow$ & \checkmark & \checkmark & \xmark & \xmark & \xmark & \xmark \\
\hline
Cooperative Learning & \xmark & \xmark & \checkmark & \checkmark & \xmark & \xmark & \xmark & \xmark & $\nearrow$ & \xmark \\
\hline
Stochastic Transitions & \checkmark & \checkmark & \checkmark & \checkmark & \checkmark & \checkmark & \checkmark & \checkmark & \checkmark & \checkmark \\
\hline
Partially Observable & \checkmark & \checkmark & \checkmark & \checkmark & \checkmark & \checkmark & \checkmark & \checkmark & \checkmark & \checkmark \\
\hline
Graph-Based Observations & $\nearrow$ & $\nearrow$ & \xmark & \xmark & $\nearrow$ & $\nearrow$ & $\nearrow$ & \checkmark & $\nearrow$ & $\nearrow$ \\
\hline
\end{tabular}}
\caption{An overview of \acogyms and their properties, 
with \checkmark and \xmark indicating whether the
\env fulfills a criteria. The third symbol, $\nearrow$
indicates that the environment does not fulfil the criteria currently.
This may be due to enabling software features not being present, whether by design
or due to incomplete implementations. 
However, the required features could be
implemented with a modest amount of development effort, without the need for
major extensions of environment scope.}
\label{tab:aco_envs}
\end{table}

\subsection{\update{Relevant Non-Cyber Defence Environments}}

Since the specific observations and actions are \acogym dependant, it can be surmised
that these are not particular to the \aco problem. 
\update{Indeed, there are non-cyber defence environments 
that confront learners with similar challenges, and where, therefore, solutions 
might translate to the \aco domain. 
For example, drawing inspiration from the action selection approach used by AlphaStar~\citep{vinyals2019grandmaster},
an approach that was designed to select complex actions for the video game StarCraft,
\cite{shashkov2023adversarial} developed a corresponding multi-stage neural network based policy that could
be applied to a complex action space within the CyberBattleSim environment.
Through applying the multi-stage approach, the policy builds actions via 
neural networks focusing on attack types, then target, source, etc., resulting in the 
required action signature.}
\update{Therefore,} any other environment which shares core features in its internal dynamics, 
and structure of observation and action spaces, should provide a suitable 
platform for developing methodologies to tackle \aco challenges before the
\acogyms themselves are able to fully represent the problem.
\update{As such, any approaches that have already been developed that work well on
environments that confronts learners with the same challenges as \aco, should provide a 
suitable approach for our current problem formulation.} 

A good example of an environment that confronts learners with many of the
same challenges present in \aco is \emph{MicroRTS}, 
a simple Real-Time Strategy (RTS) game~\citep{huang2021gym}. 
It is designed to enable the training of \RL agents on an environment 
which is similar to {PySC2}~\citep{vinyals2017starcraft}, the \RL
\env adaptation of the popular RTS game \sctwo which possesses huge complexity. 
MicroRTS is a lightweight version of PySC2, without the extensive
(and expensive) computational requirements.
The game features a gridworld-like observation space, where for a map of size $h \times w$, the observation space
is of size $h \times w \times f$, where $f$ is a number of discrete features which may exist in any square.
This observation space can be set to be partially observable.
The action space 
is a large multi-discrete space, where each worker is commanded via a total
of seven discrete actions.
The number of workers will change during the game, making the core RTS action space intractable.
In order to handle this, the authors of {MicroRTS} implemented action decomposition as part of the environment,
and the action space is separated into the unit action space (of 7 discrete actions), and the player action
space.
While the authors discuss two formulations of this player action space, the one which is documented in code is
the \emph{GridNet} approach, whereby the agent predicts an action for each cell in the grid, with only the
actions on cells containing player-owned units being carried out.
This leads to a total action space size of $h \times w \times 7$. 
The challenge of varying numbers of agents is remarkably similar to the challenge of varying numbers of
nodes in \aco. 
The {MicroRTS} environment, if action decomposition were to be stripped away, would be a
strong candidate for handling large and variably-sized discrete action spaces.


%
%
\update{In summary, in this section, we have critically reviewed multiple cyber-defence environments 
for training and evaluating autonomous cyber-defence agents. 
While none of the environments reviewed meet all of our desirable criteria,  
each desirable property is represented in at least one environment.
Particularly, noteworthy is that current \aco environments lack a facility for conducting 
adversarial learning.
In addition, all environments could benefit from graph-based observation wrappers to
facilitate the processing of observations by flexible function approximators, \ie 
graph-neural networks and transformers~\citep{kortvelesy2022qgnn,hong2022structureaware,munikoti2022challenges,parisotto2020stabilizing}.}

\section{Coping with Vast High-Dimensional \update{States}} \label{sec:hdss_approaches}

\update{Gathering and processing cyber-security data 
represents a challenging undertaking.
Due to a lack of common standards that 
apply across vendor-specific device implementations, 
large networks, in particular those incorporating 
IoT systems, have the potential to confront autonomous
cyber-defence solutions with high-dimensional and multimodal data~\citep{10136827}.
As a result, the design and implementation of efficient algorithms 
for inferring the state of a large network, from limited
measurements, has been a topic of much interest to
network control researchers over the past decade~\citep{montanari2022functional}.
Here, AI-based security systems are increasingly turning to deep learning 
approaches thanks to their ability to automatically extract relevant features 
from input data~\citep{10142590}. 
}

\update{When training Deep Neural Networks (DNNs), 
learning efficiency is often reduced due to unnecessary 
features contributing noise~\citep{zebari2020comprehensive}.}
Dimensionality reduction techniques are a natural choice
for dealing with unnecessary/noisy features. 
Benefits include: 
the elimination of irrelevant data and redundant features, 
while preserving the variance of the dataset;
improved data quality;
reduced processing time and memory requirements;
improved accuracy;
shorter training and inference times (\ie reduced computing costs), 
and; improved performance~\citep{zebari2020comprehensive}.
Two popular approaches are \emph{feature selection} and \emph{feature extraction}.
Feature selection aims to find the optimal sub-set of relevant features 
for training a model, which is an Non-deterministic Polynomial (NP)-hard problem~\citep{meiri2006using}. 
In contrast, feature extraction involves 
creating linear combinations of the features, 
while preserving the original relative distances
in the latent structures~\citep{zebari2020comprehensive}.
The dimensionality is decreased without losing 
much of the initial information~\citep{zebari2020comprehensive}.
However, the resulting encodings are uninterpretable for humans.

DRL uses DNNs to directly extract features from high-dimensional data~\citep{mousavi2018deep,almasan2022deep}.
\update{Nevertheless, feature selection can provide a valuable pre-processing step for training DNNs for cyber-security applications, 
with the potential to significantly boost performance~\citep{davis2020feature}.}
For example, anomaly detection DNNs for cyber-security applications, tasked with
differentiating benign from malicious packets, were found to perform 
better when trained on data where feature selection had 
been applied~\citep{9216403}.
\update{Furthermore, through applying signal-to-noise ratio feature selection
to network traffic data from the 2003–2007 and 2009 Department of 
Defence Cyber Defence Exercises (CDXs),
\cite{moore2017feature} observe that a 90\% data reduction is possible 
with a negligible reduction in performance.
\cite{mohammadi2019cyber} propose an intruder detection
systems that is based on feature selection and a clustering
method that has a lower false positive rate compared to existing methods.
For an overview of feature selection techniques based
on filtering methods for cyber-attack detection, see the survey from~\cite{lyu2023survey}.}

\subsection{\update{A System-of-Systems Observation Approach}}

\update{While the above observations hint at the benefits of feature selection 
methods for DRL agents trained on granular cyber-defence data, 
the resulting models can also provide the necessary components 
for a \emph{system-of-systems} approach. 
Here the outputs from models trained to detect malicious packets and 
intruders feed into a DRL agent, capable of learning how to best respond.
However, as evident from the observation spaces
featured within current \aco-gyms that essentially adopt this
formulation (see \autoref{sec:envs}),
cyber-defence agents will still be confronted
with high-dimensional observation spaces.
Specifically, each host, on a potentially large network,
contains features indicating whether the host has been compromised, 
along with operational status of services, hardware and software, 
etc.~\citep{YAWNING,PrimAITE,cyborg_acd_2021}.}

\subsection{The Benefits of GNNs for ACD-DRL Agents} \label{sec:function_approximators}

\update{\aco agents are confronted with the challenge
of the size of the observation space potentially fluctuating, 
\eg due to hosts joining and leaving the network.
As a result, in this section we will consider the topic of feature
extraction for our idealised ACD-DRL agent on large dynamic networks. 
Specifically, we shall focus on the potential advantages of using GNNs,
an approach that is particularly well suited for processing network data
resulting from this discussed system-of-systems approach.}
\update{In the sections that follow,} we shall discuss state of the art 
DRL approaches for further eliminating unnecessary 
information through learning \emph{abstract} states
and discuss advanced exploration strategies
towards enabling the sufficient visitation of all \emph{relevant} states
to obtain accurate utility estimates~\citep{burden2021latent,pmlr-v134-perdomo21a}. 
Finally, we shall take a closer look at approaches towards mitigating
catastrophic forgetting, DNN's tendency to unlearn previous 
knowledge~\citep{pmlr-v119-ota20a,de2015importance}.

Many of the successes in \RL over the past decade rely on the ability
of DNNs to identify intricate structures 
and extract compact features from complex high-dimensional samples
~\citep{Goodfellow-et-al-2016,karpathy2014large,lecun2015deep}.
These approaches work well for numerous domains that confront learners
with the \cod, such as the~\emph{Arcade Learning Environment} (ALE)~\citep{bellemare2013arcade},
where DNNs are used to encode image observations.
However, many of these domains are fully observable and contain state spaces
with a dimensionality that is manageable for current DNNs.
\update{In addition, the dimensionality of the observations is often static
for standard DRL benchmarking environments.} 
In contrast, for \aco environments  
%
the architectures
used by standard DRL approaches often cannot scale, necessitating
innovative solutions, \eg due to feature extraction layers that 
cannot handle arbitrary sized inputs.  
Here, considerations are required with respect to 
efficiently encoding an \emph{overwhelmingly} large observation space to a low dimensional
representation, while limiting concessions regarding performance.

\update{Below, we will provide an overview of works that have applied a 
graph machine learning based feature extraction technique to 
cyber-defence problems.
For background reading on the topic of GNNs we recommend the following
seminal papers~\citep{kipf2016semi,hamilton2017inductive,GAT,you2019position,nicolicioiu2019recurrent} 
and surveys~\citep{wu2020comprehensive,zhou2020graph,zhou2022graph,waikhom2023survey,khemani2024review}.
In addition, a survey by \cite{munikoti2022challenges} provides a comprehensive
overview of GNN approaches that can be combined with DRL, along with opportunities and challenges. 
The authors define an idealised GNN for DRL as:
i.) \emph{dynamic}; 
ii.) \emph{scalable}; 
iii.) providing \emph{generalizability},
and; iv.) applicable to \emph{multiagent} systems.}

\update{Dynamic refers to DRL's need for GNN approaches that 
can cope with time varying network configurations and 
parameters.
Autonomous cyber-defence tasks for instance may require GNNs that 
can cope with a dynamic network where 
the number of hosts vary over time~\citep{9789835,sun2020combining}. 
While GNN architectures have been proposed for dynamic
graphs (\eg spatial-temporal GNNs~\citep{nicolicioiu2019recurrent}), 
tasks including node classification, link prediction, community detection, and graph
classification in dynamic graphs could benefit from further improvements~\citep{munikoti2022challenges}.}

\update{With regard to generalizability, there is a danger that 
a DRL agent can overfit on the graph
structure(s) seen during training, and become unable to generalize
across different graphs~\citep{munikoti2022challenges}.
Therefore, to obtain GNN-DRL policies that can generalize, 
training environments are required that confront learners
with different network topologies. 
We shall now turn our attention towards instances where GNNs have successfully
been applied to the topic of cyber-defence to-date.}

\update{GNNs for cyber-security are becoming increasingly topical, 
with a recent surveys by \cite{ruan2023deep} and \cite{mitra2024use} highlighting
the benefits of graph based data structures and graph machine learning solutions for 
an ever evolving cyber-threat landscape.
The strength of GNNs lies in the fact that they can view the intricate nature of the 
cyber-defence problem through a graph-based lens, allowing them to capture 
nuanced patterns that emerge between interconnected entities, yielding 
insights that evade conventional methods~\citep{mitra2024use}.}

\update{Despite the above benefits, while end-to-end trainable 
DRL-GNN solutions have been utilized outside of the context 
of cyber-defence~\citep{9793853,HUANG2023119,9826438,park2021learning},
to-date the autonomous cyber-defence literature lacks an evaluation
of DRL-GNN approaches on cyber-defence environments.
Nevertheless, this topic is on the \aco community's radar. 
\cite{9789835} noted the need for scalable function approximators 
upon training cyber-defence agents to defend against attack
graphs formulated using the Meta Attack Language, 
observing that as the networks increase in size traditional 
approaches are no longer viable. 
Their proposed solution is to aggregate
information from a neighbourhood of nodes using 
a GNN when making decisions.}
%
%
%
%

\update{Beyond DRL for autonomous cyber-defence, 
there exists literature on the application of GNNs 
to detect and mitigate attacks, and the extent to which methods 
take advantage of their knowledge propagation and learning capabilities~\citep{mitra2024use}.
GNNs being capable of such aggregations have been shown in a number of existing works.
\cite{biswas2023intrusion} introduce a GNN based intruder detection model
using Lyapunov optimization, showing an improved detection accuracy
compared to baseline methods on the Aegean Wi-Fi Intrusion Dataset~\citep{9360747}.
Similarly, \cite{lo2022graphsage} take advantage of the fact that training and evaluation data 
for network intrusion detection systems (NIDS) can be represented as flow records that can naturally 
be represented by a graph, demonstrating the advantages of this approach on NIDS benchmark datasets.
Such benefits were also observed by \cite{pujol2022unveiling}, finding that their proposed GNN model 
achieves state-of-the-art results on the CIC-IDS2017 dataset. 
In addition, the authors find that their GNN based approach is more robust towards common adversarial attacks. 
These intentionally modify the packet size and interarrival times to avoid detection. 
The authors find that while the F1 scores of current SOTA ML techniques degrade up to 50\% when confronted 
with these types of attacks, the GNN architecture manages to maintain the same level of accuracy.}

\update{While we were unable to find end-to-end trainable DRL-GNN solutions for cyber-defence,
the DRL literature does provide examples of cyber-defence frameworks for 
cyber-physical systems that feature both approaches.
\cite{lin2022privacy} propose a deep reinforcement learning-based 
privacy-enhanced intrusion detection and defence mechanism
that consists of three modules:  
i.) A privacy-enhanced topology graph generation module; 
ii.) A GNN-based user evaluation module, and; 
iii.) A DRL-based intruder identification and handling module. 
The framework was found to excel in intrusion detection accuracy, 
intrusion defence percentage, and privacy protection.}

\update{
\cite{sun2020combining} used DRL-GNN to obtain a Network Function Virtualization (NFV) placement policy, 
tasked with deploying software to implement network functions as virtual instances. 
The authors turned to DRL-GNN solutions due to standard DRL-based 
solutions being unable to generalize well to different topologies. 
The authors showed that their DRL-GNN solution, DeepOpt, 
outperforms SOTA NFV placement schemes 
while demonstrating a better generalization ability across different network topologies.
Recently, \cite{lee2023graph} leveraged GNNs for imitation learning, enabling a mapping from defenders’ 
local perceptions and their communication graph to their actions, resulting in 
a scalable decentralized multi-agent perimeter defence system.
The above results, combined with the fact that GNNs are being incorporated into solutions from other ML paradigms,
hints at exploring the benefits of GNN-DRL solutions for \aco being a rich topic area for
future developments.}

\subsection{State Abstraction} \label{sec:state_abstraction}

\update{While the \emph{system-of-systems} approach 
discussed at the beginning of this section acts as 
a buffer between DRL agents and raw cyber-defence
data, the challenge of dealing with a large observation 
space remains.
For example, Blue observation vectors for the 
default network configuration in CybORG CAGE Challenge 2 
contain 11,293 features~\citep{cage_challenge_2_announcement}.
Given that this is for a small network (featuring 3 subnets, 8 hosts, 
3 enterprise servers and an operational server) further considerations 
are required to reduce the dimensionality of the observation space.
In this section we consider the suitability of \emph{state abstraction} approaches 
for this task.}
 
The aim of state abstraction is to obtain a compressed model of an environment 
that retains all the useful information -- enabling the efficient training
and deployment of a DRL agent over an abstract formulation. 
Solving the abstract MDP is equivalent to solving the underlying 
MDP~\citep{pmlr-v108-abel20a,abel2019state,burden2018using,burden2021latent}. 
%
State abstraction groups together semantically similar 
states, abstracting the state space to a representation with 
lower dimensions~\citep{yu2018towards}. 
A motivating example for the importance of state abstraction is the
ALE game Pong~\citep{bellemare2013arcade}, where success only
requires access to the positions and velocities of the two 
paddles and the ball~\citep{pmlr-v97-gelada19a}. 
%

\begin{figure}[h]
\centering
\includegraphics[width=0.4\columnwidth]{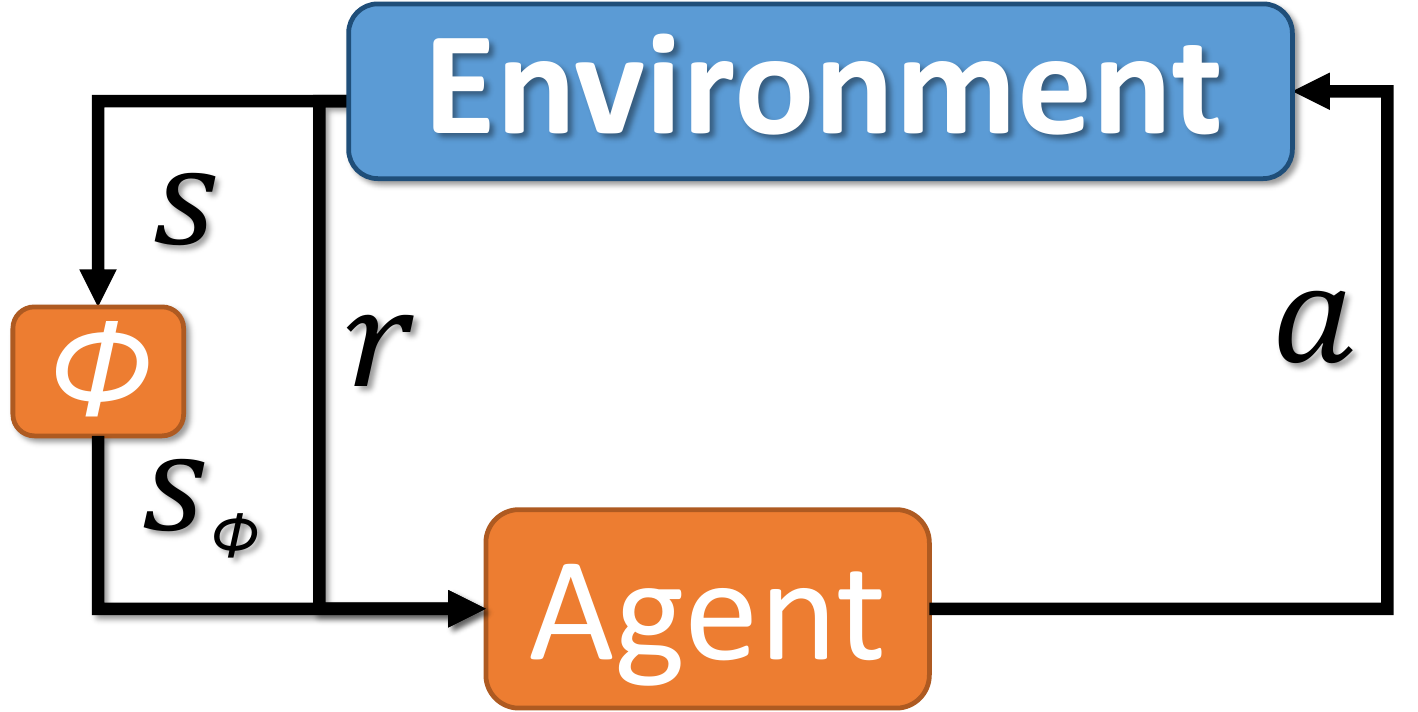}
\caption{Depiction of an Abstract MDP, that includes a mapping 
$\phi : \states \rightarrow \states_{\phi}$ from the full state 
$\state$ to an abstract state $\state_\phi$.}
\label{fig:state_abstraction}
\end{figure}

\cite{abel2019state} identify various types of abstraction discussed in the 
literature that can involve states and also actions, 
including: 
state abstraction~\citep{pmlr-v80-abel18a},
temporal abstraction~\citep{precup2000temporal},
state-action abstractions~\citep{pmlr-v108-abel20a},
and hierarchical \RL approaches~\citep{kulkarni2016hierarchical,dietterich2000overview}.
In this section we shall focus our attention on state abstraction.
%
%
Formally, given a high-dimensional state space $\states$, the goal of state abstraction 
is to implement a mapping $\phi : \states \rightarrow \states_{\phi}$ from each
state $\state \in \states$ to an abstract state $\state_{\phi} \in \states_{\phi}$, 
where $\rvert\states_{\phi}\rvert \ll \rvert\states\rvert$~\citep{abel2019theory}, 
and 
$\phi$ is an encoder~\citep{abel2019theory}.
This expands the set of RL problem definitions defined in \autoref{sec:background},
introducing the notion of an Abstract MDP, as illustrated in \autoref{fig:state_abstraction}.

In practice low dimensional representation are often obtained using
Variational AutoEncoder (VAE) based architectures~\citep{kingma2013auto}.
For example, \cite{9287851} apply 
neural discrete representation learning, mapping high-dimensional raw video
observations from an \RL agent's interactions with the
environment to a low dimensional discrete latent representation,
using a Vector Quantized AutoEncoder (VQ-AE) trained to reconstruct the
raw video data. 
%
%
The benefits of the approach are demonstrated within a 3D navigation task
in a maze environment constructed in Minecraft. 

The work from \cite{9287851} and others~\citep{tang2017exploration,burden2018using,burden2021latent}  
demonstrates the ability of state abstraction
to reduce noise from raw high-dimensional inputs.
However, discarding too much information can result in the encoder
failing to preserve essential features. 
Therefore, encoders
must find a balance between appropriate degree of compression 
and adequate representational power~\citep{abel2016near}. 
Using \emph{apprenticeship learning}, where the availability of 
an expert demonstrator providing a policy $\pi_E$ is assumed, 
\cite{abel2019state} seek to understand the role of 
information-theoretic compression in state abstraction for
sequential decision making. 
The authors draw parallels between 
state-abstraction for \RL and compression as understood in information theory.
The work focuses on evaluating the extent to which an
agent can perform on par with a demonstrator, while using as little
(encoded) information as possible.
Studying this property resulted in a novel objective function with which
a VAE~\citep{kingma2013auto} can be optimized,
enabling a convergent algorithm
for computing latent embeddings with a
trade-off between compression and value.

\cite{pmlr-v97-gelada19a}
and \cite{zhanglearning}
observe that encoder-decoder approaches are typically task agnostic -- 
encodings represent all dynamic elements that they observe,
even those which are not relevant. 
An idealised encoder, meanwhile, would learn a robust 
representation that maps two observations to the same 
point in the latent space while ignoring irrelevant 
objects that are of no consequence to our learning 
agent(s). 
%
Both works rely on the concept of bisimulation to avoid training
a decoder. 
The intuition behind bisimulation is as follows.
\begin{definitionsec}[Bisimulation.] 
Given an MDP $\M$, 
an equivalence relation $B$ between states is a bisimulation relation if, 
for all states $\state_i, \state_j \in \states$ that are equivalent
under B (denoted $\state_i \equiv_{B} \state_j$ ) the following conditions hold:
\begin{equation}
\R(\state_i, \action) = \R(\state_j, \action), \forall \action \in \actions,
\end{equation}
\begin{equation}
\Prob(G\rvert\state_i, \action) = \Prob(G\rvert\state_j, \action), \forall \action \in \actions, \forall G \in \states_{B}, 
\end{equation}
where $\states_{B}$ is the partition of $\states$ under the relation $B$
(the set of all groups $G$ of equivalent states), and
$\Prob(G\rvert\state, \action) = \sum_{\state' \in G} \Prob(\state'\rvert\state,\action)$.
\end{definitionsec}

%
\cite{zhanglearning}
propose \emph{deep bisimulation for control} (DBC) that learns directly 
on a bisimulation distance metric. 
This allows
the learning of invariant representations that can be used effectively for 
downstream control policies, and are invariant with respect to 
task-irrelevant details. 
The encoders are trained in a manner such that distances in
latent space equal bisimulation distances in the actual state space.
The authors evaluated their approach on visual Multi-Joint dynamics with Contact 
(MuJoCo) tasks where
control policies must be learnt from natural videos with moving
distractors in the background. 
Exactly partitioning states with bisimulation is generally not
feasible when dealing with a continuous state space, therefore 
a pseudometric 
space $(\states, d)$ is utilized, where distance function
$d : \states \times \states \rightarrow \rats_{\geq 0}$ measures the similarity 
between two states. 
%
%
%
DBC significantly outperforms SAC and DeepMDP~\citep{gelada2019deepmdp}
on CARLA, an open-source simulator for autonomous driving 
research~\footnote{\url{https://carla.org/}}.
%
%
While DBC was applied to image data, in principle it could 
also be applied to observations from \aco domains. 
%
%

\update{To the best of our knowledge the literature currently lacks
an evaluation of state abstraction approaches on \aco environments.  
Therefore, evaluating the methods discussed in this 
section on the \aco environments listed in \autoref{sec:envs} represents 
a research topic that is ripe for exploration. 
Here, a particularly tantalizing avenue is exploring the extent to which
these methods can cope with the \aco challenges of 
variable network sizes and topologies.}

\update{In \autoref{sec:function_approximators} we observe 
that an idealised DRL-ACD agent is implemented with
feature extractors that can efficiently deal with
this challenge.
Given that the above approaches frequently turn to VAEs
and similar architectures to obtain compact representations,
Variational Graph Auto Encoders (VGAE) represent a promising
solution~\citep{kipf2016variational,Park_2019_ICCV}.
However, the selection of an appropriate VGAE architecture
will require considerations regarding the extent to which
it can mitigate two well documented graph machine learning pathologies:
i.) \emph{Over smoothing}, 
where adding too many layers to the GNN, 
in the absence of an attention mechanism, 
can result in the same information being aggregated on all nodes, and; 
ii.) \emph{Over-squashing}, where node features are insensitive 
to information contained in distant nodes~\citep{rusch2023survey,giraldo2023trade}.
For the latter pathology, the graph topology plays the greatest role, 
with over-squashing occurring between nodes with a high commute time~\citep{di2023over}.
Here, the GNN literature provides examples of VGAE approaches designed
to mitigate these pathologies, including \cite{xu2024hc}'s Hierarchical Cluster-based-GAE 
and \cite{liu2023graph}'s Graph Positional and Structural Encoder.
We also recommend the consideration of solution concepts from the graph machine learning
literature for mitigating these challenges, \eg graph rewiring~\citep{di2023over}.}

\subsection{Exploration} \label{sec:exploration}

Successful state abstraction has numerous applications, including
the scaling of principled exploration strategies to DRL~\citep{tang2017exploration}  
and potential-based reward shaping~\citep{burden2021latent,burden2018using}. 
\update{The former, being able to efficiently discover the structure of a system,
has been identified as a key requirement for implementing both \aco and penetration
testing agents~\citep{zennaro2023modelling,tran2021deep,tran2022cascaded,zhou2021autonomous,selim2023adaptive}.}
%
%
However, learning an abstract state space that
meets all the desired criteria, \update{to enable efficient exploration},
remains a long-standing problem.
\RL agents often gradually unlock new 
abilities, that in turn result in new areas
of the environment being visited.
Many initial encodings may be learned before an 
agent has sufficiently explored the state space~\citep{pmlr-v119-misra20a}.
In addition, exploration is intractable for domains suffering from the 
\cod. 
Principled exploration strategies are required 
that enable the sufficient 
visitation of abstracted states
~\citep{burden2021latent,pmlr-v119-misra20a,wong2022deep}.
%
%
%
To address this, \cite{pmlr-v119-misra20a} introduce \HOMER,
a state abstraction approach that accounts for the fact
that the learning of a compact representation for states
requires comprehensive information from the environment
- something that cannot be achieved via random exploration
alone. 
%
%

\begin{figure}[h]
\centering
\includegraphics[width=\columnwidth]{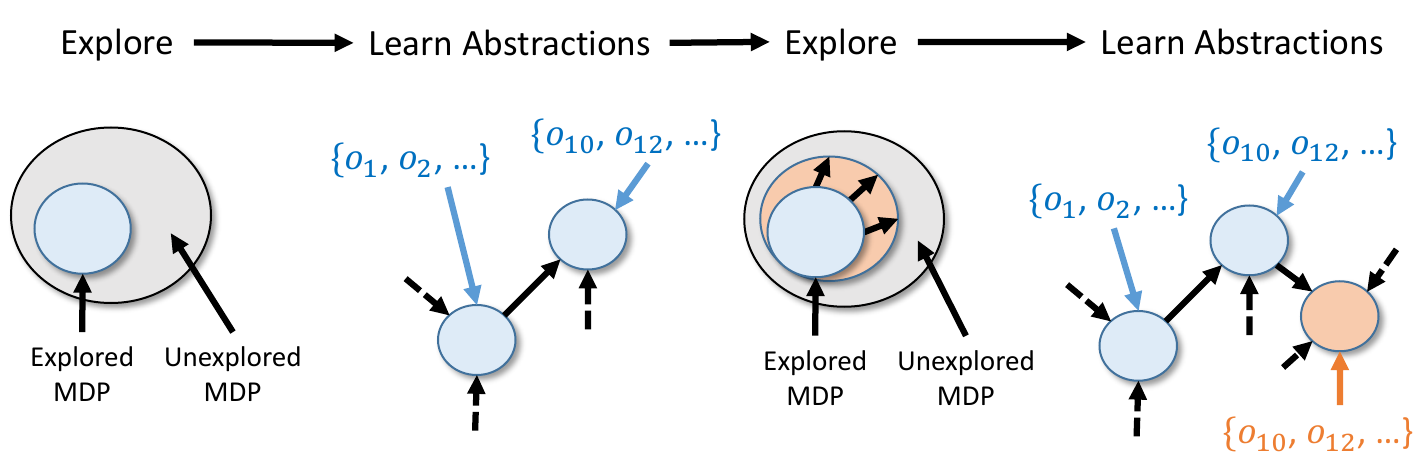}
\caption{\HOMER (Adapted from~\citep{pmlr-v119-misra20a}).}
\label{fig:homer}
\end{figure}

\HOMER is designed to learn a reward-free
state abstraction termed \emph{kinematic inseparability},
aggregating observations that share the same forward and backward
dynamics. 
The approach iteratively explores the environment by training policies to
visit each kinematically inseparable abstract state. 
Policies are constructed using contextual bandits and a synthetic 
reward function that incentifies agents to reach an abstract state. 
In addition, \HOMER interleaves learning the state
abstraction and the policies for reaching the new 
abstract states in an inductive manner, meaning
policies reach new states, which are abstracted, and
then new policies are learned, iteratively, until a \emph{policy cover}
has been obtained.
This iterative learning approach is depicted in \autoref{fig:homer}.
Once \HOMER is trained a near-optimal policy can be found for any reward
function.
\HOMER outperforms PPO and other baselines on an environment named
the \emph{diabolical combination lock}, a class of 
rich observation MDPs where the wrong choice leads to states
from which an optimal return is impossible.

There are numerous \emph{count-based} exploration approaches
with strong convergence guarantees for tabular \RL when applied
to small discrete Markov decision processes~\citep{tang2017exploration}.
%
%
\cite{ladosz2022exploration} define three desirable criteria for
exploration methods, including:
i.) determining the degree of exploration based on the agent’s learning;
ii.) encouraging actions that are likely to result in new outcomes, and;
iii.) rewarding the agent for exploring environments with sparse rewards. 
A popular approach towards encouraging exploration is via intrinsic rewards,
where the reward signal consists of extrinsic and
intrinsic components. 
%
When combined with state-abstraction,
these tried and tested methods can be applied
to environments suffering from the \cod.

\cite{tang2017exploration} introduce a count-based exploration 
method through static hashing, using SimHash.
The hash codes are obtained via a trained AutoEncoder, and provide a means
through which to keep track of the number of times semantically similar
observation-action pairs have been encountered. 
A count-based reward encourages the visitation of less 
frequently explored semantically similar 
observation-action pairs.
\cite{bellemare2016unifying} proposed 
using Pseudo-Counts, counting salient events derived from 
the log-probability improvement according to a 
\emph{sequential density model} over the state space. 
In the limit this converges to the empirical count.
\cite{martin2017count}
focus on counts within the feature 
representation space rather than for the raw inputs.
Other approaches for computing intrinsic rewards are based
on prediction errors 
~\citep{pathak2017curiosity,stadie2015incentivizing,savinovepisodic,burdaexploration,bougie2021fast,ladosz2022exploration}
and memory based methods~\citep{fu2017ex2,badianever},
using models trained to distinguish
states from one another, where easy to distinguish states are 
considered novel~\citep{ladosz2022exploration}.
\begin{figure}[h]
\centering
\includegraphics[width=0.8\columnwidth]{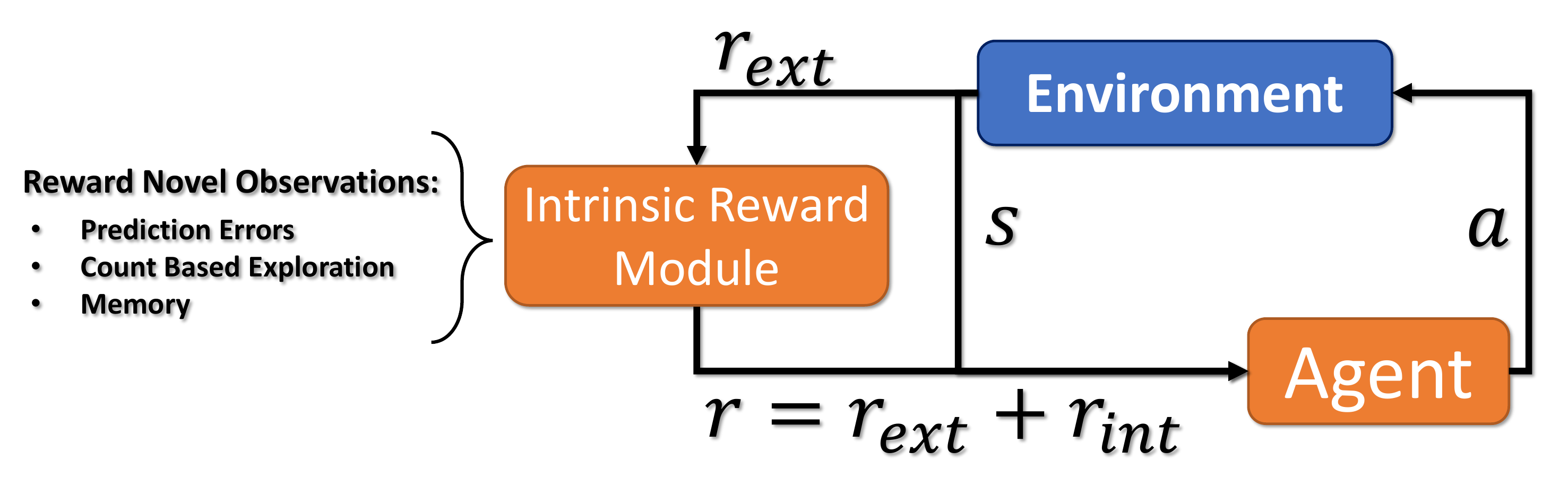}
\caption{Approaches rewarding agents for visiting novel states
(Adapted from~\citep{ladosz2022exploration}).}
\label{fig:reward_novel_states}
\end{figure}

Goal based exploration represents another class of
methods, which \cite{ladosz2022exploration} further divide into: 
meta-controllers, where a controller with a high-level overview of the environment provides goals for a worker agent~\citep{forestier2017intrinsically,colas2019curious,vezhnevets2017feudal,hester2013learning,kulkarni2016hierarchical};
sub-goals, finding a sub-goal for agents to reach, \eg bottlenecks in the environment~\citep{machado2017laplacian,machadoeigenoption,fangadaptive},
and; goals in the region of highest uncertainty, where exploring uncertain states with respect to the rewards are the sub-goals~\citep{kovac2020grimgep}.

Despite the above advances, learning over the entirety of the environment
is neither feasible nor desirable when dealing with increased complexity. 
Here principled methods are required that can determine
which parts of the state space are most relevant~\citep{pmlr-v134-perdomo21a}.
However, this requirement in itself leads to the dilemma of 
how one can determine with minimal effort that an area of the state space
is irrelevant, which is an open research question. 

%
%

\subsection{Knowledge Retention} \label{sec:knowledge_retention}

\update{A recent systematic review by \cite{kreutz2024impact} on the  
impact of AI on enterprise information security management 
highlighted the 
need for human oversight, in particular for safeguarding
against a well documented limitation of deep learning 
in online settings: \emph{catastrophic forgetting}, the unlearning 
of previously acquired knowledge~\citep{atkinson2021pseudo,schak2019study}.
As with many online learning tasks, DRL agents are
also prone to catastrophic forgetting.  
In order to be sample efficient, DRL approaches
often resort to experience replay memories, which 
store experience transition tuples that are sampled
during training~\citep{foerster2017stabilising}. 
However, as the state space can grow exponentially with the size 
of the state representation~\citep{burden2021latent},
maintaining a sufficient sample spread over the 
state-action space becomes challenging~\citep{de2015importance}.
This challenge is with respect to both gathering a sufficient
spread (as discussed in Section \ref{sec:exploration}) and 
the need for principled data retention strategies.}

\update{For current commonly used DRL solutions, samples are either stored medium term, as in off-policy approaches such as DQN and DDPG,
or short term, \eg  samples gathered using multiple workers for PPO, 
an approach where samples gathered on-policy are discarded after performing 
several epochs of optimization~\citep{schulman2017proximal}. 
For the former, paying attention to the experience replay memory composition
can mitigate catastrophic forgetting~\citep{de2015importance}.
However, in practice, a large number of transitions are discarded, 
since there is a memory cost associated with storage. 
An additional challenge is that the stationarity of the environment,
and one's teammates, opponent(s), and even green agents, are a strong assumption.
%
Samples stored inside a replay buffer can become deprecated,
confronting learners with the same challenges seen in data streaming~\citep{hernandez2018multiagent,omidshafiei2017deep,palmer2020independent,palmer2018negative,palmer2018lenient}. 
Here DRL agents require continual learning, 
the ability to continuously learn and build on previously 
acquired knowledge~\citep{wolczyk2021continual}.}

One approach to solve this problem is to utilize a \emph{dual memory}
where a freshly initialized DRL agent, a 
short-term agent (network), is trained on a new task, upon which
knowledge is transferred to a DQN designed to retain long-term
knowledge from previous tasks. 
A generative network is used to generate short sequences from previous
tasks for the DQN to train on, in order to prevent catastrophic forgetting
as the new task is learned~\citep{atkinson2021pseudo}. 
However, this approach relies on a stationary \env, as an additional
mechanism would be required to determine the relevance of past knowledge,
given drift in the state transition probabilities.  

\emph{Elastic Weight Consolidation} (EWC) is another popular approach for mitigating catastrophic
forgetting for DNNs~\citep{kirkpatrick2017overcoming,huszar2018note,huszar2017quadratic}.
EWC has been applied to DRL using an 
additional loss term using the Fisher information matrix
for the difference between the old and new parameters, 
and a hyperparameter $\lambda$ which can be used to specify
how important older weights are~\citep{ribeiro2019multi,nguyen2017system,kessler2022same,wolczyk2021continual}.
 
%

\subsection{\update{High-Dimensional State Approaches Summary}}

\update{To enable the timely discovery, tracking, identification, and mitigation of
Red agents, our idealised autonomous cyber-defence agent must be capable of efficiently 
gathering and processing data from a high-dimensional state space, resulting from the
 combinatorial explosion of monitoring properties on each host within a network.
In this section we observe that,
even when our idealised DRL agent can benefit from a system-of-systems approach,
e.g., where the task of intruder detection and the identification of compromised
nodes and services is outsourced to other models, our cyber-defence agent
will remain confronted with a high-dimensional observation space.
To address this challenge, 
in this section we have surveyed DRL methodologies designed for domains
that push traditional DRL approaches, and their DNN function approximators,
to their limits. 
These, therefore have much potential for the domain of cyber-defence.}

\update{We outlined immediate next steps that the autonomous cyber-defence
community can take to evaluate the benefits of utilizing and fusing the identified approaches, 
while also predicting the challenges that will likely need to be overcome.
In particular, we discussed that our idealised cyber-defence agent can benefit from 
an appropriate observation representation, such as using 
a graph-based representation and leveraging graph neural networks (GNNs),
combined with 
state abstraction methods
and principled exploration approaches.
However, we observe that a number of graph machine learning pathologies
will need to be considered for our learners 
to reap the full benefit from this approach.} 

\update{We have seen that DRL agents for domains with a large state 
spaces face significant challenges \wrt learning encodings, requiring an
iterative approach in order to learn embeddings for 
areas of an environment \say{unlocked} thanks to learning new behaviour~\citep{pmlr-v119-misra20a}.
Given that our idealised \aco-DRL agent is situated within an adversarial game, 
considering the implications of using approaches like \HOMER~\citep{pmlr-v119-misra20a} 
in an adversarial
learning context represents an interesting direction for future research.
Finally, in this section we also discuss the knowledge retention challenge, 
something that is critical within adversarial settings, to prevent cyclic 
behaviour and forgetting a best response policy towards
a previously encountered opponent policy.
We shall revisit this challenge when we discuss the topic of adversarial
learning in \autoref{sec:adv_learning}.
A summary of the approaches discussed in this section is provided in 
\autoref{tab:state_abstraction} in \autoref{appendix:states}.}

\section{Approaches for combinatorial action spaces} \label{sec:hdas_approaches}

Due to an explosion in the number of state-action pairs, traditional
DRL approaches do not scale to 
high-dimensional combinatorial action spaces. 
Scalable methods will need to meet the following criteria.
%
%
%
%
%
%

\spacer
\noindent \textbf{\generalizability:} 
For our target domains a sufficient visitation of 
all state-action pairs to obtain accurate value 
estimates is intractable. 
Formulations are required that allow for 
generalization 
over the action space~\citep{dulac2015deep}. 
\update{For instance, once deployed, a cyber-defence policy 
must be able to execute action and target-host
combinations that were not executed during 
training.}

\spacer
\noindent \textbf{\unseenactions} can result in a policy 
being trained on a subset of actions $\actions' \subset \actions$.
This has implications when the agent is later asked to choose from 
a larger set of actions~\citep{9507301} or applying actions to previously
unseen objects~\citep{chandak2020lifelong,fang2020learning,9507301},
\eg a new host on a network for \aco.

\spacer
\noindent \textbf{Computational Complexity}: 
An efficient formulation is to use a DNN with $\rvert\actions\rvert$ output 
nodes, requiring a single forward pass to compute an output for each action.
However, this approach will not generalize well. 
Alternatively, for a value function with \emph{a single output}, 
one could input observation-action pairs and estimate the 
utility of an arbitrary number of actions:
$Q : \observationf \times \actions \rightarrow \rats$. 
However, this approach is intractable due to the 
computational cost growing linearly with $\rvert\actions\rvert$.
Instead, methods with sub-linear complexity are required.

\spacer

\begin{figure}[h]
\centering
\subfloat[Proto Actions]{\label{fig:actions:proto}\includegraphics[width=0.2\columnwidth]{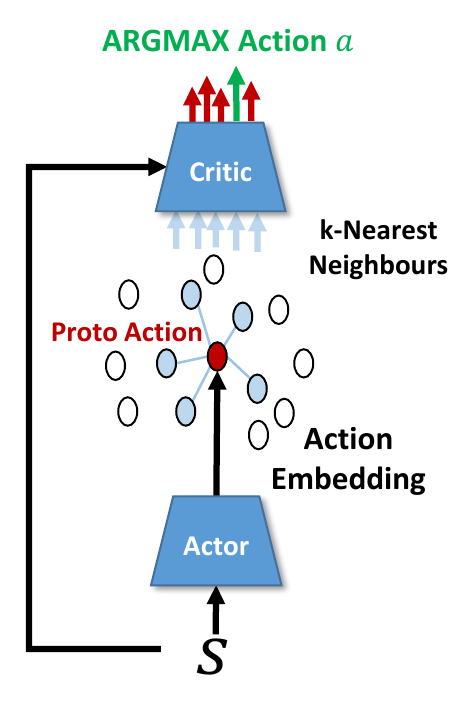}}
\subfloat[Action Decomp.]{\label{fig:actions:composition}\includegraphics[width=0.2\columnwidth]{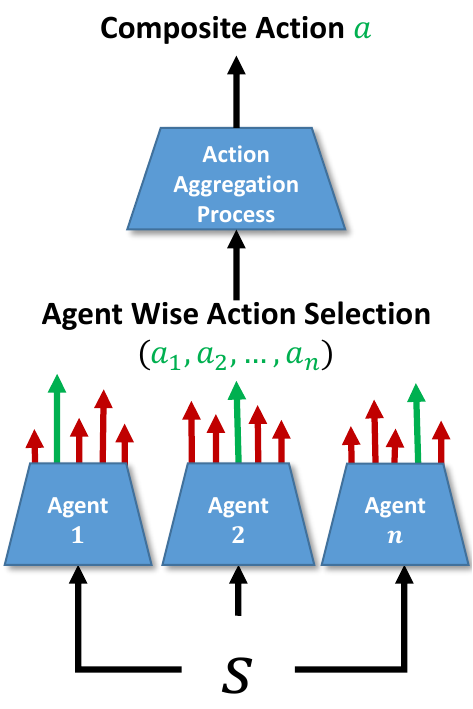}}
\subfloat[Action Elimination]{\label{fig:actions:elimination}\includegraphics[width=0.2\columnwidth]{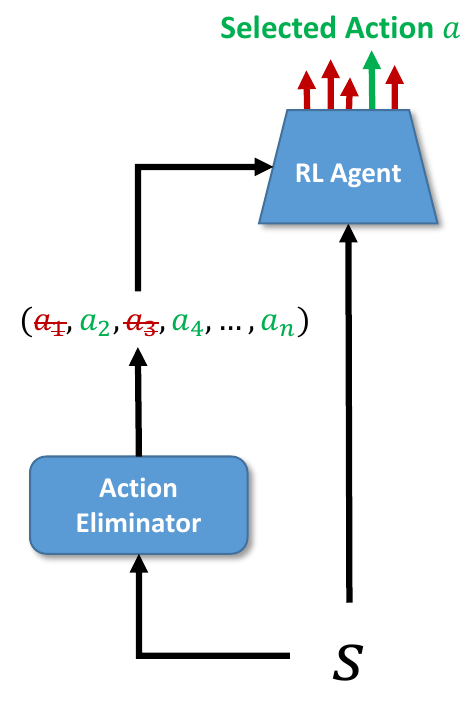}}
\subfloat[Hierarchical]{\label{fig:actions:hierarchcial}\includegraphics[width=0.2\columnwidth]{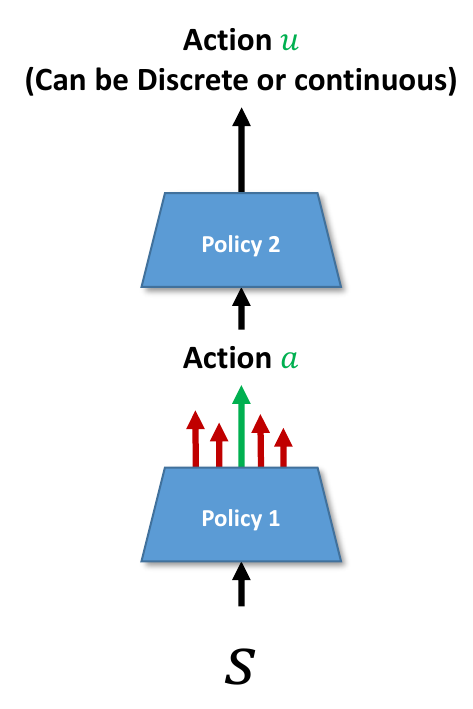}}
\subfloat[Curriculum]{\label{fig:actions:curriculum}\includegraphics[width=0.2\columnwidth]{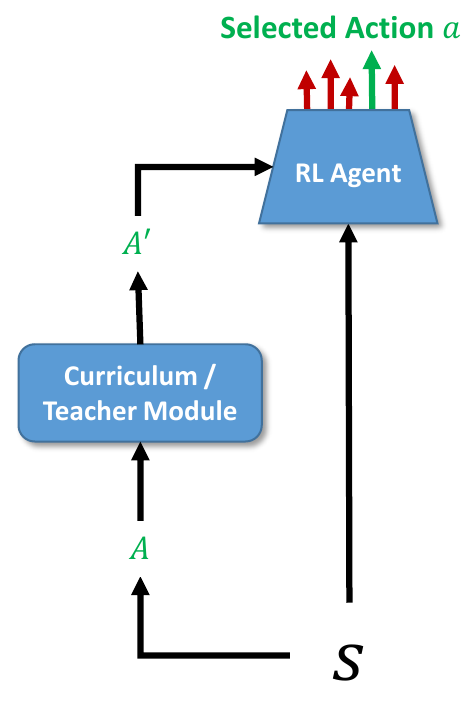}}
\caption{Categories of DRL approaches for high-dimensional action spaces.
\autoref{fig:actions:proto}: Proto actions leverage prior domain knowledge and embed actions into a continuous space, before applying $k$-NN
to pick the closest discrete actions, which are passed to a critic.
\autoref{fig:actions:composition}: Action Decomposition reformulates the single agent problem as a Dec-POMDP with a composite action space.
\autoref{fig:actions:elimination}: Action elimination approaches use a module that determines which actions are redundant for a given observation. 
\autoref{fig:actions:hierarchcial}: Hierarchical, the action selected by a policy influences the action selected by a sub-policy.
\autoref{fig:actions:curriculum}: Curriculum learning approaches for gradually increasing the number of available actions.}
\label{fig:high_level_action_approaches_categories}
\end{figure}

The criteria listed above provide the axes along which 
the suitability of approaches for autonomous cyber defence
can be measured. 
Through reviewing the literature on high-dimensional
action spaces we were able to identify five categories 
that conveniently cluster the approaches: 
i.) \emph{proto action} based approaches (\autoref{sec:hdas_approaches:wolpertinger});
ii.) \emph{action decomposition} (\autoref{sec:action_decomposition}); 
iii.) \emph{action elimination} (\autoref{sec:action_elimination}); 
iv.) \emph{hierarchical} approaches (\autoref{sec:hrl}), and; 
v.) \emph{curriculum learning} (\autoref{sec:actions:cl}).
\autoref{fig:high_level_action_approaches_categories} provides an 
illustrative example and short description for each category.
\update{As we shall see, action decomposition approaches often rely
on solutions from the Multi-Agent Deep Reinforcement Learning (MADRL)
literature.
Furthermore,
recent CAGE Challenges (3 \& 4) require 
decentralized cyber-defence solutions using MADRL. 
Therefore, before discussing action decomposition approaches in \autoref{sec:action_decomposition},
we will discuss relevant MADRL solutions in \autoref{sec:hdas_approaches:marl_and_coordination}.}
In \autoref{tab:high_dim_action_approaches} in \autoref{appendix:action_approaches} we provide an overview 
of the literature and a short contributions summary.
%

\subsection{Proto Action Approaches} \label{sec:hdas_approaches:wolpertinger}

\cite{dulac2015deep} proposed the first DRL
approach to address the \hdasp, the \WP architecture,
which embeds discrete actions into a continuous space~$\rats^n$.
A continuous control policy 
$f_{\pi} : \observationf \rightarrow \rats^n$,
is trained to output a \emph{proto} action.
Given that a proto action $\hat{\action}$ is unlikely 
to be a valid action ($\hat{\action} \notin \actions$),
$k$-nearest-neighbours ($k$-NN) is used to map the proto action 
to the $k$ closest valid actions:
$g_k(\hat{\action}) = \argmin_{\action \in \actions}^k \rvert\action-\hat{\action}\rvert_2$.
To avoid picking outlier actions, and to refine the action selection,
the selected actions are passed to a critic $Q$, which then selects
the $\argmax$. 
%

The \WP architecture meets many of the above requirements. 
While the time-complexity scales 
linearly with the number of actions $k$, 
the authors show both theoretically and in practice
that there is a point at which increasing $k$ delivers a 
marginal performance increase at best. 
Using 5-10\% of the maximal number of actions was found
to be sufficient, allowing agents to generalize over the set of 
actions with sub-linear complexity.
Here, the action embedding space does require a logical 
ordering of the actions along each axis. 
Currently prior information about the action space is
leveraged to construct the embedding space.
However, \cite{dulac2015deep} note that learning action 
representations during training could also provide a solution. 
The approach has also been criticised for 
instability during training due to 
the $k$-NN component preventing the gradients
from propagating back to boost the training of 
the actor network~\citep{tran2022cascaded}.

\begin{figure}
\centering
\includegraphics[width=0.7\columnwidth]{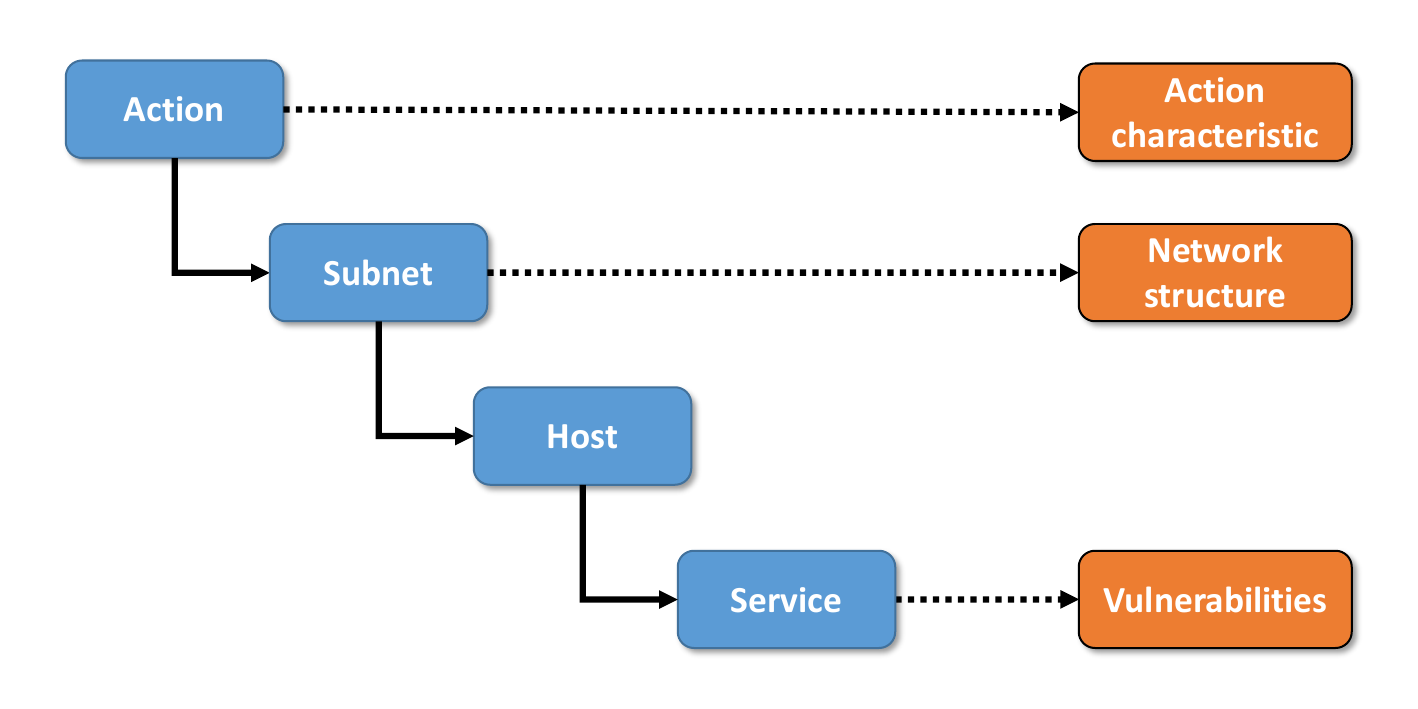}
\caption{\WP attack embedding used by \cite{nguyen2020multiple} on NASim~\citep{schwartz2019autonomous}.}
\label{fig:NASim_attack_embedding}
\end{figure}

\update{To date work on the applicability of the \WP approach to cyber-defence
scenarios include using it for penetration testing on NASim~\citep{nguyen2020multiple}.}
%
%
\cite{nguyen2020multiple}
evaluate the ability of
\WP to learn a policy that 
launches attacks on vulnerable services on
a network.
\WP is applied using an embedding space that consists of three levels (illustrated in \autoref{fig:NASim_attack_embedding}):
i.) the action characteristics (\emph{scan subnet}, \emph{scan host} and \emph{exploit services}); 
ii.) the subnet to target, and;
iii.) services that are vulnerable towards attacks.
The second dimension focuses on the destination of the action with respect 
to the subnet, \eg selecting \say{scan subnet} along axis $1$ and selecting the subnet on
axis $2$.  
\emph{Node2Vec} is used for expressing the network structure, and the authors also
train a network to produce similar embeddings for correlated service vulnerabilities. 
%
%
%
%
\WP was shown to outperform DQN on NASim~\citep{schwartz2019autonomous}.

\update{While the work from \cite{nguyen2020multiple} focused on cyber-attacking 
agents, there are nevertheless valuable takeaways
for using the approach for the cyber-defence task.
For instance, when training \WP to generalize across different network
topologies, the computational efficiency for computing actions
needs to be considered alongside the computational complexity of
running Node2Vec every time the network topology changes. 
This problem is amplified when applying \WP to dynamic networks
such as mobile ad-hoc networks.}

\update{Upon applying \WP for the task of 
energy efficient task scheduling in data centres,
\cite{9522309} noted the algorithm's exploration method struggles with 
sampling a sufficient number of servers early in training. 
This causes the critic to mistakenly overestimate a subset of servers,
resulting in the actor selecting from a limited number of servers later 
in training and once deployed. 
Therefore, while the \WP approach meets many of desirable criteria for
addressing our idealized \aco's agents need for an HD-action approach,
numerous challenges remain with respect to obtaining a computationally
efficient embedding space for \aco and an advanced exploration policy.}

\subsection{Multi-Agent Learning Approaches} \label{sec:hdas_approaches:marl_and_coordination}

\update{
In this section we shall discuss suitable methods for tasks that require
a decentralized cyber-defence solution, such as the problem formulations
from CybORG CAGE Challenges 3 and 4.
Here, approaches are required that can consistently converge upon, or at least approach, a Pareto
optimal solution~\citep{JMLR:v17:15-417,lauer2000algorithm,kapetanakis2002reinforcement,palmer2020independent}.
However, this task is non-trivial, due to a number of well documented multi-agent learning pathologies (See \autoref{sec:background:marl_pathologies}). 
Nevertheless, fully cooperative MADRL is an active field with a plethora of solutions designed to overcome the above learning pathologies. 
For comprehensive overview of existing methods in this area, we recommend the following surveys~\citep{gronauer2022multi,hernandez2018multiagent,nguyen2020deep,oroojlooy2023review,wong2023deep}.
In this section we provide a high-level overview of applicable approaches, 
and a summary of the extent to which they have been utilized within the \aco-DRL literature.}

\update{Popular MADRL classes include:
\begin{itemize}
\item{{\textbf{Independent Learning (IL):}}} 
Independent learners each learn through gathering their own experiences 
and treating other agents as part of the environment~\citep{palmer2020independent,JMLR:v17:15-417,lauer2000algorithm}. 
There are a number of IL versions of popular DRL approaches, 
\eg Independent Q-Learning (IQL), 
Independent synchronous Advantage Actor-Critic (IA2C) 
and Independent Proximal Policy Optimisation (IPPO)~\citep{papoudakis1benchmarking}. 
For independent Deep Q-learning approaches, 
the non-stationarity challenge is amplified 
due to a use of large experience replay memories.
While policy gradient methods frequently discard training data,
deep IQL approaches typically store data long term.
This has motivated the development of methods that reduce the
impact of deprecated state transitions, \ie through using 
optimistic approaches~\citep{omidshafiei2017deep,palmer2018negative,palmer2018lenient}, 
agents sampling \emph{concurrent experience replay trajectories} (CERTs) \citep{omidshafiei2017deep}
and sample fingerprinting~\citep{foerster2017stabilising}. 
\item{{\textbf{Centralized Multi-Agent Policy Gradient:}} Here, agents benefit from a CTDE approach that uses a centralized critic during training. 
The centralized critic is optimized using information from all actors~\citep{papoudakis1benchmarking}.
Approaches include 
Counterfactual Multi-Agent (COMA) Policy Gradient~\citep{foerster2018counterfactual};
Multi-Agent A2C (MAA2C) is an actor-critic algorithm in which the critic learns a joint
state value function, conditioned on both states and actions~\citep{li2023research}, and;
Multi-Agent PPO (MAPPO), an extension of IPPO with a joint state-action value function as for MAA2C~\citep{yu2022surprising}.
Thanks to being a PPO variant, the latter can perform multiple update epochs per training batch, and as a result is more
data efficient compared to MAA2C~\citep{papoudakis1benchmarking}.}  
\item{{\textbf{Value Decomposition Approaches:}} This class of CTDE MADRL algorithms decomposes  
a joint state-action value function into individual state-action value functions. 
Popular variations include
Value Decomposition Networks (VDN), where each agent maintains its own 
state-action value network. 
The VDN decomposes the joint Q-values into the sum of individual Q-values~\citep{sunehag2017value}.
During training the gradients of the joint temporal difference error flow back to the network of each agent.
QMIX is an extended version of VDN that enables a more complex decomposition via a parameterized mixing network 
that utilizes a hyper-network~\citep{rashid2018qmix}.}
\item{{\textbf{Communicating Agents:}} This set of agents are implemented with the ability to communicate at deployment time. Examples include fully differentiable communication solutions, such as a 
Differentiable Inter Agent Learning (DIAL) \citep{foerster2016learning}, and the use of graph structured communication enabled by GNNs~\citep{chen2021graph,su2020counterfactual,xiao2023graph}.}
\end{itemize}}

\update{
A number of the above approaches have been evaluated within the context of cyber-defence.
\cite{hicks2023canaries} train a single neural policy shared by multiple agents within CybORG CAGE Challenge 3.
The authors propose a curriculum learning approach for addressing the non-stationarity pathology, 
initially pairing a single DRL agent with a set of competent rules-based agents, 
and gradually increasing the proportion of learning cyber-defence agents.}

\update{\cite{contractor2024learning} demonstrated that DIAL can enable cyber-defence agents to successfully communicate
with each other in response to incidents on a multi-agent adaption of CAGE Challenge 2. 
Communication was also found to facilitate coordinate in response to cyber threats.}

\update{\cite{cheah2023co} introduce a co-operative decision-making framework (Co-Decyber). 
This breaks up a big multi-contextual action space into smaller decisions. 
The authors successfully apply this framework to an autonomous vehicle platooning scenario, 
showing that Co-Decyber outperforms a random reference agents, also on a multi-agent version of CAGE Challenge 2.}

\update{\cite{wiebe2023learning} evaluated the ability of a team of cyber-defence agents to jointly mitigate an attacker, 
comparing the performance of agents trained with IQL and QMIX. 
The authors find that both were capable of learning various objectives across different 
network sizes, and perform robustly on dynamic problem spaces.  
Interestingly both IQL and QMIX performed similarly on the majority of evaluation scenarios, 
with the performance of IQL indicating that it would scale well to larger networks.}

\update{\cite{wilson2024multi} evaluate the use of MADRL for autonomous cyber defence decision-making on generic maritime based
cyber-defence scenario, finding that MAPPO outperforms IPPO, noting that a big improvement was achieved upon conducting
an extensive hyperparameter tuning.}

\update{\cite{TANG2024103871} use tools from game theory to model cyberspace conflicts as a Stackelberg hypergame. 
This enables a hierarchical MADRL approach for facilitating cooperation within multi-agent version 
of CAGE Challenge 2.}

\update{The above works demonstrate the potential of MADRL solutions for cyber-defence
scenarios that require a decentralized solution.
However, the plethora of available MADRL approaches that can be found
within the MADRL literature raises the question regarding which approach(es)
are most suitable.
Recently, \cite{papoudakis1benchmarking} conducted an extensive
benchmarking of MADRL approaches on cooperative tasks using non-cyber environments, including
IQL, IA2C, IPPO, COMA, MAA2C, MAPPO, VDN and QMIX. 
The authors performed hyperparameter optimisation separately for each approach in each environment.
Interesting takeaways included that: 
\begin{enumerate}
\item Despite ILs having a reputation for being inferior to CTDE
methods, the authors found that IA2C and IPPO can learn effective policies, in particular in environments where each agent
can independently learn policies that achieve high returns without requiring extensive coordination with other agents.
Therefore, IL approaches may very well be successful in decentralized cyber-defence scenarios where each agent takes
care of a network segment, without requiring a significant amount of interaction with other agents.
\item The availability of joint-information (observations and actions) over all agents, 
can significantly boost performance during training, in particular when the full state of the environment is not available.
For instance, MAA2C performs better than IA2C on such tasks.
\item Finally, while VDN and QMIX performed well on Multi-Agent Particle Environment (MPE) tasks, 
both perform significantly worse than policy gradient methods on sparse reward tasks. 
\end{enumerate}
The above findings indicate that interesting insights could be 
gained from using the work from \cite{papoudakis1benchmarking} as a template,
and conducting a similar extensive evaluation of MADRL approaches on CybORG
CAGE Challenges 3 and 4, or through converting other cyber-defence environments
into MADRL formulations, as done by~\cite{wiebe2023learning}.}

\subsection{Action Decomposition Approaches} \label{sec:action_decomposition}

\update{For action decomposition approaches the}
action space is decomposed into actions
provided by multiple agents, \eg
having each agent control an action dimension~\citep{tavakoli2018action},
or via an algebraic formulation for combining the actions~\citep{tran2022cascaded}.
However, learning an optimal policy requires the underlying agents to converge upon an \emph{optimal joint-policy}. 
Therefore, approaches must be viewed through the lens of MADRL within a Dec-POMDP,
using equilibrium concepts from multi-agent learning.
We shall now consider the different types of action decomposition approaches that
can be found in the literature.

\subsubsection{Branching Dueling Q-Network} \label{sec:BDQ}

\begin{figure}
\centering
\includegraphics[width=\columnwidth]{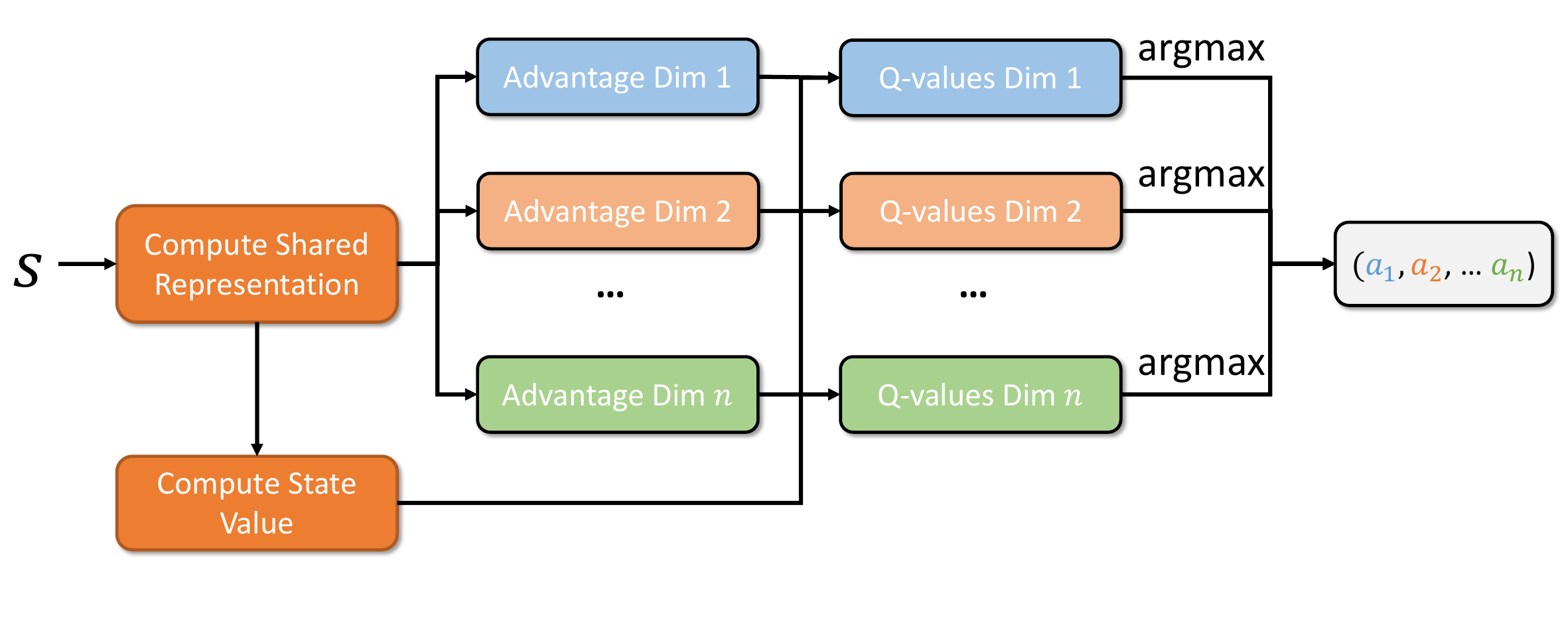}
\caption{An illustration Branching Dueling Q-Networks (adapted from~\citep{tavakoli2018action}).}
\label{fig:BDQ}
\end{figure}

Designed for environments where the action-space can be split 
into smaller action-spaces
Branching Dueling Q-Network (BDQ)~\citep{tavakoli2018action} 
is a branching version of Dueling DDQN~\citep{wang2016dueling}~\footnote{
Dueling DDQNs consist of two separate estimators for the 
value and state-dependent action advantage function.}.
%
%
Each branch of the network is responsible for proposing
a discrete action for an actuated joint. 
%
The approach features a \emph{shared decision module}, 
allowing the agents to learn a common latent representation
that is subsequently fed into each of the $n$
DNN branches, and can therefore be considered a CTDE approach. 
A conceptional illustration of the approach can be found 
in \autoref{fig:BDQ}. 

BDQ has a linear increase of network outputs
\wrt number of degrees of freedom, thereby allowing a level of
independence for each individual action dimension.
It does not suffer from the combinatorial growth 
of standard vanilla discrete action algorithms. 
%
%
%
However, the approach is designed for discretized continuous 
control domains. 
Therefore, BDQ's scalability to the MultiDiscrete
action spaces from \aco requires further investigation. 

While BDQ achieves sub-linear complexity, 
the formulation is vulnerable towards the MA(D)RL pathologies 
outlined above. 
To evaluate the benefit of the 
shared decision module, BDQ is evaluated against
Dueling-DQN, DDPG, and \emph{independent Dueling DDQNs} (IDQ). 
The only mentioned distinction between
BDQ and IDQ is that the first two
layers were not shared among IDQ agents~\citep{tavakoli2018action}.
BDQ uses a modified Prioritized Experience Replay memory~\citep{schaul2015prioritized}, 
where transitions are prioritized based on the \emph{aggregated distributed TD error}.
%
%
In essence, a prioritized version of Concurrent Experience Replay Trajectories (CERTS)
are being utilized, a method from the MADRL literature 
that has previously been shown to facilitate coordination~\citep{omidshafiei2017deep}.
\commentout{There is no explicit mention of any differences between
sampling techniques for BDQ and IDQ.
Specifically, for IDQ each agent would typically have
its own experience replay buffer $\mathcal{D}_i$, where
the use of CERTS cannot be assumed.
This raises the question, were CERTS primarily 
responsible for facilitating cooperation, 
rather than the shared representations?}

BDQ has been evaluated on numerous discretized MuJoCo environments~\citep{tavakoli2018action}. 
The evaluation focused on two axes: granularity and degrees 
of freedom. 
BDQ's benefits over Dueling-DQNs become noticeable 
as the number of degrees of freedom are increased.
In addition, BDQ was able to solve granular, high degree
of freedom problems for which Dueling-DDQNs was not applicable.
On the majority of the environments DDPG still outperformed
BDQ, with the exception of Humanoid-v1.
Unfortunately a comparison of BDQ 
against the \WP architecture was not
provided.
%
%
\update{Furthermore, to date the \aco literature lacks an evaluation of BDQ on both cyber-defence and attacking tasks.
Here, we observe that, for numerous cyber-defence environments listed in \autoref{sec:envs}
the action space can be decomposed into MultiDiscrete action signatures (\eg CAGE Challenge 2 and PrimAITE).
Therefore, conducting an extensive benchmarking of BDQ against other approaches listed in this section
could yield interesting insights.}

%
%
%
%
%
%
%
%
%

\subsubsection{Cascading Reinforcement Learning Agents} \label{sec:Cascading_RL}

\begin{figure}
\centering
\includegraphics[width=\columnwidth]{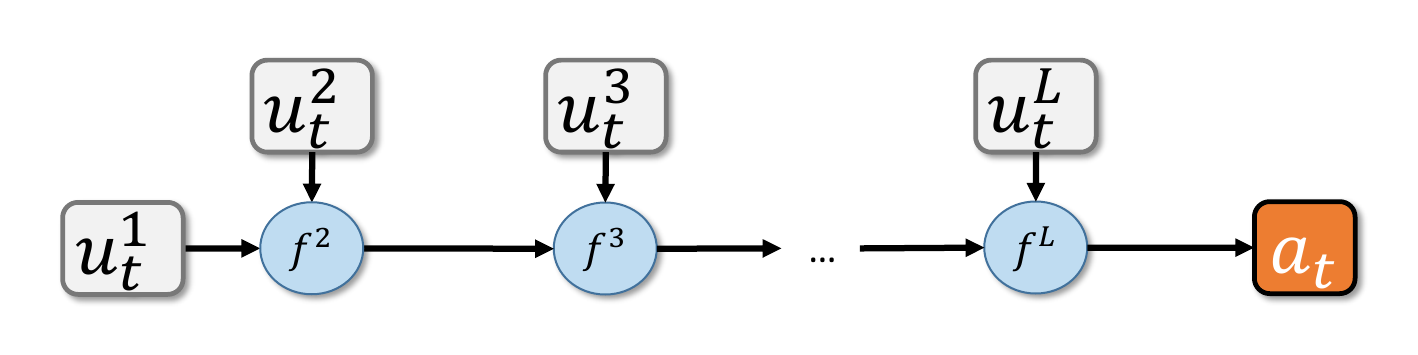}
\caption{Action space composition. Action $u_t$ at time
step $t$ is algebraically constructed using actions $a^i_t$ obtained
from action subsets $\mathcal{A}^i$ (adapted from ~\citep{tran2022cascaded}).}
\label{fig:cascading}
\end{figure}

For \emph{cascading reinforcement learning agents}~(CRLA) the action space $\actions$ is 
decomposed into smaller sets of actions 
$\crlaactions^1, \crlaactions^2, ..., \crlaactions^L$~\citep{tran2022cascaded}.
For each subset of actions $\mathcal{A}^i$, the size of the dimensionality
is significantly smaller than that of $\actions$, \ie $\rvert\crlaactions^i\rvert \ll\ \rvert\actions\rvert$ $(\forall i \in [1, L])$. 
In this formulation a primitive action $\action_t$ at time step $t$ is given by a 
function over actions $\crlaaction_t^i$ obtained from each respective subset $\crlaactions^i$: 
$\crlaaction_t = f(\crlaaction_t^1, \crlaaction_t^2, ..., \crlaaction_t^L)$.
The action components $\crlaaction_t^i$ are chained together to algebraically build 
an integer identifier. 
This provides a formulation through which larger
identifiers can be obtained using the smaller integer values provided
by each action subset. 
A CTDE approach is used to facilitate the training 
of $n$ agents, where the joint action space $\bm{\actions}$ 
is comprised of $\crlaactions^1, \crlaactions^2, ..., \crlaactions^n$, with $\crlaactions^i$ 
representing the action space of an agent $i$.
The concept is illustrated in \autoref{fig:cascading}. 

\begin{figure}[h]
\centering
\includegraphics[width=0.7\columnwidth]{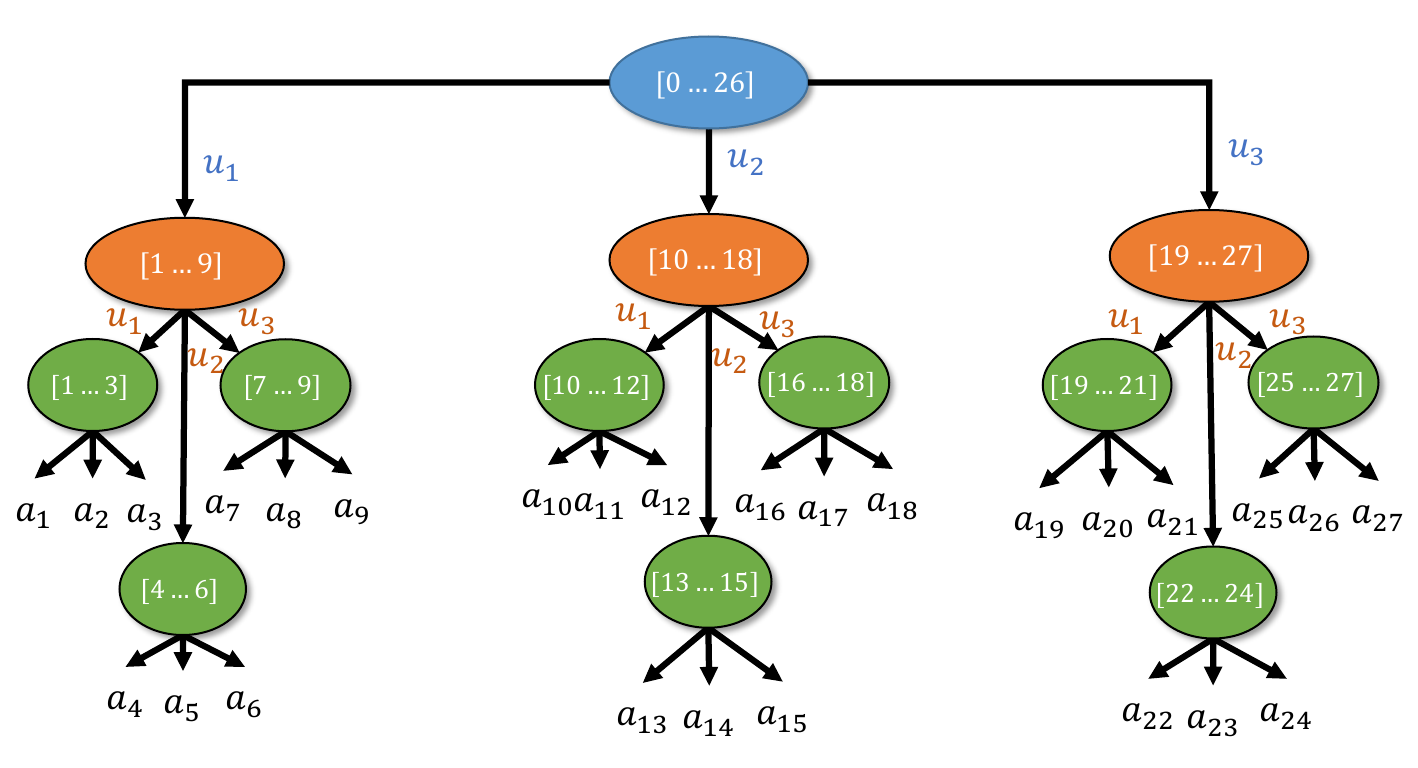}
\caption{An illustration of the action selection process used by CRLA 
(adapted from \citep{tran2022cascaded}). The leaf nodes represent the 
primitive actions from $\actions$, while each internal node contains
the action range for its children. Through using an algebraic 
formulation that makes use of an offset, only three agents with
an action space $\rvert\crlaactions^i\rvert=3$ are needed to capture $\actions$.}
\label{fig:cascading_selection_tree}
\end{figure}

CRLA yields a solution that allows for large combinatorial action spaces 
to be decomposed into a branching tree structure. 
Each node in the tree represents a decision by an agent regarding which 
child node to select next.
Each node has its own identifier, and all nodes in the tree have the same
branching factor, with a heuristic being used to determine the number of tree levels:
$\rvert T \rvert = \log_b(\rvert\actions\rvert)$.  
Instead of having an agent for each node, the authors propose
to have a linear algebraic function that shifts the action component identifier
values into their appropriate range:
%
$\crlaaction^{i+1}_{out} = f(\crlaaction^{i}_{out}) = \crlaaction^{i}_{out} \times \beta^{i+1} + \crlaaction^{i+1}$,
%
with $\beta^{i+1}$ being the number of nodes at level $i+1$. 
%
More generally, given two actions $\crlaaction^i$ and $\crlaaction^{i+1}$ obtained from agents at 
levels $i$ and $i+1$, 
where for both actions we have identifiers in the range $[0\isep \rvert\crlaactions^i\rvert - 1]$,
and $[0\isep \rvert\crlaactions^{i+1}\rvert - 1]$ respectively, 
then the action component identifier at level $i+1$ is 
$\crlaaction^{i} \times \rvert\crlaactions^{i+1}\rvert + \crlaaction^{i+1}$.
The executive primitive action meanwhile will be computed via:
$\action = \crlaaction^{L-1} \times \rvert\crlaactions^{L}\rvert + \crlaaction^{L}$.
An illustration of this tree structure is provided in \autoref{fig:cascading_selection_tree}.

Cooperation among agents is facilitated via QMIX~\citep{rashid2018qmix}, 
%
which uses a non-linear combination of the value estimates 
to compute the joint-action-value during training.
The weights of the mixing network are 
produced using a hypernetwork~\citep{ha2016hypernetworks},
conditioned on the state of the environment. 
In CRLA agents share a replay buffer. 
Therefore, as with BDQ, a synchronised sampling 
equivalent to CERTS~\citep{omidshafiei2017deep} is being used.
%
%
The authors~\citep{tran2022cascaded} recommend limiting the size of the action sets 
to 10 -- 15 actions, and to choose an $L$ that allows the approach
to reconstruct the intended action set $\crlaactions$.
%
%
%
CRLA-QMIX was evaluated on two environments against a 
version of CRLA using independent learners, and a single 
DDQN:
\begin{itemize}
\item A toy-maze scenario with a discretized action space, representing
the directions in which the agent can move.
The agent received a small negative reward for each step, and positive
one upon completing the maze.
An action size of 4096 was selected, with $n=12$ actuators.
\item A partially observable CybORG capture the flag scenario from Red's perspective. 
%
%
Upon finding a flag a large positive reward was received. 
A smaller reward was obtained for successfully hacking a host.
%
%
\end{itemize} 
\commentout{
\begin{table}[h]
\centering
\resizebox{0.8\columnwidth}{!}{
\begin{tabular}{ |p{2cm}||p{2cm}|p{2cm}|p{2cm}|}
 \hline
 \textbf{Hosts} & 
 \textbf{State Space} &  
 \textbf{Action Space}  &  
 \textbf{Agents}  \\  
\hline
50 & 573 & 1326 & 3 \\
\hline
60 & 685 & 1830 & 4 \\
\hline
70 & 797 & 2414 & 4 \\
\hline
100 & 1133 & 4646 & 4 \\
\hline
\end{tabular}}
\caption{CybORG configurations used
for evaluating CRLA~\citep{tran2022cascaded}.}
\label{tab:CRLA_CybORG_Eval_Configs}
\end{table}}

CRLA significantly outperformed DDQN on both the maze task and CybORG
scenarios with more than 50 hosts. 
CRLA-QMIX had less variance and better stability than CRLA with ILs.
However, in an evaluation scenario with 60 hosts the converged 
policies show similar rewards and steps per episode, potentially
explained by the fact that CRLA-ILs also makes use of CERTs. 
The authors note that hyperparameter tuning
for the QMIX hypernetwork was time consuming.
\update{Given that the action signatures for Blue and Red have 
the same format in CybORG CAGE Challenge 2, it is very likely
that the same challenges will be encountered when applying CRLA
to the task of defending a network.
Nevertheless, we recommend including CRLA in future endeavours 
that attempt to establish which current high-dimensional
action approach is most suited for the task of cyber-defence.}

%
%

\subsubsection{Discrete Sequential Prediction of Continuous Actions}

\begin{figure}[h]
\centering
\includegraphics[width=\columnwidth]{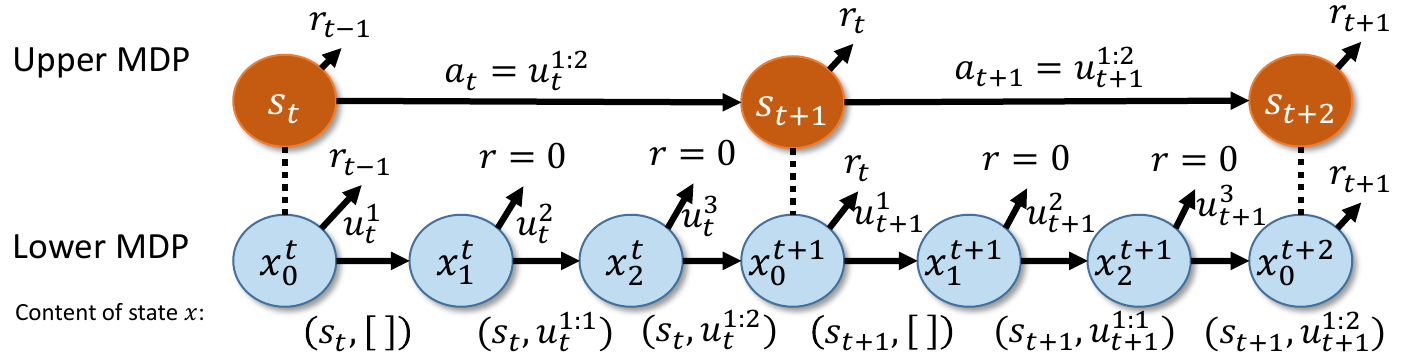}
\caption{Sequential DQN Architecture (adapted from \citep{metz2017discrete}).}
\label{fig:SDQN}
\end{figure}

\cite{metz2017discrete} propose a Sequential DQN (SDQN)
for discretized continuous control.
The original (\emph{upper}) MDP with $N$ (actuators) 
times $D$ (dimensions) actions is transformed into 
a \emph{lower} MDP with $1 \times D$ actions.
The lower MDP consists of the compositional action component, 
where $N$ actions are selected sequentially. 
Unlike BDQ actuators take turns selecting actions, and can
observe the actions that have been selected by others 
(see~\autoref{fig:SDQN}).
Also, the action composition was obtained using a single DNN that learns
to generalize across actuators.
An LSTM~\citep{hochreiter1997long} was used to keep 
track of the selected actions.
For stability, SDQN learns Q-values for both the upper and lower 
MDPs at the same time, performing a Bellman backup from the lower to the 
upper MDP for transitions where the Q-value should be equal. 
In addition, a zero discount is used for all steps 
except where the state of the upper MDP changes. 
Another requirement is the pre-specified ordering of actions.
The authors hypothesise that this may negatively impact training
on problems with a large number of actuators. 

The authors evaluate SDQN against DDPG on a number of continuous control
tasks from the OpenAI gym~\citep{1606.01540}, including 
Hopper ($N=3$), 
Swimmer ($N=2$), 
Half-Cheetah ($N=6$), 
Walker2d ($N=6$),
and Humanoid($N=17$).
SDQN outperformed DDPG on all domains except Walker2d.
With respect to granularity SDQN required $D \geq 4$. 
The authors also evaluated 8 different action 
orderings at 3 points during training on Half Cheetah.
All orderings achieved a similar performance.
\update{Similar to BDQ, SDQN is a natural benchmarking
candidate for cyber-defence environments either featuring
a MultiDescrete action space, or where a suitable wrapper
could be added.}

\subsubsection{Time-Varying Composite Action Spaces}

\cite{9507301} propose the 
\emph{Structured Cooperation Reinforcement Learning (SCORE)}
algorithm that accounts for dynamic time-varying action spaces,
\ie environments where a sub-set of actions become 
temporarily invalid~\citep{9507301}.
SCORE can be applied to heterogeneous action spaces
that contain continuous and discrete values.
A series of DNNs model the composite action space.
A centralized critic and decentralized actor (CCDA) approach~\citep{li2020f2a2} 
facilitates cooperation among the agents.
In addition a Hierarchical Variational Autoencoder (HVAE)~\citep{edwards2017towards} 
maps the sub-action spaces of each agent to a common 
latent space. 
This is then fed to the critic, allowing the critic 
to model correlations between sub-actions, enabling 
the explicit modelling of dependencies between the 
agents' action spaces. 
A graph attention network (GAT)~\citep{GAT} is used as the critic, 
in order to handle the varying numbers of agents (nodes). 
The HVAE and GAT are critical for SCORE to cope with
varying numbers of heterogeneous actors.
As a result, SCORE is a two stage framework, that must first learn 
an action space representation, before learning a robust
and transferable policy. 
For the first phase a sufficient number of trajectories must be gathered for each sub-action space. 
The authors use a random policy to generate these transitions. 
Once the common latent action representation is acquired the training can
switch to focusing on obtaining robust policies.

SCORE is evaluated on a proof-of-concept task --
a Spider environment based on the MuJoCo Ant environment
-- 
and a Precision Agriculture Task, where the benefits of
a mixed discrete-continuous action space comes into play.
SCORE outperforms numerous baselines on both \envs,
including MADDPG~\citep{lowe2017multi}, PPO~\citep{schulman2017proximal}, 
SAC~\citep{haarnoja2018soft}, H-PPO~\citep{fan2019hybrid}, 
QMIX~\citep{rashid2018qmix} and MAAC~\citep{iqbal2019actor}.

\update{Score has much potential for \aco, given that it addresses 
one of the key challenges for cyber-defence agents:
a time-varying action space resulting from hosts joining/leaving 
the network, and new services being launched, including decoys.}
However, the code for SCORE has not been made publicly available.
\update{Therefore, benchmarking SCORE against the other approaches
listed in this section would require a reimplementation of the framework.}

%

\subsubsection{Action Decomposition Approaches for Slate-MDPs}

Action decomposition approaches have also been applied to Slate-MDPs.
%
Two noteworthy efforts are
\emph{Cascading Q-Networks} and
\emph{Slate Decomposition}.
  
\textbf{Cascading Q-Networks (CDQNs):}
\cite{pmlr-v97-chen19f} introduce a model-based \RL approach for
the recommender problem that utilizes Generative Adversarial Networks (GANs)\citep{goodfellow2014generative}
to imitate the user's behaviour dynamics and reward function.
The motivation for using GANs is to address the issue
that a user's interests can evolve over time, and the fact that the recommender
system can have a significant impact on this evolution process.
In contrast, most other works in this area use a manually designed
reward function. 
A CDQN is used to address the large action space, through which
a combinatorial recommendation policy is obtained. 
CDQNs consist of $k$ related Q-functions, where actions
are passed on in a cascading fashion. 
%

CDQNs were evaluated on six real-world recommendation datasets --
\emph{MovieLens}, \emph{LastFM}, \emph{Yelp}, \emph{Taobao}, \emph{YooChoose}, and \emph{Ant Financial} --
against a range of non-RL recommender approaches, including IKNN, S-RNN,
SCKNNC, XGBOOST, DFM, W\&D-LR, W\&D-CCF, and a Vanilla DQN.
On the majority of  these datasets, 
the generative adversarial model is a better fit to user behaviour 
with respect to held-out likelihood and click prediction. 
With respect to the resulting model policies, better cumulative 
and long-term rewards were obtained.
The approach took less time to adjust compared to approaches that 
did not make use of the GANs synthesized user.
However, we caution that applying model-based RL approaches to complex 
asymmetrical adversarial games, such as \aco, requires further 
considerations.

\textbf{Slate Decomposition:} 
Slate decomposition, or \emph{SlateQ}, is an approach where the Q-value estimate 
for a slate $\bm{\action}$ can be decomposed into the item-wise Q-values of its constituent 
items $u_i$~\citep{ie2019reinforcement}.
Having a decomposition approach that can learn $\bar{Q}(\state, \action_i)$ for an item $i$ 
mitigates the generalization and exploration challenges listed above.
However, the ability to successfully factor the Q-value of a slate $\bm{\action}$ 
relies on two assumptions: \emph{Single Choice} (SC) and 
\emph{Reward/Transition Dependence on Selection} (RTDS).
%
%
The authors show theoretically that given the standard 
assumptions with respect
to learning and exploration~\citep{sutton2018reinforcement}, as well as SC and RTDS, 
SlateQ will converge to the true slate Q-function $Q_{\pi}(x, \bm{\action})$.
SlateQ is evaluated in a simulation, while validity and scalability were 
tested in live experiments on YouTube. 

\update{Recommender systems have the potential to reduce the
response time to cyber threats \citep{gadepally2016recommender}. 
Methods such as CDQN and SlateQ can serve as valuable filtering systems,
allowing human operators to make timely decisions based on suggested actions.
However, this will require the formulation and implementation of cyber-defence
environments that adhere to the Slate-MDP problem formulation,
and can serve as a training environment for DRL based recommender systems.}

\subsection{Action Elimination Approaches} \label{sec:action_elimination}

\begin{figure}[h]
\centering
\includegraphics[width=0.4\columnwidth]{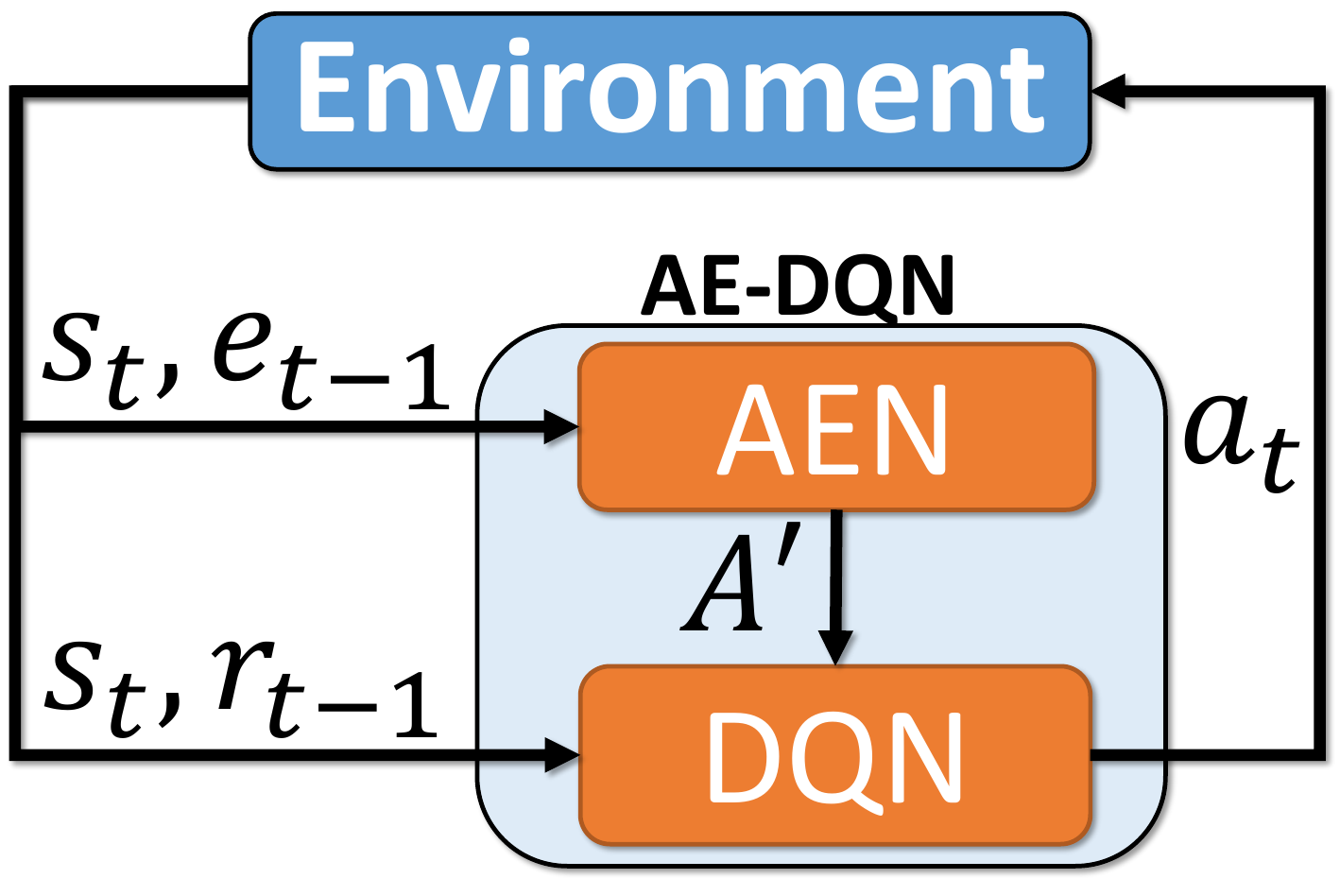}
\caption{Action Elimination DQN (adapted from~\citep{zahavy2018learn}). 
The agent selects an action $\action_{t}$, 
and observes a reward $r_{t-1}$, the
next observation $\observation_{t}$ and an elimination signal $e_{t-1}$. 
The agent uses this information to learn two function approximation 
deep networks: a DQN and an AEN. The AEN provides an admissible actions set $\actions' \subseteq \actions$
to the DQN, from which the DQN can pick the next action $\action_{t+1}$.}
\label{fig:aedqn}
\end{figure}

Learning with a large combinatorial action space is often 
challenging due to a large number of actions being either redundant
or irrelevant within a given state~\citep{zahavy2018learn}. 
\update{Eliminating redundant actions can assist exploration, 
as demonstrated by the CAGE Challenge 2 winning solution submitted
by Cardiff University, who used domain knowledge to decide which
actions to exclude~\citep{cage_challenge_2_announcement}.} 
\update{However, standard} \RL agents lack the ability to determine a sub-set of relevant 
actions. 
\update{To address this}, there have been efforts towards the state dependent 
elimination of actions.
\cite{zahavy2018learn}
combine a DQN with an action-elimination
network (AEN), which is trained 
via an elimination signal $e$, resulting in AE-DQN 
(\autoref{fig:aedqn}).
After executing an action $\action_t$, the environment will return
a binary action elimination signal in $e(\state_t, \action_t)$ in addition to the 
new state and reward signal.
The elimination signal $e$ is determined using
domain-specific knowledge.
A linear contextual bandit model is applied to the outputs of the
AEN, that is tasked with eliminating irrelevant actions with a
high probability, balancing out exploration/exploitation. 
Concurrent learning introduces the challenge that the learning 
process of both the DQN and AEN affect the state-action 
distribution of the other. 
However, the authors provide theoretical guarantees
on the convergence of the approach using linear contextual bandits.
While the AE-DQN was designed for text based games 
the authors note that the approach is applicable to any \env 
where an elimination signal can be obtained via 
a rule-based system. 

AE-DQN is evaluated on both Zork and a $K$-Room Gridworld
\env. 
For the $K$-rooms Gridworld \env a significant gain for the 
use of action elimination was observed as the number of
categories $K$ was increased.  
Similar benefits were observed in the Zork \env, \eg AE-DQN using 215 actions being able
to match the performance of a DQN trained with a reduced action space of 35 actions, while
significantly outperforming a DQN with 215 actions.
However, questions remain regarding the extent to which
the approach is applicable to \emph{very} high dimensional
action spaces, where additional considerations may be required
as to how the AEN can generalize over actions.

Action elimination has also been applied to Slate-MDPs.
\cite{10.1145/3289600.3290999} adapted
the REINFORCE algorithm into a top-$k$ neural candidate generator
for large action spaces.
The approach relies on data obtained through previous recommendation
policies (behaviour policies $\beta$), which are utilized as a means
to correct data biases via an importance sampling weight while training 
a new policy. 
Importance sampling is used due to the model being trained without access 
to a real-time environment. 
Instead the policy is trained on logged feedback of actions 
chosen by a historical mixture of policies, which will have 
a different distribution compared to the one that is being updated. 
A recurrent neural network is used to keep track of the evolving user interest.

With respect to sampling actions, instead of choosing the $k$
items that have the highest probability, the authors use
a stochastic policy via Boltzmann exploration.
However, computing the probabilities for all $N$ actions 
is computationally inefficient.
Instead the authors chose the top $M$ items, select their
logits, and then apply the softmax over this smaller set $M$
to normalize the probabilities and sample from this smaller 
distribution.
The authors note that when $M \ll K$, one can still retrieve
a reasonably sized probability mass, while limiting the risk
of bad recommendations.
Exploration and exploitation are balanced
through returning the top $K'$ most probable items 
(with $K' < k$), and sample $K-K'$ items from the remaining
$M - K'$ items. 
The approach is evaluated in a production RNN candidate generation model 
in use at YouTube, and experiments are performed to validate the 
various design decisions. 

\update{Applying the above methods to cyber-defence environments could
provide interesting insights into the types of actions that are frequently 
eliminated.
These insights could subsequently inform the development of novel
action masking methods for cyber-defence environments.
Finally, comparing the actions eliminated by an approach such as AE-DQN 
to those eliminated by domain experts would also represent another 
interesting direction for future research in this area, \eg through 
applying AE-DQN to CAGE Challenge 2 and comparing the eliminated 
actions with those selected by Cardiff University~\citep{cage_challenge_2_announcement}.}

\subsection{Hierarchical Reinforcement Learning} \label{sec:hrl}

The RL literature features a number of hierarchical formulations. 
Feudal RL features Q-learning with a managerial hierarchy,
where \say{managers} learn to set tasks for
\say{sub-managers} until agents taking atomic actions
at the lowest levels are reached~\citep{dayan1992feudal,vezhnevets2017feudal}. 
There have also been factored hierarchical approaches that decompose
the value function of an MDP into smaller constituent 
MDPs~\citep{dietterich2000hierarchical,guestrin2003efficient}.

\update{Hierarchical DRL approaches have also been applied to cyber-defence scenarios,
including solutions submitted by Cardiff University and Team Mindrake for CybORG CAGE Challenge 2~\citep{foley2022autonomous,cage_challenge_2_announcement}.
Both feature a single agent policy that specializes on one of the Red opponents provided by CAGE Challenge 2,
and a controller for determining when to load each specialist policy.
For Cardiff University this controller uses rules-based fingerprinting, 
while Team Mindrake's solution consists of RL based controller~\citep{foley2022autonomous},
tasked with selecting one of the pre-trained agents at each time-step, to execute low-level actions.}

\begin{figure}[h]
\centering
\includegraphics[width=0.5\columnwidth]{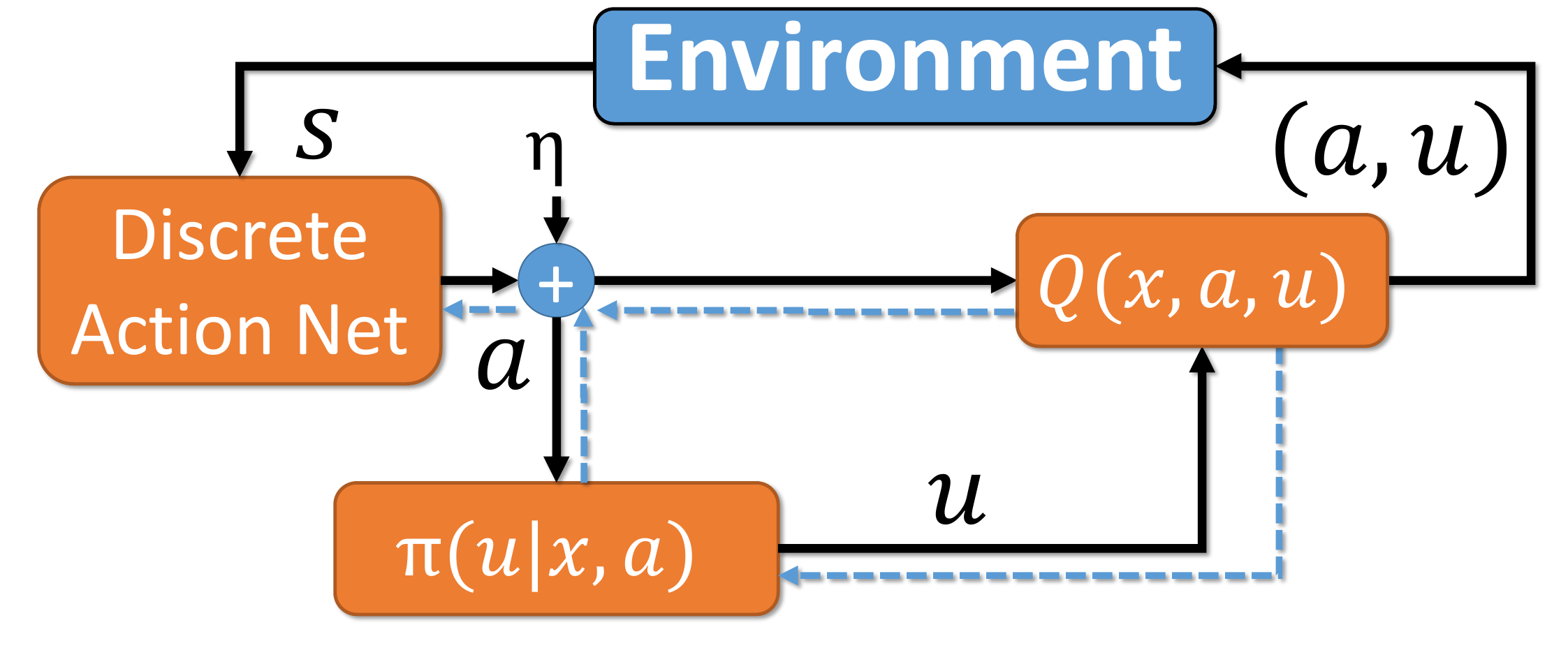}
\caption{PASVG(0) (adapted from~\citep{wei2018hierarchical}).}
\label{fig:pasvg}
\end{figure}

Hierarchical RL has also been applied to
high-dimensional parameterized action MDPs.
\cite{wei2018hierarchical} propose 
Parameterized Actions Trust Region Policy Optimization (TRPO)~\citep{schulman2015trust} and
Parameterized Actions SVG(0),
hierarchical \RL approaches designed for Parameterized Action MDPs.
The approaches consist of two policies implemented by neural networks. 
The first network is used by the discrete action policy $\pi_{\theta}(\action \rvert \state)$
to obtain an action $\action$ in state $\state$. 
The second network is for the parameter policy.
It takes both the state and discrete action as inputs, and 
returns a continuous parameter (or a \emph{set} of continuous parameters) 
$\pi_{\vartheta}(\continuousaction \rvert \state, \action)$.
Therefore, the joint action probability for $(\action, \continuousaction)$ given a state $\state$
is conditioned on both policies: 
$\pi(\action, \continuousaction \rvert \state)\ = \pi_{\theta}(\action \rvert \state)\pi_{\vartheta}(\continuousaction \rvert \state, \action)$.
This formulation has the advantage that since the action
$\action$ is known before generating the parameters, there is
no need to determine which action tuple $(\action, \continuousaction)$ has the
highest Q-value for a state $\state$.
In order to optimize the above policies, methods are required
that can back-propagate all the way back through the discrete action
policy.  
Here the authors introduce a modified version of TRPO~\citep{schulman2015trust}
that accounts for the above policy formulation, 
and a parameterized action stochastic value gradient 
approach that uses the Gumbel-Softmax trick for drawing
an action 
$u$ and back-propagating through $[\theta, \vartheta]$.
The approach is depicted in \autoref{fig:pasvg}.
With respect to evaluation, PATRPO outperforms PASVG(0) and PADDPG
within a Platform Jumping \env. PATRPO also outperforms PADDPG 
within the Half Field Offense soccer~\citep{HFO} \env with no goal 
keeper~\citep{hausknecht2015deep}.

\update{While we include hierarchical DRL methods here for completeness, 
their current benefit for \aco environments featuring high-dimensional action
spaces is less clear.
However, methods such as PATRPO and PASVG(0) could become relevant for any
future cyber-defence challenges requiring solutions featuring a parameterized
action space.}

\subsection{Curriculum Learning} \label{sec:actions:cl}

Curriculum learning (CL) approaches attempt to accelerate the learning process
through initially subjecting the \RL agent to 
a simplified version of the problem, and subsequently gradually
increasing the task complexity, \eg via a progression function
~\citep{bassich2019continuous,bassich2020curriculum}.
This approach has also been applied to \RL
for combinatorial action spaces. 
\cite{pmlr-v119-farquhar20a} introduce a CL 
approach using a growing action space (GAS).
For an MDP with unrestricted action space $\actions$
the authors define a set of $N$ action spaces $\actions_l, l \in \{0, ..., N-1\}$.
Each action space is a subset of the next level $l$: 
$\actions_0 \subset \actions_1 \subset ... \subset \actions_{N-1} \subset \actions$.
A policy restricted to an action space $\actions_l$ is denoted as $\pi_l(\action, \state)$.
The optimal policy for this restricted policy class is $\pi^*_l(u, x)$,
and the corresponding action-value and value functions are:
$Q^*_l(\state, \action)$ and $V^*_l(\state) = \max_\action Q^*_l(\state, \action)$.
Domain knowledge is used to define a hierarchy of actions.
For every action $\action \in \actions_l$ where $l>0$
there is a parent action $\texttt{parent}_l(\action)$ in the space of $\actions_{l-1}$.
Given that at each level, subsets of action spaces are subsets
of larger action spaces, the actions available in  $\actions_{l-1}$
are their own parents in $\actions_{l}$.
%
The authors note that in 
many environments Euclidean distances are a valid measure
for implementing a heuristic for defining a hierarchy over
actions.
%
%
%

\commentout{During training the authors apply off-action-space learning
meaning that, given that off-policy learning is used,
the authors simultaneously learn value function $Q_l(\state,\action)$
for multiple action spaces $\actions_l$, using samples drawn
from the behaviour policy $\pi_l$ to train all of
the higher $l$ value functions.
The approach also relies on the (strong) assumption
that a value function trained on actions for a
restricted action space can generalize to unexplored
parts of a less restricted action space. 
With respect to estimating values, the authors note:
$V^*_i(\state) \leq V^*_j(\state) \forall \state$ 
if $i < j$.
As a result there is a monotonicity property where
each $Q^*_{l+1}(\state, \action)$ can be estimated through the 
sum  $Q^*_{l}(\state, \action)$ and some positive $\Delta_l(\state, \action)$.
However, due to the mismatch in support the parent of $\action$
must be used instead for this estimate:
%
$Q^*_{l+1}(\state, \action) = Q^*_{l+1}(\state, \texttt{parent}_l(\action)) + \Delta_l(\state, \action)$,
%
where the authors initialize $\Delta_l(\state, \action)$ to be a small value.

GAS uses function approximators that learn a joint state 
representation, that can iteratively be decoded into 
estimates of $Q^*$ for each $l$, and also allows for a flexible
implementation where additional network
layers for each $Q^*_l$ can be added. 
With respect to progression a linear progression function is used.
For multi-agent settings the authors use an unsupervised clustering
algorithm to identify $k$ clusters of agents, and then apply
an action that applies to the entire group.
At level $l=0$ the agents are treated as one group,
but gradually the number of clusters increase over time.
First state-value estimates $V(\state)$ and group-action deltas $\Delta_l(\state, \action_g,g)$
are learnt for each group $g$ at each level $l$.
Subsequently estimated group-action values are computed using $Q^*_{l+1}(\state, \action_g) = Q^*_{l+1}(\state, \texttt{parent}_l(\action)) + \Delta_l(\state, \action_g, g)$.
The authors use $Q^*_{-1}(\state, \cdot ) = V(\state)$.}

An ablation study on discretized Acrobat and Mountaincart environments shows the value
of efficiently using the data collected during training across levels.
The authors also evaluate their approach on SMAC, using a far larger number 
of units compared to those usually used in MARL experiments --
\ie scenarios with 50-100 instead of 20-30. 
In addition, the task difficulty is increased through
having randomized starting positions, and scripted opponent
logic that holds its position until any agent-controlled unit
is in range. 
Subsequently the enemy focus-fires on its closest enemy.
Having to locate the enemy first increases the exploration challenge.
The authors demonstrate the advantage of their approach GAS(2) (GAS with 
2 levels) against various ablations of their approach and methods that
directly train on the full action space. 

\cite{yu4167820curriculum} take a similar approach to GAS~\citep{pmlr-v119-farquhar20a} 
for decision making in intelligent healthcare.
A Progressive Action Space (PAS) approach allows the learner
to master easier tasks first, via generalized actions, before gradually
increasing the granularity, \eg modifying the precise volume 
of a drug to be given to a patient. 
This work focuses on an offline RL setting, where traditional approaches lead
to inaccurate state-action evaluation for those actions seldom applied 
by the physicians in the historical clinical dataset --
a common problem for offline \RL~\citep{fujimoto2019off}. 
Similar to GAS~\citep{pmlr-v119-farquhar20a}, PAS also requires domain 
knowledge for defining $N$ abstracted action spaces $\{\action_1, \action_2, ..., \action_N\}$
and a corresponding number of curricula $\{M_1, M_2, ..., M_N\}$.
Value and policy transfer approaches
are used to transfer knowledge across levels. 

\update{In contrast to the above works, 
the literature also features curriculum 
learning approaches that have been evaluated on \aco environments.
As mentioned in \autoref{sec:hdas_approaches:marl_and_coordination},
\cite{hicks2023canaries} use curriculum learning for overcoming the
non-stationarity pathology of MADRL. 
This approach allows for the identification of the optimal symbolic
and learning agent split.
Meanwhile, \cite{10165310} used curriculum learning to determine exploitability
of software vulnerabilities and the complexity of the network topologies.
The authors find that both are helpful for obtaining robust cyber-defence agents.}

\subsection{High-Dimensional Action Approaches Summary}

\update{Due to an explosion in the number of state-action pairs, traditional
DRL approaches do not scale well to environments that confront learners
with a high-dimensional combinatorial action space. 
Therefore, our idealised DRL-ACD agent requires
the ability to generalize over an action space,
handle time varying actions, and
provide sublinear complexity. 
In this section we have identified a number of approaches that attempt to 
meet these criteria, including approaches belonging to the
following five categories: 
proto action based approaches,
action decomposition,
action elimination,
hierarchical approaches, and
curriculum learning.}

\update{Out out the above categories proto action (\WP~\citep{schwartz2019autonomous})
and action decomposition (CRLA~\citep{tran2022cascaded})
have been evaluated within a cyber-defence context, 
albeit from the perspective of a Red cyber-attacking agent.
While the two approaches demonstrated some potential at managing
the same type of high-dimensional action space that our idealised cyber-defence
agent will be confronted with, both have glaring weaknesses,
These weaknesses include the computational complexity of maintaining an action-embedding space for the
\WP approach when dealing with varying network topologies, 
along with exploration deficiencies.
CRLA, meanwhile, requires an extensive hyperparameter tuning. 
Furthermore, while its algebraic formulation provides an expressive and efficient 
means for encoding high-dimension actions,
considerations are required as to how one can structure a MultiDiscrete action space 
appropriately
to enable generalization across actions.
An evaluation is also required on how CRLA would react to time varying actions.}

\update{
Given that there now exists a plethora of cyber-defence environments
that can confront DRL agents with the high-dimensional action 
space challenge (see \autoref{sec:envs}), an extensive empirical
evaluation of the approaches discussed in this section 
could provide answers regarding which method is most likely to
scale to the full autonomous cyber-defence challenge.
In addition, a hybrid approach that combines the discussed components could prove beneficial.
For example, given the fact that removing unnecessary actions has been demonstrated
to facilitate learning within cyber-defence environments~\citep{cage_challenge_2_announcement},
combining either proto or action decomposition approaches with a learned action 
masking facility, derived from approaches discussed in \autoref{sec:action_elimination},
could simplify the learning task.
Naturally curriculum learning approaches could provide further benefits, 
in particular as the work from \cite{hicks2023canaries} demonstrates.
Other topics that merit further exploration, including the evaluation of 
principled exploration strategies within 
the \WP architecture's action embedding space,
and an extensive evaluation of action decomposition methods for MultiDiscrete
action spaces when combined with action masking.}

\section{Adversarial Learning} \label{sec:adv_learning}
 
\update{While the cyber security community is looking to ML for fully automating cyber-defence systems,
the learning dynamics resulting from defenders and attackers,
that can be both strategic and adaptive in in their behaviour~\citep{kott2018autonomous},
is receiving insufficient attention~\citep{de2021fixed,bakker2021solvability}.
While a number of approaches discussed in the previous sections
have been applied to adversarial domains, 
including \aco (\eg \WP~\citep{nguyen2020multiple} and CRLA~\citep{tran2022cascaded}), 
these approaches were trained against a small set of stationary opponents.}
\update{In practice, attack behaviors benefit from the speed of automation and machine-level reactions, 
enabling AI approaches to rapidly compute a plethora of attack strategies~\citep{TANG2024103871,apruzzese2020deep}.  
As a result, it is impossible for human cyber-security experts to enumerate 
all possible attack scenarios in advance~\citep{TANG2024103871}.
Therefore, given that the \aco problem is non-stationary, with
Red and Blue adjusting their approach over time,
considerations are required as to what solutions cyber-defence 
and attacking agents will converge upon,
as they iteratively attempt to compute \emph{approximate
best responses} (ABRs) to each others' policies~\citep{de2021fixed}.}

\subsection{The Adversarial Learning Challenge}

\update{To address the fact that work on cyber-security 
lacks discussions on the interaction of multiple players, 
along with a modeling of how different individual behaviors
influence learning dynamics and the resulting game equilibrium,
\cite{TANG2024103871} propose using game theory as a tool for analysing 
the strategic interactions between learning cyber-defence and 
attacking agents. 
This is an approach that has previously been found to deliver 
interesting insights within the MADRL literature beyond \aco~\citep{leibo2017multi,yang2020alphaalpha,lanctot2017unified}.
Therefore, in this section we shall formally define the adversarial 
learning problem, desirable solution concepts, and the approaches
designed to help learning agents converge upon policies that are 
hard to exploit, and can generalize across opponents.}

\update{First, we note that, while \aco environments are adversarial,
they are not always \emph{zero-sum games}.}
\update{\cite{shashkov2023adversarial} observe that 
a zero-sum game formulation is a simplified assumption, 
given the complexity of cybersecurity realities.}
For instance, in CAGE Challenge 2 Blue receives 
a penalty for restoring a compromised host due to the
action having consequences for any of the Green agents 
currently working on this facility.
Here, Red does not receive a corresponding reward $r_{Red}=-r_{Blue}$.
%
%
Therefore, \aco lacks 
a key property from two player zero-sum games, where the gain of 
one player is equal to the loss of the other player.
%
%

\update{The extent to which a game can be considered zero or 
general-sum may depend on the objectives of the attacker.
For example, \cite{de2021fixed} state that botnet attacks
(that attempt to hijack hosts for processes
such as sending spam emails, distributing malware, or framing
DDoS attacks) are general-sum games. 
This is due to both the attacker and the
defender having an aligned interest in the 
network services remaining operational.}
However, in this section we shall 
treat the \aco problem 
as a quasi zero-sum game, with the
assumption that the consequences of Red winning
outweigh other factors.  
%
%
Therefore, we shall assume: 
$\G{1}{{\pi, \mu}} \approx -\G{2}{{\pi, \mu}}$.
An equilibrium concept commonly used in this class of games to define solutions is 
the Nash equilibrium~\citep{nash1951non}: 
\begin{definitionsec}[Nash Equilibrium]
A joint policy $\bm{\pi}^*$ is a Nash equilibrium \emph{iff} no player $i$ can improve their gain through unilaterally deviating from $\bm{\pi}^*$:
\begin{equation} \label{eq:nash_equilibrium}
\forall i, \forall \pi_i \in \Delta(\states,\actions_i), \forall \state \in \states, 
\G{i}{\JointPolicy{\pi^*_i}{\bm{\pi}^*_{-i}}}
\geq  \G{i}{\JointPolicy{\pi_i}{\bm{\pi}^*_{-i}}}.
\end{equation}
\end{definitionsec}
Our focus is on finite two-player (quasi) zero-sum games, 
where an equilibrium is referred to as a saddle point,
representing the value of the game~$v*$. 
Given two policies $\pi_1$, $\pi_2$, the equilibria of a finite zero-sum game is:
\begin{theoremsec}[Minmax Theorem] In a finite zero-sum game:
\begin{equation} \label{eq:minmax_theorem}
max_{\pi_1} min_{\pi_2} \G{i}{\JointPolicy{\pi_1}{\pi_2}} = 
min_{\pi_2} max_{\pi_1} \G{i}{\JointPolicy{\pi_1}{\pi_2}} = v^*.
\end{equation}
\end{theoremsec}

Above is one of the fundamental theorems of game theory,
which states that every finite, zero-sum, two-player game 
has optimal mixed strategies~\citep{v1928theorie}. 
We shall discuss the implications of the above equations 
for \aco from the perspective of Blue.
Given a joint-policy $\JointPolicy{\pi^*_{Blue}}{\pi^*_{Red}}$
where $\pi^*_{Blue}$ and $\pi^*_{Red}$ represent optimal mixed 
strategies, then by definition Red will be unable to learn
a new best response $\pi_{Red}$ that improves on $\pi^*_{Red}$.
Obtaining $\pi^*_{Blue}$ guarantees that Blue will perform well, 
even against a worst case opponent~\citep{perolat2022mastering}.
This means that, even if the value of the game $v^*$ 
is in Red's favour, assuming that Blue has found $\pi^*_{Blue}$,
then Blue has found a policy that limits the extent to which 
that Red can \emph{exploit} Blue.  
\update{The extent to which an agent can be exploited using this formulation
also offers an insightful evaluation metric within an adversarial
learning context, termed \emph{exploitability}~\citep{lanctot2017unified}.}

\update{We now turn our considerations towards approaches that
our idealised cyber-defence agent can use to find a policy
$\pi^*_{Blue}$, along with an overview of the attempts at using 
adversarial learning approaches that can be found in the \aco
literature.
Even beyond \aco,} one of the long-term objectives of MARL is to limit the 
exploitability of agents deployed in competitive 
environments~\citep{lanctot2017unified,oliehoek2018beyond,heinrich2016deep}.
While a number of methods for limiting exploitability exist 
that are underpinned by theoretical guarantees, 
in practice finding the value of the game 
is challenging even for simple games. 
%
This is due to DRL using function approximators
that are unable to compute \emph{exact}
best responses to an opponent's policy.
%
The best a DRL agent can achieve is an 
ABR.
In addition, for complex games finding a 
Nash equilibrium is intractable.
Here the concept of an approximate Nash equilibrium ($\epsilon$-NE) is 
helpful~\citep{oliehoek2018beyond,lanctot2017unified}: 
\begin{definitionsec}[$\epsilon$-Nash Equilibrium]
The joint-policy $\bm{\pi}^*$ is an $\epsilon$-NE \emph{iff}:
\begin{equation} \label{eq:eqp_ne}
\forall i, \forall \pi_i \in \Delta(\states,\actions_i), \forall \state \in \states, 
\G{i}{\JointPolicy{\pi^*_i}{\bm{\pi}^*_{-i}}} \geq  
\G{i}{\JointPolicy{\pi_i}{\bm{\pi}^*_{-i}}} - \epsilon.
\end{equation}
\end{definitionsec}

\subsection{Towards Addressing Agent Overfitting} \label{sec:agent_overfitting}

\update{Simply training Blue cyber-defence agents against a diverse set 
of Red attacks is insufficient for reducing exploitability.
Here, \cite{de2021fixed} state the importance of considering 
opponents that can rapidly learn more sophisticated attacks,
noting that automated cyber-defence and attack systems of the 
future will interact and co-evolve on increasingly smaller
timescales, potentially learning exploits and best responses
against each other within hours, or, down the line, even milliseconds.
The consequences of deploying policies that were trained against a limited set of opponents 
have already been observed within simple cyber-defence simulation environments. 
Upon conducting experiments in FARLAND, \cite{molina2021network} show that 
when training a Blue agent using known TTPs (without accounting for potential poisoning or
evasion attacks targeting the learning algorithm), its performance
cannot be guaranteed to generalize. 
While the trained Blue agent may achieve an acceptable performance 
against a Red agent that is only assumed to behave in a manner 
consistent with a subset of behaviors described in the ATT\&CK
framework, the performance of the Blue agent degrades
significantly when simply perturbing the behavior of the Red agent using only
gray-like action.
As a result the authors emphasis that autonomous network defenders must
learn to do more than mitigate common adversaries (such
as those described in MITRE’s ATT\&CK dataset \citep{strom2018mitre}),
and be capable of mitigating more sophisticated
(autonomous) cyber-attacking agents.}

The above raises the question: how can we \emph{efficiently} 
limit the exploitability of \aco agents? 
A key insight here from the adversarial learning literature
is that finding $\pi^*_{Blue}$ will require learning
to best respond to principled Red agents,
ideally through computing a best response against $\pi^*_{Red}$.
However, similarly Red will need to face strong Blue agents
to find $\pi^*_{Red}$. 
This raises the need for an iterative adversarial learning
process where Blue and Red learn (A)BRs
to each others' latest policies.

\update{
There have been a number of efforts towards
conducting adversarial learning within a cyber-defence settings
to address agent overfitting.
\cite{shashkov2023adversarial} compare a variety of DRL, evolutionary strategies (ES) 
and Monte Carlo Tree Search methods within 
CyberBattleSim, finding that an approach that combines DRL and ES achieves 
the best comparative performance when attackers and defenders are simultaneously trained, 
compared to when each is trained against its non-learning counterpart.
Similarly, \cite{10216719} also conducted adversarial learning on
an extended version of CyberBattleSim~\citep{MARLon}. 
The authors observed a clear benefit from jointly-training 
Blue and Red agents, finding that the blue agent benefits from being
trained jointly with non-trivial Red agents, and typically outperforming 
Blue agents trained in isolation. 
The same was not true for Red agents. 
However, neither study used exploitability as an evaluation metric, 
instead comparing the jointly-trained Blue agent against a Blue agent 
that was trained without a learning opponent.
This highlights the need for the \aco community to adopt a common 
set of metrics, such as measuring exploitability, for evaluating 
Blue agents within an adversarial learning context.}

\subsection{Adversarial Perturbation Attacks} \label{sec:pertubation_attacks}

\update{A cyber-attacking agent
may be designed to directly target ML cyber-defence models that are tasked
with defending the network.
A popular approach here is to learn how to perturb the data before 
it is fed into a ML model, with the goal of manipulating the model's 
output, \eg to make the model misclassify a sample.
DRL agents are also vulnerable towards such perturbation based attacks.}
\cite{gleave2019adversarial} show that an adversary 
can learn simple attack policies that reliably win against a static opponent 
implemented with function approximators.
Often random and uncoordinated behavior is sufficient to trigger 
sub-optimal actions~\footnote{The authors provide videos of the attack behaviours: \url{https://adversarialpolicies.github.io/}}. 
By adversarial attacks the study refers to attacks on the opponents observation space.
This is quasi equivalent to adding perturbations to images for causing a misclassification
in supervised learning, but doing so via taking actions within the \env.
A worrying finding is that DNNs are more vulnerable
towards high-dimensional adversarial samples~\citep{gilmer2018adversarial,khoury2018geometry,shafahi2018adversarial}.
The same applies for DRL. 
Empirical evaluations on MuJoCo show that the greater the dimensionality 
of the area of the observation space that Red can impact, 
the more vulnerable the victim is towards attacks~\citep{gleave2019adversarial}.
\update{
Recent studies have also shown that adversarial training techniques are vulnerable 
to many different sets of attacks from perturbations that can 
transfer~\citep{korkmaz2022deep} to natural directions~\citep{Korkmaz_2023} 
and decoupled non-robust features~\citep{korkmaz2021investigating}. 
Furthermore, to make robust decisions, while receiving potentially perturbed observations,
\cite{10.5555/3618408.3619131} showed that it is possible to detect these adversarial 
state observations based on the curvature of the deep reinforcement 
learning policy manifold.}

\update{
Naturally, cyber-attack agents are not immune towards perturbation attacks.
Instead, there is a concern that perturbation attacks may be more challenging 
to identify, compared to other data modalities, as ensuring that the overall 
information throughput remains above a certain operational threshold can
be sufficient for the attack to remain undetected~\citep{de2021fixed}. 
This is in contrast to other data modalities (\eg image data), 
where methods typically have to search for the smallest possible adversarial perturbations to cause
a desired response from the targeted model, without the attack being obvious~\citep{moosavi2016deepfool}.}

\update{Adversarial network perturbations do require considerations regarding the number of environment steps,
along with the ability to solve the credit assignment problem.
To achieve this, \cite{de2021fixed} introduce Fast Adversarial Sample Training (FAST), 
a DRL approach for generating black-box adversarial attacks on network flow classification (NFC). 
The authors evaluate the convergence of MA-DDPG agents for the purpose of the perturbation (attacking) of data 
and defending against NFC attacks.}

\update{
Perturbation attacks can also pose a problem during training.
\cite{han2020adversarial} find that for autonomous cyber-defence DRL agents within 
partially observable environments are vulnerable towards causative attacks that target 
the training process and poison the DRL agents. 
These attacks are particularly successful even when the attacker only has partial observability of the environment. 
To mitigate these poisoning attacks the authors propose an effective inversion defence method
that counteracts the attacker's perturbations.
Similarly, \cite{10004318} explore data transmission resilience, between agents,
to cyber-attacks on cluster-based systems that use heterogeneous MADRL. 
An algorithm using a proportional feedback controller and a DQN was used to 
defend against the Fast Gradient Sign Method (FGSM), a popular white-box attack~\citep{goodfellow2014explaining}.}

\subsection{Approaches Towards Limiting Exploitability} \label{sec:limiting_expl}

\update{The works by \cite{gleave2019adversarial} and others show that agents trained via
simplistic training schemes -- \eg self-play -- are very far from an $\epsilon$ bounded
Nash equilibrium.}
Furthermore, it is well established within the adversarial learning literature 
that na\"ive independent learning approaches fail to generalize 
well due to a tendency to overfit on their opponents,
also known as joint-policy correlation~\citep{lanctot2017unified}.
%
%
This raises the need for training schemes designed to limit the exploitability 
of agents. 
\cite{perolat2022mastering} identify three categories of approaches
for reducing the exploitability of agents: 
regret minimization,
regret policy gradient methods,
and best response techniques.

The first category includes approaches that scale counterfactual regret minimization (CFR) using
deep learning. 
Deep-CFR~\citep{brown2019deep} trains a regret DNN via an experience replay buffer containing 
counterfactual values.
A limitation of this approach is the sampling method,
%
in that it does not scale to games with a large branching factor~\citep{perolat2022mastering}.
There are model-free regret-based approaches which use DNNs that scale to larger games,
such as \emph{deep regret minimization with advantage baselines and model-free learning}
(DREAM)~\citep{steinberger2020dream} and the advantage regret-matching actor-critic (ARMAC)~\citep{gruslys2020advantage}. 
However, these approaches rely on an importance sampling term in order to remain unbiased.
The importance weights can become very large in games with a long horizon~\citep{perolat2022mastering}. 
To generalize, these techniques require the generation of an
average strategy, necessitating either the complete retention of all strategies from 
previous iterations, or an error prone approximation, \eg a DNN trained via supervised
learning~\citep{perolat2022mastering}.

The second category of methods approximates CFR 
%
via a weighted policy gradient~\citep{srinivasan2018actor,perolat2022mastering}. 
However, the approach is not guaranteed to converge to a Nash equilibrium~\citep{perolat2022mastering}.
In contrast Neural Replicator Dynamics (NeuRD)~\citep{hennes2019neural},
an approach that approximates the Replicator Dynamics from evolutionary game theory
with a policy gradient, is proven to converge to a Nash
equilibrium. 
NeuRD has been applied to large-scale domains, 
as part of \emph{DeepNash}~\citep{perolat2022mastering}.
It is used to modify the loss function 
for optimizing the Q-function and the 
policy~\citep{perolat2022mastering}. 
DeepNash was recently introduced as a means of tackling the game of Stratego.
However, its wider applicability is yet to be explored.
In the remainder of this section we shall therefore focus on 
the third category of approaches:
best response techniques.

\subsection{Best Response Techniques} \label{sec:best_resposne_techniques}

\update{In their work on fixed points in cyber-space with respect to 
dealing with perturbation attacks, \cite{de2021fixed} 
observe that attackers might exploit defenders of network 
flow classifiers through confining themselves to a 
perturbation subspace. 
However, there is a danger of attackers suddenly switching to a different 
perturbation subspace which, due to catastrophic forgetting, 
the defender will either have not been trained on for a prolonged period of time, 
and no longer know how to defend. 
This is in essence the same challenge as the mode omission and degeneration
pathologies observed within the generative adversarial learning literature~\citep{oliehoek2018beyond}.}

\update{Antidotes for catastrophic forgetting within 
an adversarial setting can include continual learning and ensuring replay buffers are utilized that
retain characteristics of previously encountered attacks~\citep{de2021fixed}.
Below we shall discuss a number of adversarial learning approaches from 
the literature that improve upon these approaches. 
Specifically, the approaches address the catastrophic forgetting of previous 
opponents through maintaining a history over past policies.
As we shall see, 
population based approaches can provide theoretical convergence guarantees
when combined with principled policy sampling strategies.}

In recent years a number of principled approaches from the MARL literature -- 
originally designed for competitive games 
with a low-dimensional state space --
have been scaled to \emph{deep} MARL.
Population-based and game-theoretic training regimes have shown a significant
amount of potential~\citep{li2023combining}.
Early work in this area by \cite{heinrich2016deep}
scaled \emph{Fictitious Self-Play} for domains suffering from the curse-of-dimensionality,
resulting in \emph{Neural Fictitious Self-Play} (NFSP).
This approach approximates extensive-form fictitious play by progressively training a 
best response against the average of all past policies using off-policy DRL. 
A DNN is trained using supervised learning to 
imitate the average of the past best responses.

\update{
\cite{yu4167820curriculum} have explored the benefits of self-play 
for early threat detection, through modelling the cyber-security 
problem as a zero-sum Markov game between an attacker and defender, 
implemented via DRL agents using PPO. 
The self-play approach ensured that agents were trained against
a pool of previously seen opponents.
The authors also address the fact that work in this area 
typically focuses on a single topology, or scenarios where
agents are limited to a subset of the network.
They address this second challenge through considering dynamic topologies,
where the task complexity is determined via curriculum learning, resulting
in agents that can generalize across topologies.
The authors evaluate their agents using the exploitability metric,
finding that curriculum learning improves the defender’s win
rate, compared to training with a static topology.}

\update{While self-play focuses on computing a best response against the average 
of all past policies, we now turn our attention towards approaches that use 
solution concepts from game theory for determining how to sample past behaviour.}
\cite{lanctot2017unified} introduced Policy-Space Response Oracles (PSRO),
a generalization of the Double-Oracle (DO) approach originally proposed by \cite{mcmahan2003planning},
a theoretically sound approach for finding a minimax equilibrium.
Given the amount of interest that this approach has generated within the literature 
\citep{berner2019dota,perolat2022mastering,vinyals2019grandmaster,lanctot2017unified,li2023combining,oliehoek2018beyond},
we shall dedicate the remainder of this section to DO based approaches.
First we will discuss the DO approach's theoretical underpinnings, before
providing an overview of the work that has been conducted in this area in recent years.
We shall conclude the section with open challenges, in particular with respect to 
scaling this approach to \aco. 

The DO algorithm defines a two-player 
zero-sum normal-form game $\NormalFormGame$, where actions
correspond to policies available to the 
players within an underlying 
stochastic game~$\ExtensiveFormGame$.  
Payoff entries within~$\NormalFormGame$ are determined
through computing the gain~$\mathcal{G}$ for each policy pair
within~$\ExtensiveFormGame$:
\begin{equation}\label{eq:cell_entries}
\mathcal{R}^{\NormalFormGame}_{i}(\langle \action^r_1, \action^c_2\rangle) = 
\mathcal{G}^{\ExtensiveFormGame}_{i}(\JointPolicy{\pi^r_1}{\pi^c_2}).
\end{equation}
In Equation \ref{eq:cell_entries}, $r$ and $c$ refer to the respective
rows and columns inside the normal-form (bimatrix) game, and~$\ExtensiveFormGame$ is the game where 
the policies are being evaluated. 
The normal-form game~$\NormalFormGame$ is subjected 
to a game-theoretic analysis, to find an optimal mixture over actions
for each player.
These mixtures represent a probability distribution 
over policies for the game~$\ExtensiveFormGame$. 
The DO algorithm assumes that both players
have access to a \emph{best response oracle}, 
returning a \emph{best response}~(BR) policy
against the mixture played by the opponent.
BRs are subsequently added to the list of available policies
for each agent.
As a result each player has an additional action 
that it can choose in the normal-form game~$\NormalFormGame$. 
Therefore, $\NormalFormGame$ needs to be augmented through 
computing payoffs for the new row and column entries.
Upon augmenting $\NormalFormGame$ another game theoretic analysis is conducted, and the steps 
described above are repeated.
If no further BRs can be found, then the DO algorithm 
has converged upon a minimax equilibrium~\citep{mcmahan2003planning}.

%
When applying the DO algorithm to 
MARL the oracles must compute
Approximate Best Responses (ABRs)
against a \emph{\mop}.
There are a number of approaches for implementing a
\mop,
\eg sampling individual policies according 
to their respective mixture probabilities at the beginning 
of an episode~\citep{lanctot2017unified}. 
Alternatively, the set of policies can be combined
into a weighted-ensemble, where the 
outputs from each policy are weighted by 
their respective mixture probabilities prior to aggregation.
For generality, we define a \mop as follows:
\begin{definitionsec}[\MoP]\label{def:mixture_of_policies}
$\pi_i^{\mu}$ is a \mop\ for a mixture $\mu_i$ and a corresponding set of policies 
$\Pi_i$ for agent $i$.  
\end{definitionsec}
%

%
Using an oracle 
to obtain an \emph{exact} best response is often also intractable.
In games that suffer from the curse-of-dimensionality 
an oracle can at best hope to find an ABR~\citep{oliehoek2018beyond}.
Here we can apply
\emph{Approximate Double-Oracles} (ADO),
which use linear programming 
to compute an \emph{approximate mixed-strategy NE} 
$\langle \mA{1}, \mA{2} \rangle$ for $\NormalFormGame$,
where $\mA{i}$ represents a \emph{mixture} over policies $\pi_{i, 1 .. n}$
for player $i$.
We therefore require a function $O : \Pi_i^{\mu} \rightarrow \Pi_i$
that computes an ABR $\pi_i$ to a \mop $\pi_i^{\mu}$:
\begin{definitionsec}[Approximate Best Response]\label{def:ABR}
A policy $\pi_i \in \Pi_i$ of player $i$ is an 
approximate best response against a \mop $\pi_j^{\mu}$, \emph{iff},
\begin{equation}
\forall \pi'_i \in \Pi_i, 
\G{i}{\JointPolicy{\pi_i}{\pi^{\mu}_j}}
\geq 
\G{i}{\JointPolicy{\pi'_i}{\pi_j^{\mu}}}.
\end{equation}
\end{definitionsec}
ABRs estimate the exploitability $\Exploitability$ of the current mixtures:
\begin{equation} \label{eq:termination_condition}
\Exploitability \gets
\G{i}{\JointPolicy{\O{i}{\pi^\mu_j}}{\pi^\mu_j}} + 
\G{j}{\JointPolicy{\pi^\mu_i}{\O{j}{\pi^\mu_i}}}
\end{equation}
If $\Exploitability \leq 0$, then the oracle has failed to find 
an ABR, 
and a \emph{resource bounded Nash equilibrium} (RBNE)
has been found~\citep{oliehoek2018beyond}.
Resources in this context refers to the amount of computational power 
available for obtaining an ABR:
\begin{definitionsec}[Resource Bounded Nash Equilibrium] \label{def:RBNE}
Two mixtures of policies $\JointPolicy{\pi^\mu_1}{\pi^\mu_2}$ are a resource-bounded Nash equilibrium (RBNE) \emph{iff},
\begin{equation}
\forall_i \G{i}{\JointPolicy{\pi^\mu_i}{\pi^\mu_j}}
\geq
\G{i}{\JointPolicy{\O{i}{\pi^\mu_j}}{\pi^\mu_j}}.
\end{equation}
\end{definitionsec}

Each agents' oracle is unable 
to find a response that 
outperforms the current \mop against the 
opponent's one.
Therefore, an RBNE has been found. 
The ADO approach is outlined in Algorithm \ref{alg:ADO}
from the perspective of Blue and Red agents for \aco.
The algorithm makes use of a number of functions, including i.) $\funcName{InitialStrategies}$,
for providing an initial set of strategies for each agent, which can also include
rules-based and other approaches; ii.) $\funcName{AugmentGame}$, that computes
missing table entries for the new resource bounded best responses, and; iii.)
$\funcName{SolveGame}$, for computing mixtures once the payoff table has 
been augmented~\citep{oliehoek2018beyond}. 
%

\begin{algorithm}[ht]
\caption{Approximate Double Oracle Algorithm}
\label{alg:ADO}
\providecommand{\commentSymb}{//}
\begin{algorithmic}[1]
\small
\State{$\langle \aA{\blue}, \aA{\red} \rangle \gets \funcName{InitialStrategies}()$}
\State{$\langle \mA{\blue}, \mA{\red} \rangle \gets \langle \{\aA{\blue}\}, \{\aA{\red}\} \rangle$} \Comment{set initial mixtures}
\While{True}
    \State{$\aA{\blue} \gets \funcName{RBBR}( \mA{\red})$}  \Comment{get new res.\ bounded best resp.}
    \State{$\aA{\red} \gets \funcName{RBBR}( \mA{\blue} )$}
    \State{$ \mathcal{G}_{RBBRs} \gets \G{\blue}{\aA{\blue},  \mA{\red} } + \G{\red}{\mA{\blue}, \aA{\red}} $} \Comment{Exploitability.} 
    \If{$ \mathcal{G}_{RBBRs} \leq \epsilon$}
	\State \textbf{break} \Comment{found $\epsilon$-RBNE}
    \EndIf
    \State{$SG \gets \funcName{AugmentGame}(SG, \aA{\blue}, \aA{\red})     $}
    \State{$\langle \mA{\blue}, \mA{\red} \rangle \gets \funcName{SolveGame}(SG)$}
\EndWhile
\end{algorithmic}
\end{algorithm}

%
%

\subsection{Towards Scaling Adversarial Learning Approaches} \label{sec:scale_adv_learning}

\update{The ADO approach outlined above provides a theoretically sound
approach for training cyber-defence and attacking agents within zero-sum 
game settings.
Naturally, for cyber-defence scenarios that represent general-sum games (\eg botnet attacks~\citep{de2021fixed}) 
alternative solvers need to be considered for computing equilibria~\citep{muller2020generalized}. 
Next, considerations are required regarding the scalability of ADO.}

Versions of ADO have been used at scale within other challenging domains, 
e.g., \cite{vinyals2019grandmaster}
achieved a grandmaster level performance in \sctwo,
beating top human players using AlphaStar, an ADO inspired
approach that computes best responses using a match-making mechanism
that samples strong opponents with an increased probability. 
Imitation learning was utilized to obtain an initial pool of agents. 
OpenAI Five made use of similar mixture of self-play method 
and a dynamically-updated meta-distribution over past policies,
beating top human players at Dota~\citep{berner2019dota,perolat2022mastering}. 
Both achievements, although very impressive, come at a high cost with respect
to engineering and training time. 
Indeed, despite their success,  
questions remain regarding the scalability of \ado
to complex domains without making concessions
at the cost of theoretical guarantees. 
\ado approaches are expensive due to: 
%

\textbf{Computing ABRs}
in non-trivial \envs is a lengthy process~\citep{lanctot2017unified}.
%
%
%
%
Meanwhile, even for simple \envs such as Kuhn poker,
\ado can require over 20 ABR 
iterations to approach the value of the game. 
For the slightly more complex Leduc this number grows to over 200, with
further improvements still being obtainable~\citep{lanctot2017unified,li2023combining}.
An even larger number of iterations would likely be necessary for 
\ado to approach $\pi^*_{Blue}$ for \aco.
Computing ABRs from scratch in each iteration 
would therefore be wasteful, especially if the same skills are 
learnt from scratch in each iteration~\citep{liuneupl}.
In addition, due to the ABRs being resource 
bounded~\citep{oliehoek2018beyond}, each agent can end up with a 
population of under-trained policies~\citep{liuneupl}.
An idealized \ado reuses \emph{relevant} knowledge 
acquired over past iterations.
%
%

\textbf{Payoff Matrix Augmentation.} 
The second challenge with respect to scalability is the 
growing normal-form game~$\NormalFormGame$. 
%
Augmenting the payoff table with entries for
new best responses takes exponential time with respect to the 
number of policies~\citep{lanctot2017unified}.
Even for simple games, obtaining a good payoff estimate 
for each $\G{i}{\JointPolicy{\pi^{r}_i}{\pi^{c}_j}}$ can
require thousands of evaluation episodes.
Principled strategies are required for 
keeping the payoff tables to a manageable size.

\textbf{Large Policy Supports.} There is 
an overhead for having to store and utilize 
a large number of function approximators -- one for each policy 
in the support~\citep{ijcai2019-66}. 
%
%
%
%
%

While the above challenges limit the scalability of \ado, there have been
efforts towards remedying this approach without harming too many
of the underlying principles.
\cite{liuneupl} propose Neural Population Learning (NeuPL)
a method that deviates from the standard \ado algorithm PSRO in
two specific ways: i.) all unique policies within the population 
are trained continuously, and; ii.) uses a single policy that contains
the entire population of policies, conditioned on an opponent mixture
identifier. 
This allows for learning to be transferred across policies without a loss 
of generality. 
The approach is capable of outperforming PSRO while maintaining a population
of eight agents on running-with-scissors, which extends the 
rock-paper-scissors game to the spatio-temporal and partially-observed 
Markov game~\citep{vezhnevets2020options}. 

It is also worth noting that the benefits of PSRO have been explored 
within single agent domains, specifically within the context
of robust adversarial reinforcement learning (RARL)~\citep{pinto2017robust}.
For RARL a single agent environment is converted into a zero-sum game
between the standard agent, the protagonist, and an adversarial domain
agent, that can manipulate the environment, \eg through perturbing the 
protagonist's actions. 
\cite{yang2022game} recently applied a variant of PSRO
to this problem, using model agnostic meta learning (MAML)~\citep{finn2017model} 
as the best-response oracle.
The authors conducted an empirical evaluation on MuJoCo environments,
finding that the proposed method outperforms state-of-the-art baselines 
such as standard MAML~\citep{yang2022game}.

\ado has also been applied to general-sum games,
which can have more than one Nash equilibrium. 
This gives rise to the equilibrium selection problem. 
To remedy this, and enable scalable policy evaluation in general, 
\cite{muller2020generalized} apply the
$\alpha$-Rank solution concept, which does not face 
an equilibrium selection problem.
The $\alpha$-Rank solution concept establishes an ordering over policies
within the support of each player.
Specifically, $\alpha$-Rank uses Markov-Conley Chains to
identify the presence of cycles in game dynamics, and
thereby rank policies. 
%
The approach attempts to compute stationary distributions
by evaluating the strategy profiles of $N$ agents through 
an evolutionary process of mutation and selection. 
\cite{yang2020alphaalpha} reviewed the claim
of $\alpha$-ranks tractability. 
The authors find that instead of being
a polynomial time implementation, (with respect to the total
number of pure strategy profiles) solving $\alpha$-Rank
is NP-hard. 
The authors introduce $\alpha^\alpha$-Rank, 
a stochastic implementation of $\alpha$-Rank
that does not require an exponentially-large
transitions matrix for ranking policies, and can terminate early.

\cite{feng2021neural} replace the computation of mixtures
using linear-programming with a Neural Auto-Curricula (NAC), using meta-gradient 
descent to automate the discovery of the learning update rules (the mixtures) 
without explicit human design. 
The NAC is optimized via interaction with the game engine,
where both players aim to minimise their exploitability. 
Even without human design, the discovered MARL
algorithms achieve competitive or even better performance with the SOTA
population-based game solvers on a number of benchmarking environments,
including: Games of Skill, differentiable
Lotto, non-transitive Mixture Games, Iterated Matching Pennies, and Kuhn Poker.
%
%
%
However, the approach does not solve scalability issues with respect
to the increase in time for augmenting the payoff tables, and the time 
required for computing ABRs. 

Finally, we note that DO based methods have also been 
utilized for reducing the number of actions for an agent to choose
from, given a state $\state$ in two player games.
\cite{bakhtin2021no} propose an \ado based algorithm
for action exploration and equilibrium approximation in games with combinatorial 
action spaces: Double Oracle Reinforcement learning for Action exploration (DORA). 
This algorithm simultaneously performs value iteration while learning a policy 
proposal network. 
An \ado step is used to explore additional actions to add to the policy proposals.
The authors show that their approach achieves superhuman performance on a 
two-player variant of the board game Diplomacy.
%

\subsection{Adversarial Learning Approaches Summary} 

\update{
In this section we have formally defined the adversarial learning
challenge that \aco agents are being confronted with, and identified
a theoretically sound adversarial learning approach for our idealised \aco agent,
using the principled ADO approach as a foundation.
However, we have also shown that ensuring convergence upon desirable 
solution concepts is expensive with respect to wall-time, which naturally
represents a significant challenge for cyber-defence. 
Here even training against stationary opponents can be time-consuming.  
Therefore, before selecting approaches discussed 
in \autoref{sec:hdss_approaches} and \autoref{sec:hdas_approaches}
for our idealized \aco agent, their suitability for wall-time efficient 
adversarial learning must be considered.
Based on the above, the quest for \ado approaches that scale, while keeping
the properties upon which the convergence guarantees rely (more or less)
intact, remains an open challenge.} 

\update{Finally, while ADO approaches abstract a cyber-defence 
scenario to a matrix game, there exist a number of other tools from game theory that
can be considered for abstracting cyber-defence scenarios.
For example, \cite{TANG2024103871} model network attack-defence mechanisms and
cooperative (multi-agent) responses using a Stackelberg hypergame model;
This can capture network characteristics such as misinformation and errors
in information and transmission. 
This model allows for the construction of a collaborative defence strategy through
an analysis of the Nash equilibrium (or equilibria) within the Stackelberg hypergame.
Therefore, exploring population based training using different types of game 
formulations represents an interesting future avenue for adversarial learning
within the context of autonomous cyber-defence.}

\section{Discussion and Open Questions} \label{sec:challenges}

The previous sections provide a comprehensive overview
of the plethora of methods for addressing: high-dimensional
states; large combinatorial action spaces; \update{facilitating
multi-agent cooperation}, and; 
reducing the exploitability of DRL agents.
We shall now consider how the lessons learned from this
process can help us formulate an idealised learner for \aco,
and distill open research questions.
 
\update{Given that many of today's networks require
decentralized cyber-defence solutions, 
an idealised learner will often also
be expected to coordinate and negotiate
with other cyber-defence agents when called upon.
To this end, we consider the extent to which
the solutions identified within this paper
can feature within two existing autonomous
cyber-defence agent reference architectures, namely the 
NATO IST 152 RTG’s Multi Agent System for Cyber (MASC) defence~\citep{theron152}
and the Autonomous Intelligent Cyber-defence Agent (AICA)~\citep{kott2018autonomous}.}

\update{The MASC and AICA reference architectures both focus on cyber-defence scenarios 
where decentralized cooperative multi-agent solutions are required, on-the-edge, to defend 
a set of military platforms such as unmanned vehicles (UXVs)~\citep{kott2018autonomous}.
While the solutions identified in this survey are more broadly
applicable across cyber-defence scenarios, below we consider 
the extent to which they can address the prerequisites for
developing the AICA and MASC architectures,  
along with the additional developments required for
for learning \aco agents to be deployed on realistic scenarios.
First, we will consider how our identified learning methods can provide suitable 
modules for addressing a number of the AICA's architecture's high-level 
functions, including \emph{sensing and world state identification},
\emph{action selection}, \emph{collaboration and negotiation}~\citep{kott2018autonomous}.
We conclude the section with considerations regarding how to \emph{counter rapidly evolving enemy attacks}
and the need for principled evaluation metrics to measure both the performance and vulnerability 
of cyber-defence agents.}

\subsection{\update{Sensing and World State Identification}}

\update{In \autoref{sec:hdss_approaches} we propose using
a \emph{system-of-systems} approach to enable our idealised learning agent to
digest a sufficient amount of cyber-defence data to facilitate sound decision making.
This solution starts with data processing modules for asserting the state of each network devices.
Example tasks here range from malware detection to suspicious activities.
Naturally, these modules may themselves 
benefit from ML solutions~\citep{lyu2023survey,mohammadi2019cyber,moore2017feature}.}

\update{The outputs from these individual modules can subsequently be mapped to a suitable
data structure. 
In \autoref{sec:function_approximators}, 
we advocate for a graph-based data structure,
enabling the application of SOTA graph machine learning approaches, 
which are increasingly gaining traction within 
the sphere of \aco~\citep{biswas2023intrusion,mitra2024use,ruan2023deep}.
However, while GNNs are capable of processing large graphs, 
large computer networks could remain challenging
for DRL-GNN solutions. 
In particular, critical information could be lost by
graph machine learning approaches through message
aggregation and pooling operations for obtaining 
compact graph representations~\citep{di2023over}. 
To address this challenge, we have provided an overview 
of SOTA state-abstraction techniques from the DRL literature,
and outlined future work directions of combining these methods
with Variational Graph Auto Encoders.}

\update{Our identified solutions and proposed system-of-systems
approach addresses a number of the requirements 
listed by the AICA reference architecture, 
including \aco agents being able to: 
i.) Observe the state and activities within the elements
under the \aco agent's protection;
ii.) Maintain a world model for observing and understanding the environment, 
and; iii.) Observe all relevant communication traffic. 
Implementing a modular sensing and world state identification framework featuring 
ML modules for edge based classifications, along 
with a representative graph based world model augmented by  
state abstraction techniques, 
would allow us to answer this research question:}
\begin{description}
\item[\textbf{Q1:}] \update{Can our proposed system-of-systems approach
deliver compact and informative observations that enable sound decision making 
by DRL agents deployed on real-world network infrastructure?}
\end{description}

\subsection{\update{Action Selection}}
 
\update{The MASC and AICA architectures consider desirable actions that an \aco
agent must be able to execute, including monitoring~\citep{theron152} and 
taking destructive actions, such as deleting or 
quarantining compromised software and data~\citep{kott2018autonomous}.
In this paper we extend the action space requirements for idealized
\aco agents to include the ability to master combinatorial high-dimensional
action spaces.
This challenge will confront the majority of cyber-defence agents, 
even in scenarios where decentralized cyber-defence solutions watch
over reasonably sized subnets. 
We address this requirement in \autoref{sec:hdas_approaches}
through providing an overview of DRL approaches that are explicitly
designed for dealing with a high-dimensional action spaces.}

\update{Current DRL approaches for \aco often address the high-dimensional action 
problem through using domain knowledge, \eg to eliminate redundant actions.
In contrast, our work recommends tackling the high-dimensional action space
problem head-on, through applying and adapting the high-dimensional
action space approaches discussed in \autoref{sec:hdas_approaches}.
While there have been works on solving this challenge 
from the perspective of penetration 
testing~\citep{tran2022cascaded,nguyen2020multiple},
currently there is a lack of literature evaluating the extent to
which these approaches can help solve the high-dimensional
action space challenge for cyber-defence agents. 
Therefore, given that there now exists a plethora
of cyber-defence environments that confront a 
MADRL agent with the high-dimensional action space
challenge, an extensive benchmarking
of current solutions could answer the research question:}
\begin{description}
\item[\textbf{Q2:}] \update{Which current high-dimensional action space DRL approaches can scale to realistic cyber-defence challenges?}
\end{description}

\subsection{\update{Collaboration and Negotiation}}
 
\update{Given that tactical military networks will often consist of (mobile) ad-hoc network
architectures deployed within a heavily contested environment, the MASC and AICA architectures 
stress the need for decentralized cyber-defence solutions. 
These solutions need to be capable of
functioning autonomously when necessary, while being able to collaborate with friendly
agents, using collaboration schemes and negotiation mechanisms~\citep{kott2018autonomous}.  
Furthermore, given the contested nature of communications and cyber-defence environments,
our idealised cyber-defence solution needs to be able to cope 
with denied, degraded, intermittent and limited (DDIL)
communication~\citep{kott2018autonomous}.
Therefore, decentralized cyber-defence agents require autonomous 
independent decision making on the edge when necessary, without 
depending on the support of other agents or an external controller.}

\update{Throughout this \report we observe that implementing
multi-agent learning algorithms that can enable learning agents to 
converge upon a set of joint-policies capable of cooperation and coordination is non-trivial~\citep{hernandez2017survey,hernandez2018multiagent}.
To formally define the cooperative multi-agent challenge, 
\autoref{sec:background:marl_pathologies} provides an overview of 
the multi-agent learning pathologies
that MADRL approaches must overcome, in order to converge upon (near) optimal
joint policies. 
In \autoref{sec:hdas_approaches:marl_and_coordination}, we outline a number
of state-of-the-art MADRL approaches that can enable the cyber-defence 
agents to cooperate with other entities within a network.
While this section provides recommendations regarding methods that could be suitable
for the decentralized cyber-defence architectures envisaged by \cite{kott2018autonomous}
and \cite{theron152}, we recommend a number of future research avenues 
to better inform the use of MADRL components for our idealized \aco agent. 
These avenues include an extensive benchmarking
of current SOTA MADRL approaches on suitable cyber-defence environments,
an exploration of the impact of DDIL communication on MADRL-Comms approaches~\citep{chen2021graph,su2020counterfactual,xiao2023graph,contractor2024learning}, 
and a game theoretic evaluation of MADRL agents situated in cyber-defence environments where
the objectives of Blue agents are not perfectly aligned, \eg through the lens of 
\emph{sequential social dilemmas}~\citep{leibo2017multi}.}

\update{Game theoretic evaluations could be particular insightful for scenarios where 
agents are only required to collaborate with other friendly agents when a 
need arises and conditions permit, requiring collaboration schemes and negotiation 
mechanisms~\citep{kott2018autonomous,zhou2016multiagent,georgila2014single,chang2021multi,tang2019towards}.
CybORG CAGE Challenges 3-4~\citep{cage_challenge_3_announcement,cage_challenge_4_announcement} represent
suitable environments for the above evaluations, given suitable extensions that allow for self-interested agents..
An evaluation could answer the research questions:
\begin{description}
\item[\textbf{Q3:}] \update{Which current MADRL approaches can scale to challenging cyber-defence scenarios that require decentralized solutions?}
\item[\textbf{Q4:}] \update{What insights can be gained from conducting an empirical game theoretic evaluation on cyber-defence scenarios that require coordination, cooperation and negotiation?}
\end{description}}

\update{In addition to the above considerations, both \cite{kott2018autonomous} and \cite{theron152} observe that an idealized \aco agent
needs to be capable of interacting with humans. 
Naturally, human operators within military scenarios are often not cybersecurity geeks; 
when dealing with stressful scenarios, there is a risk of defence operators experiencing a mental
overload if systems are not designed appropriately 
This could impact the success of a mission~\citep{theron152}.  
To facilitate the development of cyber-defence methods designed to avoid
overloading a human operator, we recommend a survey by \cite{da2020agents}.
This focuses on the agents teaching agents paradigm, where one of the agents can be a human.
The survey provides an overview of approaches designed for determining the best time for 
approaching a human operator and how to capitalize on the interaction.} 
\update{This raises the research question:}
\begin{description}
\item[\textbf{Q5:}] \update{Can we implement a cyber-defence agent capable of 
learning \emph{when} and \emph{how} to interact with a human operator 
without causing a mental overload?}
\end{description}

\subsection{Countering Rapidly Evolving Enemy Attacks}

\update{\cite{kott2018autonomous} stress the need for an autonomous cyber-defence
agent capable of countering the ability of enemy malware to rapidly
evolve its  capabilities and tactics, techniques, and 
procedures (TTPs).
In \autoref{sec:adv_learning} we consider the adversarial learning
challenge within the context of cyber-defence, acknowledging the fact that 
turning to ML methods to mitigate attacks also opens up an 
additional attack-surface~\autoref{sec:pertubation_attacks}.}

\update{An idealised learning \aco agent should be resistant to compromise~\citep{kott2018autonomous}.
Here we observe that ML based cyber-defence components, from any of the
three paradigms of supervised, unsupervised and reinforcement learning,
are susceptible to adversarial attacks. 
An idealized \aco-DRL agent, or rather a system-of-systems
featuring a DRL based \aco agent, must be capable of mitigating perturbation and other 
types of adversarial attacks on ML models.
Therefore, a diverse set of potential attacks must
already be considered during training, 
in order to ensure that our idealized cyber-defence
agents can generalize across potential opponents once deployed.}

\update{A further requirement by \cite{kott2018autonomous} is 
that, in order to learn responses
to new capabilities, techniques, and procedures of the enemy 
malware,  learning should occur both offline and online. 
However, for scenarios where online learning is possible, 
we observe that there exists a risk/benefit trade-off.
While online learning can offer benefits such as learning to adapt to a new opponent,
our summary of the adversarial learning challenges in \autoref{sec:adv_learning} 
highlights a number risks, including overfitting
on the current opponent~\citep{lanctot2017unified}, 
and leaving \aco agent vulnerable to policy poisoning attacks~\citep{han2020adversarial}. 
However, deploying a cyber-defence policy that remains fixed upon deployment
limits the extent to which suitable counter strategies can be learnt against
novel attacks. 
Therefore, our idealized \aco-DRL agent can conduct online learning, 
in a principled fashion, using suitable counter measures towards poisoning and 
perturbation attacks, along with methods to avoid cross-policy correlation
and catastrophic forgetting.}

%
%
In \autoref{sec:adv_learning} we provided an overview of the current
state-of-the-art of adversarial learning approaches designed
to mitigate the above challenges, along with a summary of the
extent to which adversarial learning has been utilized to date
within the \aco literature.
We propose that the adversarial learning challenge should be addressed 
via principled theoretically sound methods, of which 
approximate best response techniques, such as the approximate
double oracle approach, stand out as a means for obtaining
robust and resilient cyber-defence agents, in particular 
given their successful application to environments
that share many properties with \aco environments~\citep{vinyals2019grandmaster,berner2019dota,perolat2022mastering}.
However, while computing an (approximate) best response 
to measure exploitability is a reasonable expenditure,
best response techniques require this in \emph{every} iteration. 
As a result the literature on adversarial learning 
often focuses on simple games for benchmarking
such as Kuhn and Leduc poker, where the value of the game is known~\citep{lanctot2019openspiel,
feng2021neural,li2023combining,lanctot2017unified}.
Meanwhile, methods that we have identified as suitable 
for \aco can require lengthy training times.
This raises the following research questions:
\begin{description}
\item[\textbf{Q6:}] Can we implement an \emph{efficient} best response framework for cyber defence?
\end{description}

While RQ5 focuses on mechanisms within the approximate best response
framework to reduce training time (\eg through knowledge reuse~\citep{liuneupl}),
the endeavour is underpinned by the need for methods that themselves
can be trained efficiently.
However, this is not a property one would associate
with the methods discussed in 
sections~\ref{sec:hdss_approaches}~and~\ref{sec:hdas_approaches}, 
raising the question:
\begin{description}
\item[\textbf{Q7:}] Can we implement \emph{wall time efficient} best response oracles for \aco?
\end{description}

\update{Finally, among the prerequisites listed for the development of AICA's architecture,
\cite{kott2018autonomous} highlight the need for stealthy agents, that minimize
the probability with which an adversary might detect or observe
the defence agent's presence, while observing and interacting with
its environment.
The extent to which our idealised \aco-DRL agent will adopt
stealthy behaviour will depend on factors such as: i.) the reward
function formulation; ii.) environment design factors, such as the extent
to which the agent can interact with the network without revealing itself
to \red, and; iii.) stealthy behaviour outperforming non-stealthy 
behaviour when trained against different types of \red agents.
This last point raises interesting questions regarding the 
characteristics of cyber-defence agents obtained via different
learning processes, specifically, given a specific network 
architecture, what are the policy characteristics of 
an idealized \aco-DRL agent that has minimized exploitability,
and has found a means through which to behave in a robust and
resilient manner.}

\subsection{The need for improved evaluation metrics}

After reading \autoref{sec:adv_learning}, it should be
evident that success achieved against a stationary opponent 
should be enjoyed with caution.
For example, for the CAGE 
challenges~\citep{cage_challenge_announcement,cage_challenge_2_announcement,cage_challenge_3_announcement},
the evaluation process to-date has consisted of evaluating the submitted 
approach against rules-based Red agents with different trial lengths. 
This formulation is concerning. 
Solutions that overfit on the provided Red agents are likely to perform best.
To address this shortcoming we advocate for \emph{exploitability} being 
used as a standard evaluation metric, given that it is widely used 
by the adversarial learning community~\citep{lanctot2017unified,oliehoek2018beyond,heinrich2016deep}, 
raising the research question:
\begin{description}
\item[\textbf{Q8:}] How exploitable are current approaches for autonomous cyber defence?
\end{description}
\update{This question needs to be considered within the context of realistic cyber-defence 
scenarios, where our Blue cyber-defence agent is confronted with all of the challenges
that we discussed in this report with respect to combinatorially large observation and
action spaces. 
For example, reducing the exploitability of a cyber-defence agent 
that can learn to utilize a large action space, such as PrimAITE's full action space, 
would represent a first step towards obtaining a robust and resilient learning \aco 
agent for real world deployment.}

\section{Conclusions} \label{sec:conclusion}

In this work we surveyed the DRL
literature to formulate an idealised learner for
autonomous cyber defence.
This idealised learner sits at the intersection of three
active areas of research, namely environments that confront
learners with the \cod, with respect to both state and action
spaces, and adversarial learning.
While significant efforts have been made on each of these 
topics individually, each still remains an active 
area of research.

While SOTA DNNs have
allowed RL approaches to be 
scaled to domains that were previously considered
high-dimensional~\citep{mnih2013playing,mnih2015human},
in this \report we provide an overview of solutions for environments
that push standard DRL approaches
to their limits.
%
%
\commentout{Principled approaches are required
for learning abstractions for all (relevant) areas of 
a state space~\citep{pmlr-v119-misra20a} and for encouraging 
agents to take actions that are likely to result in new outcomes, 
which is critical within environments that yield sparse rewards~\citep{ladosz2022exploration}. 
Solutions to these problems are vital for scaling
DRL to the full \aco problem, with its vast amounts 
of data, and where the raw multi-discrete observations can consist of a large 
number of features~\citep{cage_cyborg_2022}.}
\commentout{Our \report provides a comprehensive overview of approaches designed
for high-dimensional action spaces, and an evaluation regarding the 
extent to which the approaches meet the criteria of being able
to generalize over actions, handle a potentially time varying action space,
and feature sub-linear complexity \wrt determining which
action to take.  
We identify five categories of high-dimensional action approaches:
proto action based approaches, 
action decomposition,
action elimination,
hierarchical approaches, and
curriculum learning.
Out of these five categories we identify two 
that have shown promise for \aco domains,
from the perspective of Red
, namely the \WP
architecture (proto action)~\citep{nguyen2020multiple} and 
CRLA (action decomposition)~\citep{tran2022cascaded}.
\commentout{While other approaches also have their merits, 
we believe the \WP
and CRLA inspired approaches represent
a natural starting point for considering the
third \aco challenge: adversarial learning.}
In \autoref{sec:adv_learning} we have formally
defined the adversarial learning challenge, and
discussed why finding an optimal 
policy will be non-trivial.
\commentout{We point out that adversarial learning
is already challenging within far simpler
games than the \aco problem,
\eg the game tree for Kuhn poker is tiny,
and yet days of training and principled methods are 
required for DRL to converge upon a 
resource bounded Nash equilibrium~\citep{lanctot2017unified}.}
Here, we provide a recap of one of 
the most popular approaches towards limiting exploitability, Double 
Oracle methods, along with an overview of efforts
towards scaling this family of approaches to high-dimensional
settings.}
\commentout{Given an approach that meets the above criteria, 
suitable benchmarking environments are required
to conduct an empirical evaluation.
However, in \autoref{sec:envs} we observe that even 
benchmarking environments designed to present learners
with the \aco challenge, don't feature all of the properties
present in a real cyber defence scenario.
To address this, we review multiple environments and domains,
and discuss the extent to which they feature cyber defence properties.
While none of the \envs pose all of the challenges present in \aco, 
each challenge is represented in at least one domain, and we highlight 
the most suitable ones, including SUMO~\citep{SUMO2018}; two \sctwo environments,
SMAC~\citep{samvelyan19smac,ellis2022smacv2} and Micro-RTS~\citep{huang2021gym}; 
and, RecSim~\citep{ie2019recsim}, a recommender system environment.}
In sections~\ref{sec:hdss_approaches} -- \ref{sec:adv_learning} 
we identify components for the implementation and evaluation 
of an idealised learning agent for \aco, and in \autoref{sec:challenges} 
we discuss both
theoretical and engineering challenges that need to be overcome 
in-order to: implement our idealised agent, and; train it at scale.
We hope that this survey will raise awareness regarding
issues that need to be solved in-order for DRL to be scalable to challenging cyber 
defence scenarios, and that our work will inspire readers to attempt 
to answer the research questions we have distilled from the
literature.  

%
%
%
%
%
%

\section{Acknowledgements}

Research funded by Frazer-Nash Consultancy Ltd. on behalf of the 
Defence Science and Technology Laboratory (Dstl) which is an 
executive agency of the UK Ministry of Defence providing world 
class expertise and delivering cutting-edge science and technology 
for the benefit of the nation and allies. 
The research supports the Autonomous Resilient Cyber Defence (ARCD) 
project within the Dstl Cyber Defence Enhancement programme.

\section{License}

This work is licensed under the Creative Commons Attribution 4.0 International License. 
To view a copy of this license, visit http://creativecommons.org/licenses/by/4.0/
or send a letter to Creative Commons, PO Box 1866, Mountain View, CA 94042, USA.

\begin{appendices}
\section{High-Dimensional State Space Approaches Overview} \label{appendix:states}

\begin{table}[h]
\centering
\resizebox{0.95\columnwidth}{!}{
\begin{tabular}{ |p{5cm}||p{12cm}|l|l|l|}
 \hline
 \multicolumn{5}{|c|}{\textbf{Papers that are relevant \wrt the high-dimensional state spaces}} \\
 \hline
 \textbf{Work} & 
 \textbf{Summary} &
 \textbf{Abstr.} &
 \textbf{Expl.} &
 \textbf{CF} \\  
\hline
\cite{pmlr-v80-abel18a} & 
Authors introduce two classes of abstractions: 
\emph{transitive} and \emph{PAC} state abstractions, 
and show that transitive PAC abstractions can be acquired efficiently, 
preserve near optimal-behavior, and experimentally reduce sample complexity 
in \emph{simple domains}.
& \checkmark & \xmark & \xmark \\
\hline
\cite{abel2016near} & 
Authors investigate approximate state abstractions, 
Present theoretical guarantees of the quality of behaviors derived 
from four types of approximate abstractions. 
Empirically demonstrate that approximate abstractions lead to reduction 
in task complexity and bounded loss of optimality of behavior in a variety of environments.
& \checkmark & \checkmark & \xmark \\
\hline
\cite{abel2019state} & Seek to understand the role of 
information-theoretic compression in state abstraction for
sequential decision making, resulting in a novel objective function. & \checkmark & \checkmark & \xmark \\
\hline
\cite{pmlr-v108-abel20a} & Combine state abstractions and options to preserve the representation of near-optimal policies. & \checkmark & \checkmark & \xmark \\
\hline
\cite{atkinson2021pseudo} & A generative network is used to generate short sequences from previous tasks for the DQN to train on, in order to prevent catastrophic forgetting as the new task is transferred. & \xmark & \xmark & \checkmark \\
\hline
\cite{badianever} & 
Construct an episodic memory-based intrinsic
reward using k-nearest neighbours over recent experiences.
Encourage the agent to repeatedly revisit all states in its
environment.
& \xmark & \checkmark & \checkmark \\
\hline
\cite{bellemare2016unifying} & Use Pseudo-Counts to count salient events, 
derived from the log-probability improvement according to a 
\emph{sequential density model} over the state space. & \checkmark & \checkmark & \xmark \\
\hline
\cite{bougie2021fast} & 
%
Introduce the concept of fast and slow curiosity that aims to 
incentivise long-time horizon exploration. 
Method decomposes the curiosity bonus into a fast reward that deals 
with local exploration and a slow reward that encourages global exploration. 
& \xmark & \checkmark & \xmark \\
\hline
\cite{burdaexploration} & 
Introduce a method to flexibly combine intrinsic and extrinsic rewards 
via a \emph{random network distillation} (RND) bonus 
enabling significant progress on several hard exploration Atari games.
& \xmark & \checkmark & \xmark \\
\hline
\cite{burden2018using} &
Automate the generation of Abstract Markov Decision Processes (AMDPs)
using uniform state abstractions. 
Explores the effectiveness and efficiency of different 
resolutions of state abstractions. & \checkmark & \checkmark & \xmark \\
\hline
\cite{burden2021latent}  & 
Introduce Latent Property State Abstraction, 
for the full automation of creating and solving an
AMDP, and apply potential function for potential based reward
shaping. & \checkmark & \checkmark & \xmark \\
\hline
\cite{pmlr-v97-gelada19a} &
Introduce DeepMDP, where the $\ell_2$ distance represents an upper bound
of the bisimulation distance, learning embeddings that ignore irrelevant 
objects that are of no consequence to the learning agent(s). & \checkmark & \checkmark & \xmark \\
\hline
\cite{colas2019curious} & 
Proposes CURIOUS, an algorithm that leverages 
a modular Universal Value Function Approximator 
with hindsight learning to achieve a diversity 
of goals of different kinds within a unique policy 
and an automated curriculum learning mechanism 
that biases the attention of the agent towards 
goals maximizing the absolute learning progress. 
Also focuses on goals that are being forgotten.
& \xmark & \checkmark & \checkmark \\
\hline
\cite{de2015importance} & Study the extent to which experience replay memory composition can mitigate catastrophic forgetting. & \xmark & \xmark & \checkmark \\
\hline
\cite{precup2000temporal} & Introduction of the \emph{options} framework, for prediction, control and learning at multiple timescales.~\footnote{A substantial amount of follow-on work exists from this work.} & \checkmark & \checkmark & $\nearrow$ \\
\hline
\cite{fangadaptive} & Introduce Adaptive Procedural Task Generation (APT-Gen), an approach to
progressively generate a sequence of tasks as curricula to facilitate reinforcement learning
in hard-exploration problems. 
& \xmark & \checkmark & \xmark \\
\hline
\cite{forestier2017intrinsically} & 
Introduce an intrinsically motivated goal exploration processes with
automatic curriculum learning.
& \checkmark & \checkmark & \xmark \\
\hline
\cite{fu2017ex2} & 
Propose a novelty detection algorithm for exploration. 
Classifiers are trained to discriminate each visited state 
against all others, where novel states are easier to distinguish. 
& \xmark & \checkmark & \xmark \\
\hline
\cite{9287851} & Map high-dim video to a low-dim discrete latent representation using a VQ-AE. & \checkmark & \checkmark & \xmark \\
\hline
\cite{hester2013learning} & 
Focus on learning exploration in model-based RL.
& \xmark & \checkmark & \xmark \\
\hline
\cite{kessler2022same} & 
Show that existing continual learning methods based on single neural network predictors with shared replay buffers fail in the presence of interference. 
Propose a factorized policy, using shared feature extraction layers, but separate heads, each specializing on a new task to prevent interference. 
& \xmark & \xmark & \checkmark \\
\hline
\cite{kovac2020grimgep} & Set goals in the region of highest uncertainty. Exploring uncertain states \wrt the rewards are the sub-goals. & \checkmark & \checkmark & \xmark \\
\hline
\cite{kulkarni2016hierarchical} & Introduce a hierarchical-DQN (h-DQN) operating at different temporal scales. & \checkmark & \checkmark & \xmark \\
\hline
\cite{dietterich2000overview} & An overview of the MAXQ value function decomposition and its support for state and action abstraction. & \checkmark & \checkmark & \xmark \\
\hline
\cite{machado2017laplacian} & 
Introduce a Laplacian framework for option discovery. & \checkmark & \checkmark & \xmark \\
\hline
\cite{machadoeigenoption} & 
Look at \emph{Eigenoptions}, options obtained from representations that encode diffusive information 
flow in the environment. 
Authors extend the existing algorithms for Eigenoption discovery 
to settings with stochastic transitions 
and in which handcrafted features are not available. 
& \checkmark & \checkmark & \xmark \\
\hline
\cite{pmlr-v119-misra20a} & Authors introduce\ \HOMER,
an iterative state abstraction approach that accounts for the fact
that the learning of a compact representation for states
requires comprehensive information from the environment. & \checkmark & \checkmark & \xmark \\
\hline
\cite{martin2017count} & Count-based exploration in feature space 
rather than for the raw inputs. & \checkmark & \checkmark & \xmark \\
\hline
\cite{pathak2017curiosity} & Intrinsic rewards based method 
that formulates curiosity as the error in an agent’s ability to predict the consequence 
of its own actions in a visual feature space learned by a self-supervised inverse dynamics model.
& \checkmark & \checkmark & \xmark \\
\hline
\cite{savinovepisodic} & 
Propose a new curiosity method which uses episodic memory to form the novelty bonus. 
Current observation is compared with the observations in memory. 
%
& \checkmark & \checkmark & \xmark \\
\hline
\cite{stadie2015incentivizing} & 
Evaluate sophisticated exploration strategies, 
including Thompson sampling and Boltzman exploration, 
and propose a new exploration method based on assigning 
exploration bonuses from a concurrently learned 
model of the system dynamics. & \checkmark & \checkmark & \xmark \\
\hline
\cite{ribeiro2019multi} & Train DRL on two similar tasks, augmented with EWC to mitigate catastrophic forgetting. & \xmark & \xmark & \checkmark \\
\hline
\cite{tang2017exploration} & Use an auto-encoder and SimHash for 
to enable count based exploration. & \checkmark & \checkmark & \xmark \\
\hline
\cite{vezhnevets2017feudal} & 
Introduce FeUdal Networks (FuNs): a novel architecture for hierarchical RL.
FuNs employs a Manager module and a Worker module. 
The Manager operates at a slower time scale and sets abstract goals 
which are conveyed to and enacted by the Worker. 
The Worker generates primitive actions at every tick of the environment. 
& \checkmark & \checkmark & \xmark \\
\hline
\cite{zhanglearning} & Propose \emph{deep bisimulation for control} (DBC). 
DBC learns directly on this bisimulation distance metric. 
Allows the learning of invariant representations that can be used effectively for 
downstream control policies, and are invariant with respect to task-irrelevant details. & \checkmark & \checkmark & \xmark \\
\hline
\end{tabular}}
\caption{An overview of approaches for addressing the \cod \wrt states, and the
extent to which works cover the topics of abstraction, \emph{advanced} exploration 
strategies, and the mitigation of catastrophic forgetting.} 
\label{tab:state_abstraction}
\end{table}

\newpage
\section{High-Dimensional Action Space Approaches Overview} \label{appendix:action_approaches}

\begin{table}[h]
\centering
\resizebox{\columnwidth}{!}{
\begin{tabular}{ |l||l|p{5cm}|l|l|l|l|}
 \hline
 \multicolumn{7}{|c|}{\textbf{High-Dimensional Action Space Approaches Overview}} \\
 \hline
 \textbf{Algorithm} & 
 \textbf{Approach} &  
 \textbf{Summary}  &  
 \textbf{Applications} & 
 \textbf{SLC} & 
 \textbf{Gen.} & 
 \textbf{TVA} \\  \hline
\hline
Wolpertinger Architecture~\citep{dulac2015deep} & Proto Actions & Passes $k$ nearest neighbours of proto action obtained from actor to critic for evaluation. & Flexible & \checkmark & \checkmark & $\nearrow$ \\
\hline
DDPG+$k$-NN Slate-MDP Solver~\citep{sunehag2015deep} & Proto Actions & Uses DDPG to learn Slate Policies and Guide Attention. & Slate-MDPs & \checkmark & \checkmark & \checkmark\\
\hline
DeepPage~\citep{10.1145/3240323.3240374} & Proto Actions & Proposes solution for page-wise recommendations based on real-time feedback. & Slate-MDPs & \checkmark & \checkmark & \checkmark\\
\hline
LIRD~\citep{DBLP:journals/corr/abs-1801-00209} & Proto Actions & Authors introduce an online user-agent interacting environment simulator and a LIst-wise Recommendation framework LIRD. & Slate-MDPs & \checkmark & \checkmark & \checkmark\\
\hline
CRLA~\citep{tran2022cascaded,tran2021deep} & Decomp. \& Hierarchical & Cascaded \RL agents that algebraically construct an action index. & Flexible & \checkmark & \checkmark & \xmark\\
\hline
\textsc{Sequential} DQN~\citep{metz2017discrete} & Decomposition & Discrete Sequential Prediction of Continuous Actions & Multi-Actuator  & \checkmark & \checkmark & \xmark \\
\hline
BDQ~\citep{tavakoli2018action} & Decomposition & Branching Dueling Q-Network for fine control of discretized continuous control domains. & Multi-Actuator & \checkmark & \checkmark & $\nearrow$ \\
\hline
Cascading DQN~\citep{pmlr-v97-chen19f} & Decomposition & Model based RL approach for recommender systems that utilizes a Generative Adversarial Network (GANs)
to imitate user behaviour dynamics and reward function.
 & Slate-MDPs & \checkmark & \checkmark & \xmark\\
\hline
Parameter Sharing TRPO~\citep{gupta2017cooperative} & Decomposition & Cooperative Multi-Agent Control Using Deep Reinforcement Learning. & Multi-Actuator & \checkmark & \checkmark & \xmark\\
\hline
SCORE~\citep{9507301} & Decomposition & Structured Cooperative Reinforcement Learning With Time-Varying Composite Action Space. & General & \checkmark & \checkmark & $\nearrow$ \\ 
\hline
Action-Elimination DQN (AE-DQN)~\citep{zahavy2018learn} & Elimination & Combines a DRL algorithm with an
\emph{Action Elimination Network} that eliminates sub-optimal actions. & Text-based games & \checkmark & \xmark & \xmark\\
\hline
Top-$k$ Candidate Generator~\citep{10.1145/3289600.3290999} & Elimination & Adapts the REINFORCE algorithm into a top-$k$ neural candidate generator for large action spaces. & Slate-MDPs & \checkmark & \xmark & \xmark\\
\hline
PATRPO \& PASVG(0)~\citep{wei2018hierarchical} & Hierarchical & Hierarchical architecture for the tasks where the parameter policy is conditioned on the output of the discrete action
policy. & PA-MDPs & \checkmark & \xmark & \xmark\\
\hline
Growing Action Spaces~\citep{pmlr-v119-farquhar20a} & Curriculum Learning & CL approach where agents gradually learn large action spaces, based on a hierarchical action space defined using domain knowledge. & Flexible & \checkmark & \xmark & \xmark\\
\hline
Progressive Action Space~\citep{yu4167820curriculum} & Curriculum Learning & Curriculum offline \RL with a progressive action space. & Flexible & \checkmark & \xmark & \xmark\\
\hline
\end{tabular}}
\caption{A summary of approaches designed for high dimensional action spaces. 
The final three columns are abbreviated: Sublinear Complexity (SLC), Generalizeabilty (Gen.) and
Time Varying Actions (TVA).
The $\nearrow$ symbol indicates that steps have been taken towards addressing one of the above criteria. 
Wolpertinger for instance can handle TVA within the current embedding space, but would require retraining
if an action embedding axis was added/removed.
}
\label{tab:high_dim_action_approaches}
\end{table}

\end{appendices}


\bibliography{main}


\end{document}